\documentclass[preprint]{elsarticle}
\usepackage[utf8]{inputenc}
\usepackage[T1]{fontenc}
\usepackage{amsmath,amssymb}
\usepackage{microtype}
\usepackage{graphicx}
\usepackage{subcaption}
\usepackage{booktabs}
\usepackage{adjustbox}
\usepackage{pifont}
\usepackage{multirow}
\usepackage{xcolor}
\PassOptionsToPackage{hyphens}{url}
\usepackage{hyperref}
\usepackage{cleveref}
\usepackage{tikz}
\usetikzlibrary{arrows.meta}
\usepackage{todonotes}

\usepackage{placeins}

\hypersetup{
  colorlinks=true,
  linkcolor=blue,
  citecolor=blue,
  urlcolor=blue,
  pdfpagelayout=SinglePage
}

\definecolor{deepblue}{HTML}{2E86AB}

\newcommand{\nsubstep}{N_{\text{sub}}}
\newcommand{\cmark}{\ding{51}}
\newcommand{\xmark}{\ding{55}}

\begin{document}

\begin{frontmatter}

\title{Learning to Advect: A Neural Semi-Lagrangian Approach for Data-Driven Weather Forecasting}

\author[ECCC]{Carlos A. Pereira\fnref{equal}}
\author[ECCC]{St\'ephane Gaudreault\corref{cor}\fnref{equal}}
\ead{stephane.gaudreault@ec.gc.ca}
\author[ECCC]{Valentin Dallerit}
\author[ECCC]{Christopher Subich}
\author[ECCC]{Shoyon Panday}
\author[ECCC]{Siqi Wei}
\author[ECCC]{Sasa Zhang}
\author[UBC]{Siddharth Rout}
\author[UBC]{Eldad Haber}
\author[USASK]{Raymond J. Spiteri}
\author[UofR]{David Millard}

\address[ECCC]{Recherche en pr\'evision num\'erique atmosph\'erique, Environnement et Changement climatique Canada, 2121 Route Transcanadienne, Dorval, Qu\'ebec H9P 1J3, Canada}
\address[UBC]{Department of Earth, Ocean and Atmospheric Sciences, University of British Columbia, 2020--2207 Main Mall, Vancouver, British Columbia V6T 1Z4, Canada}
\address[USASK]{Department of Computer Science, University of Saskatchewan, 176 Thorvaldson Building, 110 Science Place, Saskatoon, Saskatchewan S7N 5C9, Canada}
\address[UofR]{Department of Electrical and Computer Engineering, University of Rochester, 186 Cumberland St, Rochester, New York 14623, USA}
\fntext[equal]{These authors contributed equally to this work.}
\cortext[cor]{Corresponding author.}

\begin{abstract}
Machine-learning approaches to weather forecasting often employ a monolithic architecture in which distinct physical mechanisms, such as advection, diffusive mixing, thermodynamic processes, and forcing, are represented implicitly within a single large neural network. This is particularly problematic for advection, where long-range transport typically requires expensive global interaction mechanisms or deep stacks of local convolutional layers. To address this limitation, we introduce a physics-inspired neural architecture that decomposes latent-state evolution into dedicated advection, diffusion, and reaction operators. Its central component is a Neural Semi-Lagrangian operator that performs trajectory-based transport via differentiable interpolation on the sphere, allowing the network to learn both a compressed set of latent modes to be transported and their characteristic trajectories.
The atmospheric state is projected into latent space and spatially coarsened to a processor grid, where advection, diffusion, and reaction operators jointly
evolve the representation. Diffusive mixing and unresolved dissipation are represented by depthwise-separable spatial mixing, while local source terms and vertical interactions are handled through pointwise channel interactions. We evaluate a reference implementation of the proposed architecture on global weather forecasting. Evaluated on ERA5 benchmarks, the reference model achieves competitive deterministic forecast skill, with particularly strong performance at short to medium lead times, while preserving improved spectral fidelity and forecast activity relative to several leading data-driven baselines.
\end{abstract}


\end{frontmatter}

\section{Introduction} \label{sec1}

Weather forecasting is critical infrastructure for global safety and economic resilience, informing daily decisions and mitigating the impacts of extreme events. For decades, operational forecasts have relied on numerical solvers for partial differential equations (PDEs). More recently, machine-learning approaches to weather prediction have advanced rapidly, offering competitive forecast skill at substantially reduced inference cost.

Data-driven models now span a broad range of architectures. Vision Transformer models, including Pangu-Weather~\citep{bi2023pangu}, FuXi~\citep{chen2023fuxi}, FengWu~\citep{chen2024fengwu}, and Baguan~\citep{niu2025utilizing}, use self-attention to represent long-range atmospheric dependencies. Graph-based approaches such as GraphCast~\citep{lam2023learning}, the ECMWF Artificial Intelligence Forecasting System (AIFS)~\citep{lang2024aifs}, Anemoi~\citep{wijnands2025anemoi}, and the work of Keisler~\citep{keisler2022graph} operate on multiresolution meshes that provide flexible treatment of spherical geometry. Fourier Neural Operators~\citep{bonev2023spherical} instead seek resolution-robust operator mappings. Recent work has also expanded toward probabilistic forecasting and generalization, including diffusion-based generative models~\citep{price2025probabilistic, li2024seeds} and large-scale foundation models such as Aurora~\citep{bodnar2024aurora} and Prithvi WxC~\citep{hhit2024prithvi}.

Other hybrid and physics-informed models incorporate physical structure more explicitly. NeuralGCM~\citep{kochkov2024neuralgcm} integrates a differentiable dynamical core with learned physical parameterizations. ClimODE~\citep{verma2024climode} frames atmospheric evolution as a continuous-time advection process governed by a Neural ODE, while DeepPrim~\citep{chendeepprim} embeds the primitive equations as architectural constraints. These approaches share our motivation of grounding neural forecasts in physical structure, although their computational requirements have so far limited their competitiveness with state-of-the-art data-driven systems for global weather forecasting at $0.25^\circ$.

Most directly relevant is the Advection--Diffusion--Reaction Network (ADRNet) of~\cite{zakariaei2024advection}, which showed that explicitly separating advection-dominated dynamics from local processes can improve over standard convolutional architectures. ADRNet, however, was developed for regular two-dimensional Cartesian grids and was not designed to address the spherical geometry, multi-level atmospheric states, long autoregressive rollouts, or computational scale of global weather forecasting. Earlier work~\cite{debezenac2018deep} similarly used the solution form of an advection--diffusion equation to motivate a neural model for sea-surface-temperature forecasting that estimates a motion field and applies differentiable Gaussian warping.

These methods demonstrate that explicit transport can be incorporated successfully into relatively small learned models. However, this idea has seen limited adoption in large-scale data-driven weather forecasting, where state-of-the-art models continue to rely predominantly on standard machine-learning mechanisms. As a result, atmospheric transport remains implicit in most modern global forecasting architectures.  Convolutions, message passing, and attention can represent long-range dependencies, but they do not directly encode the displacement of coherent atmospheric features along characteristic trajectories. Long-range transport must instead emerge from learned spatial interactions, potentially contributing to diffusive artifacts and the attenuation of fine-scale structure during autoregressive prediction \citep{zakariaei2024advection}. This smoothing can be further amplified by pointwise training objectives such as mean-squared error, which favor conditional-mean predictions rather than realistic localized extremes \citep{subich2025fixing}. At the same time, existing approaches that represent transport more explicitly have not directly addressed the combination of spherical geometry, multi-level atmospheric states, and computational scale required for global weather forecasting.

We investigate whether an explicit semi-Lagrangian transport operator can improve data-driven global forecasting as part of a learned latent architecture. We introduce a physics-inspired architecture that decomposes latent-state evolution into advection, diffusion, and reaction operators. Its central component is a Neural Semi-Lagrangian (NSL) operator that transports learned latent modes along learned trajectories using differentiable interpolation on the sphere. In contrast to classical semi-Lagrangian schemes \citep{robert1981stable}, the NSL operator learns both \emph{which} latent features to transport and \emph{along which} trajectories.

We extend the ADR principle to global weather prediction through learned advection on the sphere and latent transport-mode compression at operational grid scale. The advection operator is inspired by backward semi-Lagrangian methods, whose spatial reach is not directly constrained by a grid-scale Courant--Friedrichs--Lewy (CFL) condition~\citep{fletcher2019semi}. In the learned formulation considered here, stability depends on the predicted displacement fields and their composition across successive Advection--Diffusion--Reaction layers, which act as learned substeps.

This work makes two main contributions:
\begin{itemize}
\item \textbf{A Neural Semi-Lagrangian operator for learned atmospheric transport:} We introduce an explicit trajectory-based mechanism for transporting latent modes on the sphere. By separating advection from local spatial mixing and pointwise transformations, the architecture provides a geometric inductive bias for long-range transport while allowing diffusion-like and reaction-like processes to be learned by specialized operators.

\item \textbf{A global weather forecasting implementation of the proposed architecture:}
We instantiate the proposed approach in a global forecasting model operating on
$0.25^\circ$ inputs and outputs, with the latent advection--diffusion--reaction
processor evaluated at $0.5^\circ$. The resulting model achieves competitive
deterministic forecast skill while preserving improved spectral
fidelity and forecast activity during autoregressive rollouts.

\end{itemize}

The remainder of this paper is organized as follows. \Cref{sec:neural_architecture} details the theoretical foundation of our architecture, introducing the Advection--Diffusion--Reaction decomposition and the Neural Semi-Lagrangian operator. \Cref{sec:architecture} describes the integration of these components into the full physics-inspired neural architecture and forecast pipeline. \Cref{sec:experiments} presents comprehensive forecast verification results for the reference model, including spectral fidelity, prediction activity, and ablation studies. \Cref{sec:learned_operators} provides an analysis of the learned operators and internal latent representations. Finally, \Cref{sec:conclusion} summarizes the main findings and outlines directions for future work. Additional technical details on the proposed model, training procedures, and evaluation metrics are provided in the appendices.

\section{From Physical Principles to Data-Driven Numerical Methods} \label{sec:neural_architecture}

\begin{figure}[tb]
\centering
\includegraphics[width=0.48\linewidth]{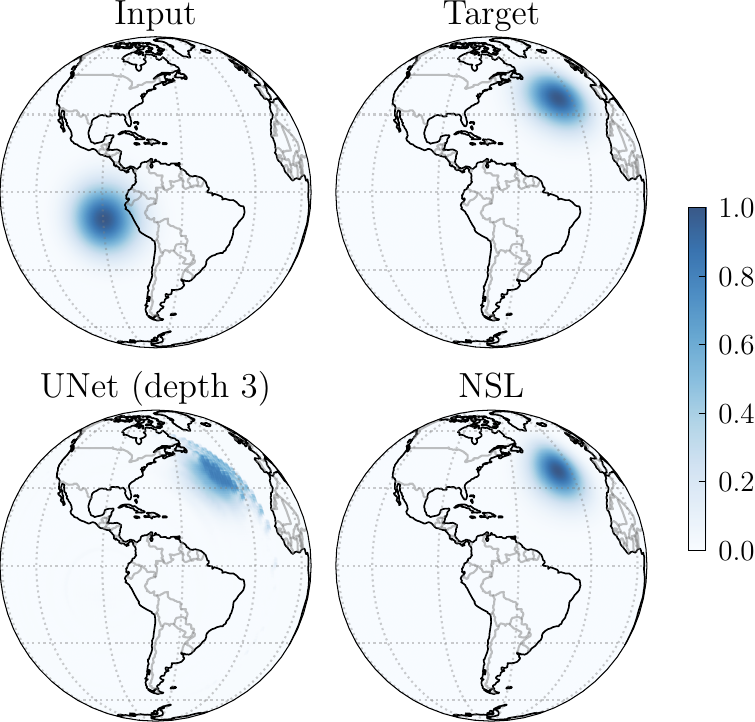}
\caption{Implicit vs.\ explicit advection on a $721\times1440$ latitude-longitude grid.
\textit{Top left:} Initial Gaussian tracer.
\textit{Top right:} Target after uniform-wind advection.
\textit{Bottom left:} UNet (depth 3, $\sim$45k parameters); the fixed receptive field limits transport range.
\textit{Bottom right:} NSL layer (20 parameters); accurately relocates the tracer while preserving its shape.}
\label{fig:advection_vs_unet}
\end{figure}

Most neural weather models treat forecasting as an image-to-image translation task, learning patterns from data without explicit physical structure. This perspective ignores the wavelike transport of atmospheric features along physical trajectories governed by wind and wave motions. 

We first isolate this issue in a simple transport problem before introducing the physically inspired architecture.

\subsection{Motivating Transport Experiment} \label{sec:motivational_example}

Advection, the transport of weather systems across the globe, is the dominant process in atmospheric dynamics. Standard CNNs struggle because convolution is a local operation with a fixed receptive field. Moving a feature across large distances, therefore, requires a very deep network, increasing the parameter count, introducing a diffusive bias, and raising the risk of unstable learning dynamics.

To demonstrate the structural limitations of standard convolutions for long-range transport, we compare U-Net models of varying depths against the NSL layer on a pure advection task. A single-channel Gaussian tracer on a native $0.25^\circ$ spherical grid is advected along a great circle trajectory over various angles $a_i$. To give the networks maximum opportunity to overfit, a separate model is trained to minimize the mean squared error for each advection angle. 

\begin{figure}[htb]
    \centering
    \includegraphics[width=0.75\linewidth]{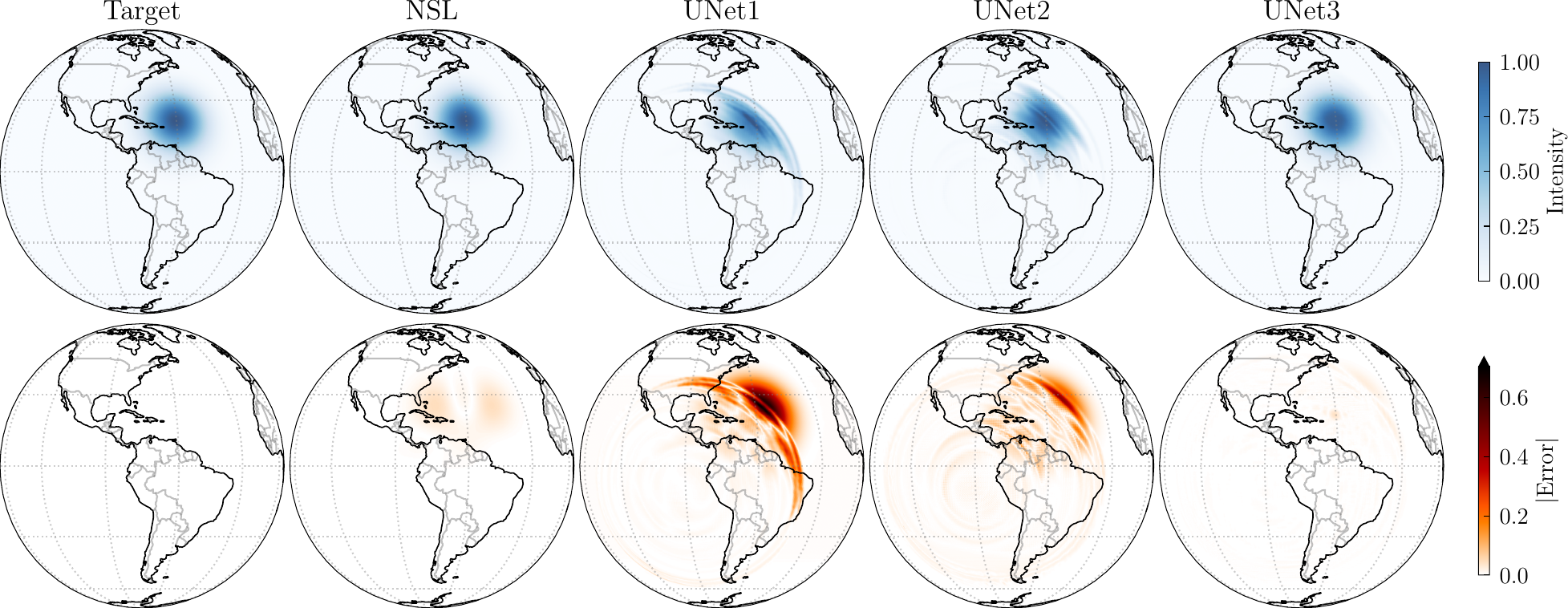} \\
    \vspace{1em}
    \includegraphics[width=0.75\linewidth]{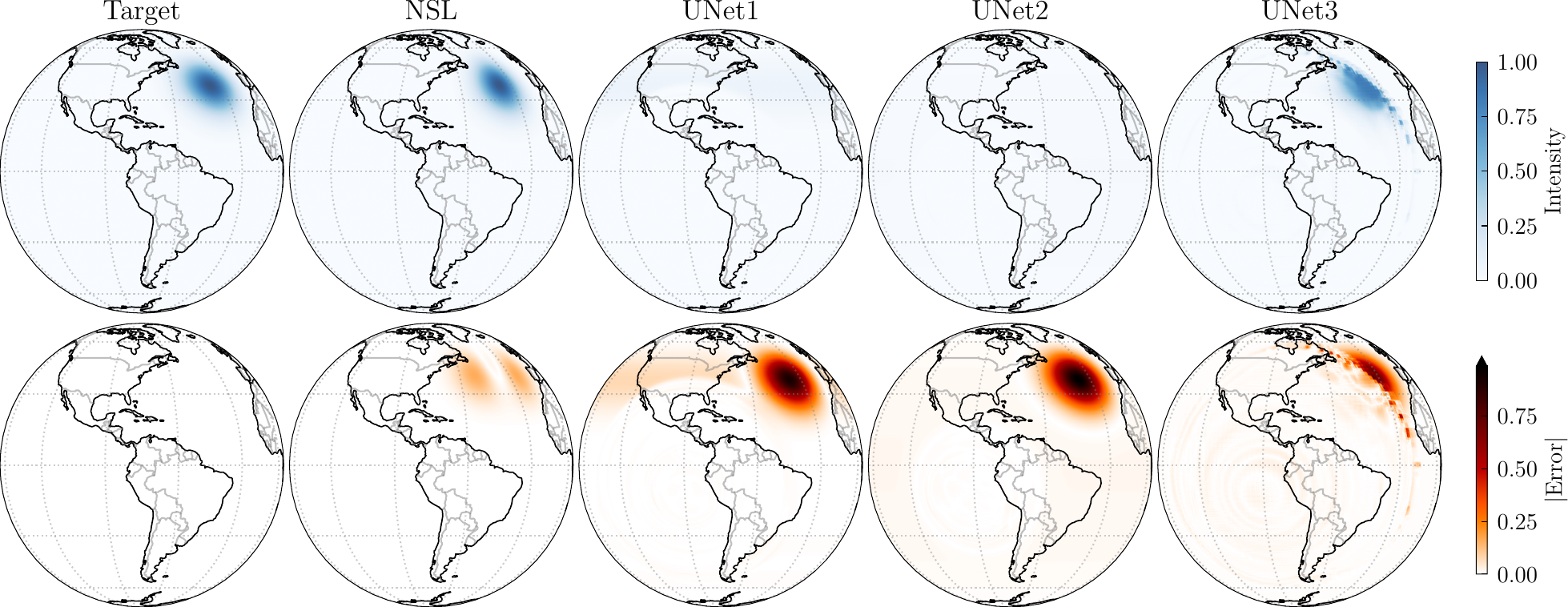}
    \caption{Rows 1 and 2: Model outputs and errors, respectively, for a $42^\circ$ rotation, inside the domain of dependence for the U-Net of depth 3, but not depth 1 or 2.  Rows 3 and 4: Outputs and errors, respectively, for a $68^\circ$ rotation, where the deeper U-Net struggles.}
    \label{fig:unet_adv_fields}
\end{figure}

While the U-Net transports information through local mixing and spatial coarsening, the NSL layer simply learns local velocities to interpolate from departure points. We parameterized the baseline U-Nets to roughly 45,000 parameters each, whereas the NSL layer uses only 20 learnable parameters (a single $3\times3$ convolution and biases). Qualitative results are shown in \Cref{fig:advection_vs_unet}, where the NSL approach is able to maintain the shape of the tracer. In \Cref{fig:unet_adv_fields}, we extend to larger displacement distances. Here, the U-Nets effectively memorize the advection only until the transport distance exceeds their fixed architectural receptive fields, at which point they fail catastrophically. In contrast, the 20-parameter NSL layer degrades slowly and gracefully across all angles, limited primarily by grid curvature rather than a hard domain of dependence. This confirms that explicitly modeling characteristic trajectories provides a highly parameter-efficient mechanism for long-range transport, freeing model capacity for local thermodynamics and forcing.


\subsection{Advection-Diffusion-Reaction Decomposition}

The proposed architecture is designed to mirror the structure that describes atmospheric evolution. At the coarsest level, the evolution of the state vector $\mathbf{q}\in \mathbb{R}^{C_\text{phys}}$ can be written as
\begin{align}
\label{eq:governing_equations}
\frac{\partial \mathbf{q}}{\partial t} + \underbrace{(\mathbf{u} \cdot \nabla) \mathbf{q}}_{\text{Advection}} = \underbrace{\mathcal{D}(\mathbf{q})}_{\text{Diffusion}} + \underbrace{\mathcal{R}(\mathbf{q})}_{\text{Reaction}},
\end{align}
where $\mathbf{u}$ is the velocity field. 

This decomposition naturally separates three physical processes: \textit{advection}, \textit{diffusion}, and \textit{reaction}. We use the terminology advection, diffusion, and reaction to describe the architectural roles of the three operators. Advection explicitly represents spatial transport; diffusion represents local spatial mixing and dissipation-like dynamics; and reaction encapsulates pointwise nonlinear channel interactions, including interactions between vertical levels. The latter two operators are not intended to correspond one-to-one with individual atmospheric parameterizations, but rather provide inductive biases appropriate to spatially local and column-local tendencies.

\begin{figure*}[tb]
\centering
\includegraphics[width=0.7\linewidth]{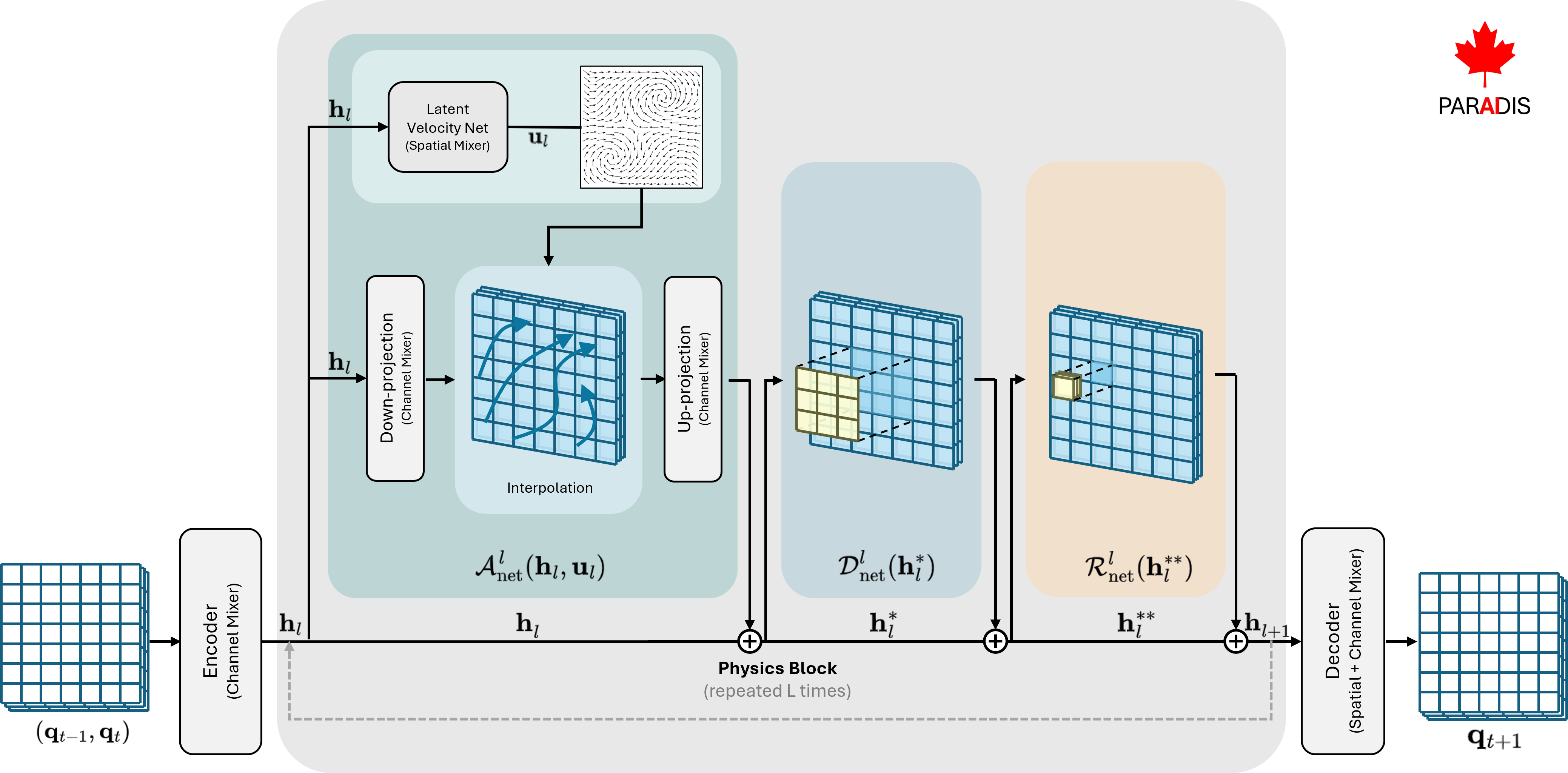}
\caption{Diagram of the Advection-Diffusion-Reaction neural architecture implementation.}
\label{fig:diagram}
\vspace{-1.5em}
\end{figure*}

Rather than learn this structure from data, we enforce it architecturally: the network is explicitly decomposed into operators for these processes. A diagram illustrating the proposed design is shown in \Cref{fig:diagram}. In the following subsections, we describe the main components of the model architecture.

\subsection{Modeling Advection with the Neural Semi-Lagrangian Operator} \label{sec:nsl_operator}

An NSL layer explicitly transports information along flow trajectories. A velocity network $\mathcal{V}_\text{net}$, implemented as a depthwise-separable convolution \citep{howard2017mobilenets}, estimates a set of transport fields from the latent state. These fields are learned from data to approximate the effective characteristics (curves) along which information propagates in the physical system.

The physical intuition behind this layer is that in a hyperbolic system of PDEs, such as those that approximate atmospheric motion, information is conserved along characteristic flow trajectories.  To find the state at $\mathbf{x}$ at time $t + \Delta t$, we ask: what information will arrive at $\mathbf{x}$? We answer this by tracing backward in time to the parcel's location at time $t$, as illustrated in \Cref{fig:compact_nsl_bicubic_clean}.

For each grid point $\mathbf{x}$, we compute its departure point
\begin{equation}
    \mathbf{x}_d = \mathbf{x} - \Delta t \cdot \mathbf{u}(\mathbf{x}).
\end{equation}
The advective contribution to the update is obtained by applying an interpolation operator $\mathcal{I}$ to the state vector $\mathbf{h}$ at the departure point:
\begin{equation}
    \mathbf{h}_{\mathcal{A}}(\mathbf{x},\, t+\Delta t) = \mathcal{I}\bigl[\mathbf{h}(\cdot,\, t)\bigr](\mathbf{x}_d),
\end{equation}
which corresponds to integrating the pure advection equation along the characteristic
from $\mathbf{x}_d$ to $\mathbf{x}$. This advected value is subsequently corrected by the diffusion and reaction operators. 

Unlike explicit Eulerian discretizations, semi-Lagrangian schemes are not subject to a grid-scale CFL restriction on the displacement over a time step. Instead, the regularity of the velocity field and the invertibility of the characteristic mapping become important considerations. This can be expressed through a Lipschitz-type condition that prevents neighboring characteristics from intersecting~\citep{smolar1992semilag,cossette2014mae}. In the learned setting considered here, the analogous behavior depends on the predicted displacement fields and on their composition across successive NSL layers.

We implement this operator on a latent set of features $\mathbf{h}\in\mathbb{R}^{C_\text{lat}}$, which we choose to have a large number of channels with respect to the physical space, i.e. $C_{\text {lat}} \gg C_\text{phys}$. Advecting the full latent state can become computationally expensive. We posit that atmospheric waves compress into a smaller set of ``transportable modes'' that move coherently. Unlike classical semi-Lagrangian schemes, the proposed architecture learns both which features to transport and their characteristic velocity fields.

We therefore project $\mathbf{h}$ down to a subspace $C_\text{adv}$ using pointwise convolutions, with $C_{\text{adv}} < C_{\text{lat}}$, apply the advection operator, and subsequently project back to the full latent space. This forces the network to discover a compressed representation of advected quantities, analogous to identifying the characteristic curves of the governing PDEs. The model learns end-to-end which combinations of atmospheric state are best represented through coherent spatial transport versus local transformation. The overall computational cost of the NSL operator is linear in the number of grid elements, $N_{\text{grid}}$, giving it lower asymptotic spatial complexity than dense global attention, whose cost scales as $O(N_{\text{grid}}^2)$.

\begin{figure}[tb]
\centering
\begin{tikzpicture}[
    scale=0.8,
    node distance=1cm,
    grid_point/.style={circle, fill=black!20, inner sep=1.2pt},
    interp_point/.style={circle, fill=deepblue, inner sep=1.4pt},
    arrival_node/.style={circle, fill=black, inner sep=2.2pt},
    departure_node/.style={circle, draw=black, line width=1.2pt, fill=white, inner sep=1.8pt},
    >=stealth,
]

    \draw[step=1cm, black!10, thin] (-0.5,-0.5) grid (5.5, 5.5);

    \draw[deepblue, thick, dashed, fill=deepblue, opacity=0.2] (0.5, 0.5) rectangle (4.5, 4.5);
    
    \foreach \x in {0,...,5}
        \foreach \y in {0,...,5}
        {
            \pgfmathparse{and(and(\x>=1,\x<=4),and(\y>=1,\y<=4)) ? 1 : 0}
            
            \ifnum\pgfmathresult=1
                \node[interp_point] at (\x,\y) {}; 
            \else
                \node[grid_point] at (\x,\y) {};  
            \fi
        }

    \coordinate (x_arr) at (5, 5); 
    \node[arrival_node, label={[black, font=\bfseries]above:$\mathbf{x} (t+\Delta t)$}] (A) at (x_arr) {};

    \coordinate (x_dep) at (2.3, 2.4); 
    \node[departure_node, label={[black, font=\bfseries, yshift=+10pt]below left:$\mathbf{x}_d (t)$}] (D) at (x_dep) {};

    \draw[->, thick, black, dashed] (A) -- node[midway, sloped, above, scale=0.9] {$-\mathbf{u} \Delta t$} (D);

    \node[deepblue, scale=0.85, anchor=south west] at (0.5, 4.5) {$4 \times 4 ~\text{Bicubic Stencil}$};

\end{tikzpicture}
\caption{The value at arrival point $\mathbf{x}$ ($\bullet$) is obtained by back-tracing to departure point $\mathbf{x}_d$ ($\circ$) and interpolating from the surrounding stencil.}
\label{fig:compact_nsl_bicubic_clean}
\end{figure}

\section{A Physics-Inspired Neural Architecture}
\label{sec:architecture}

Our physically inspired architecture structures atmospheric evolution as an explicit advection–diffusion–reaction system operating within a learned latent space. Rather than prescribing a rigid decomposition of the atmospheric state into known physical modes, the methodology provides specialized neural operators dedicated to transport, spatial mixing, and local interactions. This design allows the neural network to determine during training how information processing is distributed among these underlying physical processes.

\subsection{Physics-Inspired Decomposition in Latent Space}
\label{subsec:latent_decomposition}

Atmospheric flows arise from structures operating across distinct physical regimes. Synoptic-scale flows ($>1000$~km) are strongly influenced by rotational dynamics and Rossby waves, while smaller-scale motion includes processes such as gravity waves, turbulent mixing, and diabatic effects. These processes evolve and propagate on different characteristic scales. Rather than specifying this decomposition a priori, we project the physical state into a sufficiently rich latent space and allow the encoder and processor to learn an appropriate representation. Let
\begin{equation}
\mathbf{x}_t =
\left[
\mathbf{q}_t,
\mathbf{q}_{t-1},
\mathbf{s},
\mathbf{f}_t
\right],
\end{equation}
denote the assembled model input including atmospheric state variables $\mathbf{q}$ at times $t$ and $t-1$, geophysical static fields $\mathbf{s}$ and time-dependent forcings $\mathbf{f}_t$. The encoder first maps these fields to a
full-resolution latent representation
\begin{equation}
\mathbf{h}^{f}_0
=
\mathcal{E}(\mathbf{x}_t),
\qquad
\mathbf{h}^{f}_0
\in
\mathbb{R}^{C_{\mathrm{lat}}\times H\times W}.
\end{equation}
This representation is then spatially coarsened to the processor grid,
\begin{equation}
\mathbf{h}_0
=
\mathcal{C}_{\downarrow}(\mathbf{h}^{f}_0),
\qquad
\mathbf{h}_0
\in
\mathbb{R}^{C_{\mathrm{lat}}\times H_p\times W_p},
\end{equation}
where $H_p$ and $W_p$ denote the dimensions of the 
processor grid, and $\mathcal{C}_{\downarrow}$ denotes the spatial
coarsening operator.

Working in this latent space, we formulate the dynamics using the material derivative
\begin{equation}
    \frac{D}{Dt}
    \equiv
    \frac{\partial}{\partial t}
    +
    \mathbf{u}\cdot\nabla,
\end{equation}
which describes the rate of change following a moving fluid parcel. In the continuous-time limit, the latent dynamics are represented as
\begin{align}
    \mathbf{h}(\mathbf{x},t)
        &= \mathcal{E}(\mathbf{q}(\mathbf{x},t)), \\
    \frac{d\mathbf{x}}{dt}
        &= \mathcal{V}_{\mathrm{net}}(\mathbf{h}), \\
    \frac{D\mathbf{h}}{Dt}
        &= \mathcal{D}_{\mathrm{net}}(\mathbf{h})
        + \mathcal{R}_{\mathrm{net}}(\mathbf{h}), \\
    \mathbf{q}(\mathbf{x},t)
        &= \mathcal{G}(\mathbf{h}(\mathbf{x},t)),
    \label{eq:lagrangian_system}
\end{align}
where $\mathcal{V}_{\mathrm{net}}$ predicts latent transport velocities, $\mathcal{D}_{\mathrm{net}}$ represents spatial mixing and dissipation, $\mathcal{R}_{\mathrm{net}}$ represents local nonlinear interactions and source/sink processes, and $\mathcal{G}$ maps the evolved latent state back to physical variables. Advective transport is incorporated through the material derivative and is evaluated explicitly using the Neural Semi-Lagrangian (NSL) operator described in \Cref{sec:nsl_operator}.

To evolve the latent state from $t$ to $t+\Delta t$, we divide the forecast interval into $\nsubstep$ learned sub-steps, represented by successive processor layers. At each layer $l$, we apply a first-order Lie--Trotter operator splitting:
\begin{align}
    \mathbf{u}_l
        &= \mathcal{V}_{\mathrm{net}}^l(\mathbf{h}_l),
        \label{eq:unet} \\
    \mathbf{h}^*_l
        &= \mathbf{h}_l
        + \boldsymbol{\alpha}_l
        \left[
        \mathcal{A}_{\mathrm{net}}^l
        (\mathbf{h}_l,\mathbf{u}_l)
        - \mathbf{h}_l
        \right],
        \label{eq:step_adv} \\
    \mathbf{h}^{**}_l
        &= \mathbf{h}^*_l
        + \mathcal{D}_{\mathrm{net}}^l(\mathbf{h}^*_l),
        \label{eq:step_diff} \\
    \mathbf{h}_{l+1}
        &= \mathbf{h}^{**}_l
        + \mathcal{R}_{\mathrm{net}}^l(\mathbf{h}^{**}_l).
        \label{eq:step_react}
\end{align}
Here, $\mathcal{A}_{\mathrm{net}}^l$ is the NSL transport operator, which samples $\mathbf{h}_l$ at departure points computed from $\mathbf{u}_l$, and $\boldsymbol{\alpha}_l \in [0,1]^{C_{\mathrm{lat}}}$ is a vector of learned blending weights. The three stages specialize respectively in transport, spatial mixing, and local interactions. Their residual formulation allows each operator to learn a latent tendency over the corresponding sub-step and provides short gradient paths through the processor.

\subsection{Model Pipeline}
\label{subsec:forecast_pipeline}

 To perform a time step, the model $\operatorname{NN}$ receives the current state $\mathbf{q}_t$ and the preceding state $\mathbf{q}_{t-1}$, together with static geophysical fields and time-dependent forcings, and predicts the next state
\begin{equation}
\mathbf{q}_{t+1}
    =
    \operatorname{NN}
    (\mathbf{x}_t),
\end{equation}
where $\mathbf{x}_t$ denotes the assembled input defined in
\Cref{subsec:latent_decomposition}. The computation follows an
encoder--processor--decoder pipeline.

\paragraph{Encoder}
The assembled input $\mathbf{x}_t$ is projected into a latent representation
on the $0.25^\circ$ grid,
\begin{equation}
    \mathbf{h}^{f}_0 = \mathcal{E}(\mathbf{x}_t).
\end{equation}
Because atmospheric variables and pressure levels are represented along the
channel dimension, this projection also allows information to be exchanged
between physical variables and vertical levels.

\paragraph{Processor}
The encoded latent state is first mapped to a $0.5^\circ$ processor grid,
\begin{equation}
    \mathbf{h}_0 =
    \mathcal{C}_{\downarrow}(\mathbf{h}^{f}_0).
\end{equation}
The processor then evolves $\mathbf{h}_0$ through $\nsubstep$ successive
advection--diffusion--reaction layers,
\begin{equation}
    \mathbf{h}_{\nsubstep}
    =
    \mathcal{P}(\mathbf{h}_0)
    =
    \left(
    \mathcal{F}_{\nsubstep}\circ\cdots\circ\mathcal{F}_1
    \right)(\mathbf{h}_0).
\end{equation}
Here, each $\mathcal{F}_l$ corresponds to the split update in
\Cref{eq:unet,eq:step_adv,eq:step_diff,eq:step_react}. The advection stage
transports latent features along learned trajectories, the diffusion stage
exchanges information spatially, and the reaction stage performs local channel
interactions. All three operators are evaluated on the common $0.5^\circ$
processor grid.

\paragraph{Decoder}
After the final processor layer, the latent state is interpolated back to the
$0.25^\circ$ grid and combined with the full-resolution encoded representation,
\begin{equation}
    \mathbf{h}_{\mathrm{dec}}
    =
    \mathcal{I}_{\uparrow}
    \left(\mathbf{h}_{\nsubstep}\right)
    +
    \mathbf{h}^{f}_0.
\end{equation}
The resulting latent representation is then mapped to the physical forecast
variables,
\begin{equation}
    \mathbf{q}_{t+1}
    =
    \mathcal{G}(\mathbf{h}_{\mathrm{dec}}).
\end{equation}

\paragraph{Diffusion and reaction operators}
\label{subsec:diffusion_reaction}

The diffusion operator $\mathcal{D}_{\mathrm{net}}(\mathbf{h})$ models spatial mixing and unresolved dissipation. Depthwise-separable convolutions provide local spatial mixing within this latent representation.

The depthwise convolution acts as a learned spatial stencil independently within each latent channel, while the pointwise projection exchanges information between channels and can therefore represent cross-variable interactions.

The reaction operator $\mathcal{R}_{\mathrm{net}}(\mathbf{h})$ represents local nonlinear interactions and source/sink processes. It consists of two pointwise ($1\times1$) convolutions separated by a Swish (SiLU) activation~\citep{ramachandran2017searching}. Since atmospheric levels are stacked in the channel dimension, these transformations can model local vertical and column-wise coupling without introducing additional horizontal propagation.

Both the diffusion and reaction operators incorporate a low-rank bias mechanism described in \Cref{sec:low_rank_bias}. This provides spatially coherent static information, such as effects associated with topography and geography, without requiring a dense
$C_{\mathrm{lat}}\times H_p\times W_p$
bias tensor.

\paragraph{Spherical geometry and boundary conditions}
\label{subsec:geometry}

Global weather prediction on a latitude--longitude grid requires special treatment of the spherical topology, particularly near the poles. We use geocyclic padding~\citep{cheon2024karina} to maintain spatial continuity across these boundaries. Longitude is periodic, while features crossing either pole re-enter the domain shifted by $180^\circ$ in longitude, corresponding to their continuation across the sphere.

At the polar grid points, we additionally enforce scalar continuity by averaging features over the coincident longitudinal points independently for each channel. This prevents the accumulation of grid-scale artifacts at the convergence points of the latitude--longitude grid.

The NSL operator requires additional geometric treatment because trajectories are defined by displacements on the sphere. Departure locations are computed in a local rotated coordinate system centered at each grid point~\citep{mcdonald1989semi}, avoiding the coordinate singularity at the poles. The resulting departure points are then transformed back to latitude--longitude coordinates using spherical trigonometry. Further details of the interpolation and geometric transformations are provided in \Cref{appendix:padding}.

\section{Results} \label{sec:experiments}

The controlled transport experiment in~\ref{sec:motivational_example} demonstrated the architectural benefits of modeling transport explicitly. We now verify whether these benefits carry over to full atmospheric forecasting. To this end, we evaluate a reference implementation that integrates NSL advection with dedicated diffusion and reaction operators within an autoregressive forecasting architecture for global weather prediction at $0.25^\circ$ resolution. We refer to this reference implementation as PARADIS.

We evaluate the approach along complementary axes: computational scaling of the NSL operator, deterministic forecast skill of the ERA5-trained model, preservation of forecast activity and spectral structure, and tropical cyclone track and intensity errors.

\subsection{Performance of the NSL Operator}
We evaluate the computational scaling of the Neural Semi-Lagrangian transport operator independently of the learned projection and velocity-estimation components. The benchmark includes the spherical departure-point calculation, polar-continuity treatment, geocyclic boundary handling, and interpolation of the transported fields using PyTorch's \textit{grid sample} function. We benchmark both bilinear and bicubic interpolation for forward evaluation and for combined forward--backward evaluation.

For a grid of size $H\times W$, batch size $B$, and $C_{\mathrm{adv}}$ transported latent modes, the number of transported values is
\begin{equation}
N = B C_{\mathrm{adv}} H W.
\end{equation}
The scaling experiments use $B=1$ and $C_{\mathrm{adv}}=32$, with resolutions ranging from $32\times64$ to $1024\times2048$. The largest case therefore contains approximately $2.1\times10^6$ spatial grid points and $6.7\times10^7$ transported values, which corresponds to a resolution of approximately 20 km at the Equator.

The computational complexity of the NSL operator is linear in $N$. Departure points are computed independently at every grid location, and the interpolation uses a fixed-size local stencil. Consequently,
\begin{equation}
T_{\mathrm{NSL}} = O(B C_{\mathrm{adv}} H W),
\qquad
M_{\mathrm{NSL}} = O(B C_{\mathrm{adv}} H W),
\label{eq}
\end{equation}
where $T_{\mathrm{NSL}}$ and $M_{\mathrm{NSL}}$ denote computational and memory complexity, respectively. The interpolation order changes only the constant factor: bilinear interpolation uses a $2\times2$ stencil, whereas bicubic interpolation uses a $4\times4$ stencil. Importantly, the cost does not increase with the distance travelled by a characteristic; a long displacement requires the same number of interpolation samples as a short displacement.

The measured scaling agrees closely with this theoretical behavior. At the smallest resolutions, runtime is dominated by fixed execution overhead, whereas at larger resolutions it increases approximately proportionally to the number of transported values. Fitting a power law
\begin{equation}
t \propto N^\alpha
\end{equation}
over the three largest grid sizes gives $\alpha=0.96$ for bilinear interpolation and $\alpha=0.98$ for bicubic interpolation in the forward pass. When reverse-mode differentiation is included, the corresponding exponents are $0.93$ and $0.96$. These values are close to the ideal linear exponent $\alpha=1$.

Memory usage exhibits essentially linear scaling over the full resolution range. Each fourfold increase in the number of spatial grid points therefore produces
approximately a fourfold increase in peak memory, consistent with the $O(N)$ storage complexity in \Cref{eq}.

\begin{table}[htb]
\centering
\small
\caption{Scaling of the isolated Neural Semi-Lagrangian transport operator. Runtime exponents are fitted over the three largest resolutions. Runtime, throughput, and peak memory are reported for the largest $1024\times2048$ grid, corresponding to $6.71\times10^7$ transported values.}
\label{tab:nsl_scaling}
\begin{adjustbox}{max width=\linewidth}
\begin{tabular}{llcrrr}
\toprule
Interpolation &
Evaluation &
$\boldsymbol{\alpha}$ &
Median [ms] &
Throughput [Gvalues/s] &
Memory [MB] \\
\midrule
Bilinear & Forward            & 0.96 & 12.74 & 5.27 & 2840 \\
Bilinear & Forward + backward & 0.93 & 24.18 & 2.77 & 4384 \\
Bicubic  & Forward            & 0.98 & 19.45 & 3.45 & 2836 \\
Bicubic  & Forward + backward & 0.96 & 31.38 & 2.14 & 4384 \\
\bottomrule
\end{tabular}
\end{adjustbox}
\end{table}

At the largest tested resolution, the bicubic forward pass processes $6.71\times10^7$ transported values in $19.45$~ms, corresponding to a throughput of approximately $3.45$ billion values per second. Bilinear interpolation requires $12.74$~ms and reaches approximately $5.27$ billion values per second. Bicubic interpolation therefore increases the runtime constant, as expected from its larger interpolation stencil, but does not alter the linear scaling with resolution.

These results show that the NSL operator retains linear computational and
memory complexity as the grid is refined. This is particularly useful for
high-resolution global forecasting: long-range transport can be represented directly through characteristic trajectories without requiring the computational cost to grow with transport distance, and without introducing the quadratic spatial complexity associated with global pairwise interactions, typical of state-of-the-art models via attention mechanisms.

\subsection{Experiment Setup}  
We train the implemented model on ERA5 reanalysis at $0.25^\circ$ resolution over 1979--2019 and evaluate it on 2020 and 2022, using a 6-hour forecast time step. These evaluation years are selected to enable comparison with baseline forecasts available through the WeatherBench~2 framework~\cite{rasp_weatherbench_2024}. The model produces forecasts on the same $0.25^\circ$ grid, while
its latent ADR processor operates internally at $0.5^\circ$. Forecasts are produced autoregressively using a single recurrent model operator, while baseline models are evaluated using their standard released inference procedures. Unless otherwise stated, forecasts are initialized every 36 hours and verified against ERA5 at each lead time.

We compare PARADIS with ECMWF IFS HRES and leading data-driven forecasting systems, including GraphCast, Pangu-Weather, FuXi, GenCast, Aurora and Baguan. Evaluation follows the WeatherBench~\cite{rasp_weatherbench_2024} benchmarking protocol, with area-weighted metrics averaged over all initialization times and grid points. Baseline machine-learning models are evaluated using their publicly released configurations and pretrained weights, without retraining or restricting their input variables. This reflects their fully optimized operational usage and avoids introducing artificial handicaps. PARADIS is trained and evaluated using a comparable set of atmospheric state variables (see \Cref{tab:variable_inventory} for a detailed reference). We report results for upper-air variables including geopotential ($z$), temperature ($t$), wind components ($u,v$), and specific humidity ($q$), as well as surface variables including 2-metre temperature ($2t$), 10-metre winds, and mean sea-level pressure ($msl$). Spectral diagnostics are computed to assess scale-dependent forecast fidelity. The component-wise parameter distribution is shown in \Cref{tab:paramcount}. Detailed architectural and training hyperparameters are provided in the appendix.

\begin{table}[ht]
\centering

\caption{Parameter count for the different modules in the implemented model}
\label{tab:paramcount}
\small
\begin{adjustbox}{max width=\linewidth}
\begin{tabular}{@{}llrr@{}}
\toprule
\textbf{Module} & \textbf{Component} & \textbf{Parameters} & \textbf{\%} \\ \midrule
\multirow{3}{*}{Advection} 
    & Velocity net & 9{,}112{,}576 & 14.4\% \\
    & SL advection & 12{,}810{,}240 & 20.2\% \\
    & \textit{Subtotal} & \textit{21{,}922{,}816} & \textit{34.6\%} \\ \midrule
\multirow{5}{*}{Diffusion} 
    & Spatial mixer & 204{,}800 & 0.3\% \\
    & Channel mixer & 8{,}396{,}800 & 13.3\% \\
    & ChannelNorm & 16{,}384 & 0.0\% \\
    & Bias & 1{,}180{,}672 & 1.9\% \\
    & \textit{Subtotal} & \textit{9{,}798{,}656} & \textit{15.5\%} \\ \midrule
\multirow{4}{*}{Reaction} 
    & Channel mixer & 28{,}472{,}320 & 45.0\% \\
    & ChannelNorm & 18{,}432 & 0.0\% \\
    & Bias & 1{,}172{,}480 & 1.9\% \\
    & \textit{Subtotal} & \textit{29{,}663{,}232} & \textit{46.8\%} \\ \midrule
\multirow{4}{*}{Encoding} 
    & Input & 191{,}488 & 0.3\% \\
    & Output & 1{,}738{,}209 & 2.7\% \\
    & Static encoder & 14{,}314 & 0.0\% \\
    & \textit{Subtotal} & \textit{1{,}944{,}011} & \textit{3.1\%} \\ \midrule
\textbf{Total} & & \textbf{63{,}328{,}715} & \textbf{100.0\%} \\ \bottomrule
\end{tabular}

\end{adjustbox}
\end{table}

\paragraph{Training}
 \label{subsec:training_curriculum_main}

Training adopts a multi-stage curriculum designed to improve autoregressive forecast performance. All stages employ a reversed Huber loss defined as
\begin{equation}
\tilde{L}_\delta(e) =
\bigl(1-w(e)\bigr)\,\delta|e|
+ w(e)\,\frac{1}{2}e^2,\label{eq:revhuber}
\end{equation} where $w(e)$ is a sigmoid weighting function. 
This function penalizes quadratically if the error $e$ is large and linearly otherwise. This design is intended to strongly penalize catastrophic forecast failures while remaining tolerant of small errors. For more details on the behavior and choice of this loss function, see~\ref{subsec:training_objective}.

The training curriculum proceeds in two phases. We first pretrain the model for a 6h forecast for 260,000 steps using a 6-hour forecast horizon and a global batch size of 16 forecasts, allowing the model to learn one-step atmospheric predictions efficiently. Then, we fine-tune for multi-step forecasting through autoregressive rollouts, progressively increasing the forecast horizon to 12, 24, 36, 48, 60, and 72 hours, with 1,000 optimizer updates at each stage. Losses and gradient propagation are accumulated across the full rollout horizon window.

\paragraph{Computational cost} The model was trained on a distributed cluster of 16 NVIDIA H100 GPUs. Pre-training required approximately $0.30$ GPU-years, while autoregressive fine-tuning required an additional $0.05$ GPU-years, for a total training cost of approximately $0.35$ GPU-years. Once trained, PARADIS produces a 10-day forecast at $0.25^\circ$ resolution in approximately 35 seconds.

\subsection{Vertical Structure of Forecast Errors} 
\begin{figure*}[tb]
\centering
\includegraphics[width=\linewidth]{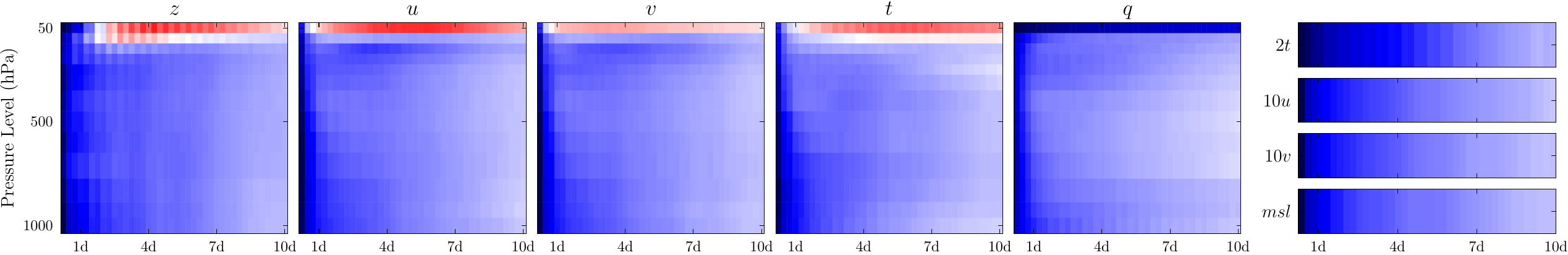}
\caption{Area-weighted RMSE differences (PARADIS minus HRES) across pressure levels as a function of lead time. Negative values indicate lower error for PARADIS (represented in blue). Skill gains for the dynamical variables are relatively uniform throughout most of the troposphere, with a localized degradation at the uppermost pressure levels.}
\label{fig:vertical_rmse}
\vspace{-1.5em}
\end{figure*}
To assess how forecast skill varies with altitude, \Cref{fig:vertical_rmse} presents the RMSE difference between PARADIS and IFS HRES as a function of pressure level and lead time. Negative values, shown in blue, indicate lower error for PARADIS, while positive values in red indicate an advantage for HRES.

For all considered dynamical variables, PARADIS exhibits lower RMSE throughout most of the troposphere. These improvements are broadly distributed across pressure levels and show relatively little dependence on altitude. They persist across much of the forecast range and are generally larger at shorter lead times. The main exception occurs at the uppermost levels, where a narrow region of degradation appears, with its magnitude varying by variable. Surface variables show a consistent advantage for the proposed approach: RMSE differences for 2-metre temperature, 10-metre winds and mean sea-level pressure favor our approach over HRES throughout the full forecast horizon. Overall, results indicate that the RMSE improvements of the proposed approach are not associated with a particular atmospheric layer, but instead reflect a broadly distributed reduction in mean-square error.

\subsection{Lead-Time Forecast Skill}
We compare PARADIS against established numerical and machine-learning baselines using a resolution-aware evaluation protocol. Because forecast availability differs across models, comparisons are reported for 2020 or 2022 depending on the available evaluation period.

\begin{figure}[htb]
    \centering

    \begin{subfigure}{\linewidth}
        \centering
        \includegraphics[width=\linewidth]{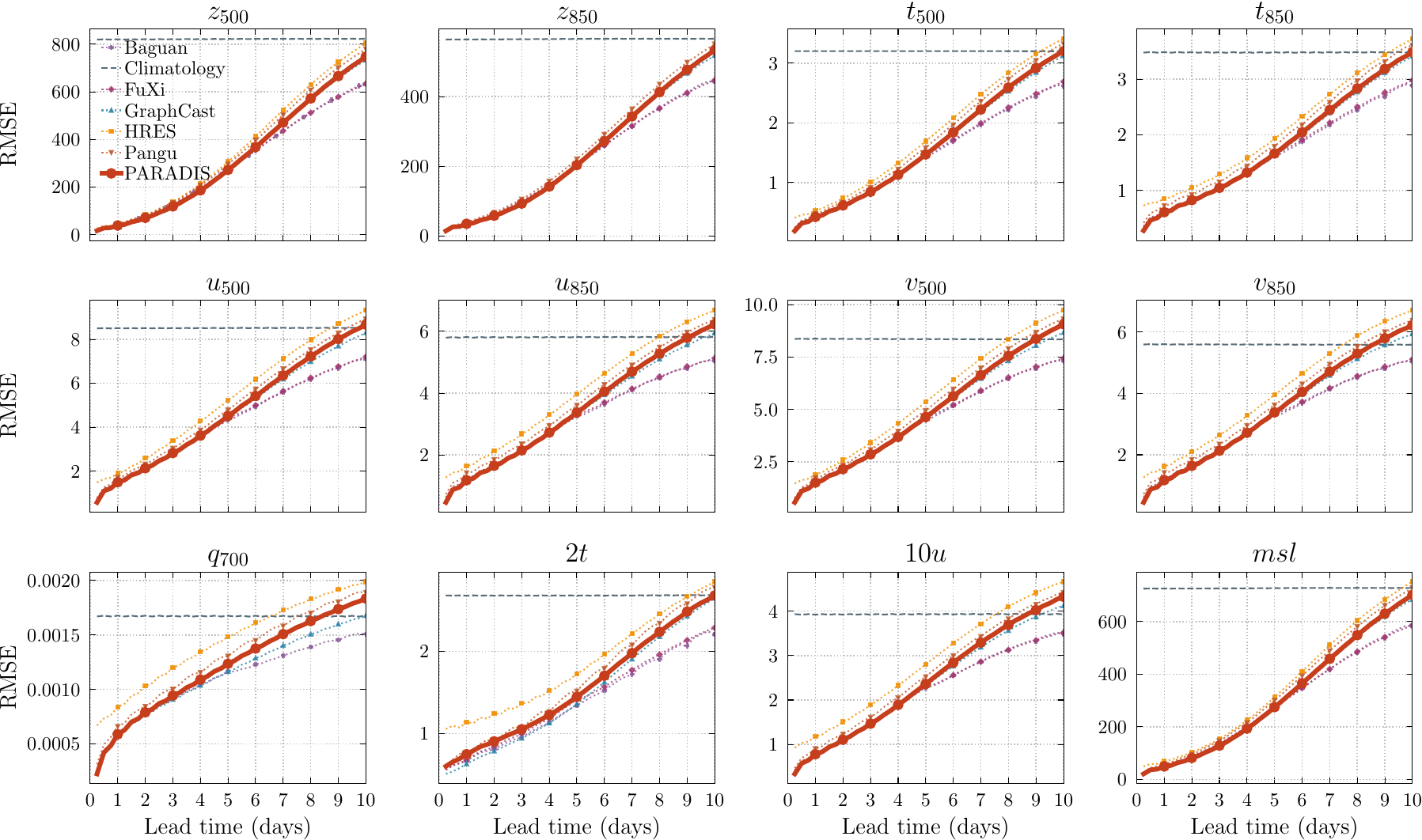}
        \caption{2020 Evaluation Period}
        \label{fig:paradis_others_rmse_2020}
    \end{subfigure}

    \vspace{0.5em}

    \begin{subfigure}{\linewidth}
        \centering
        \includegraphics[width=\linewidth]{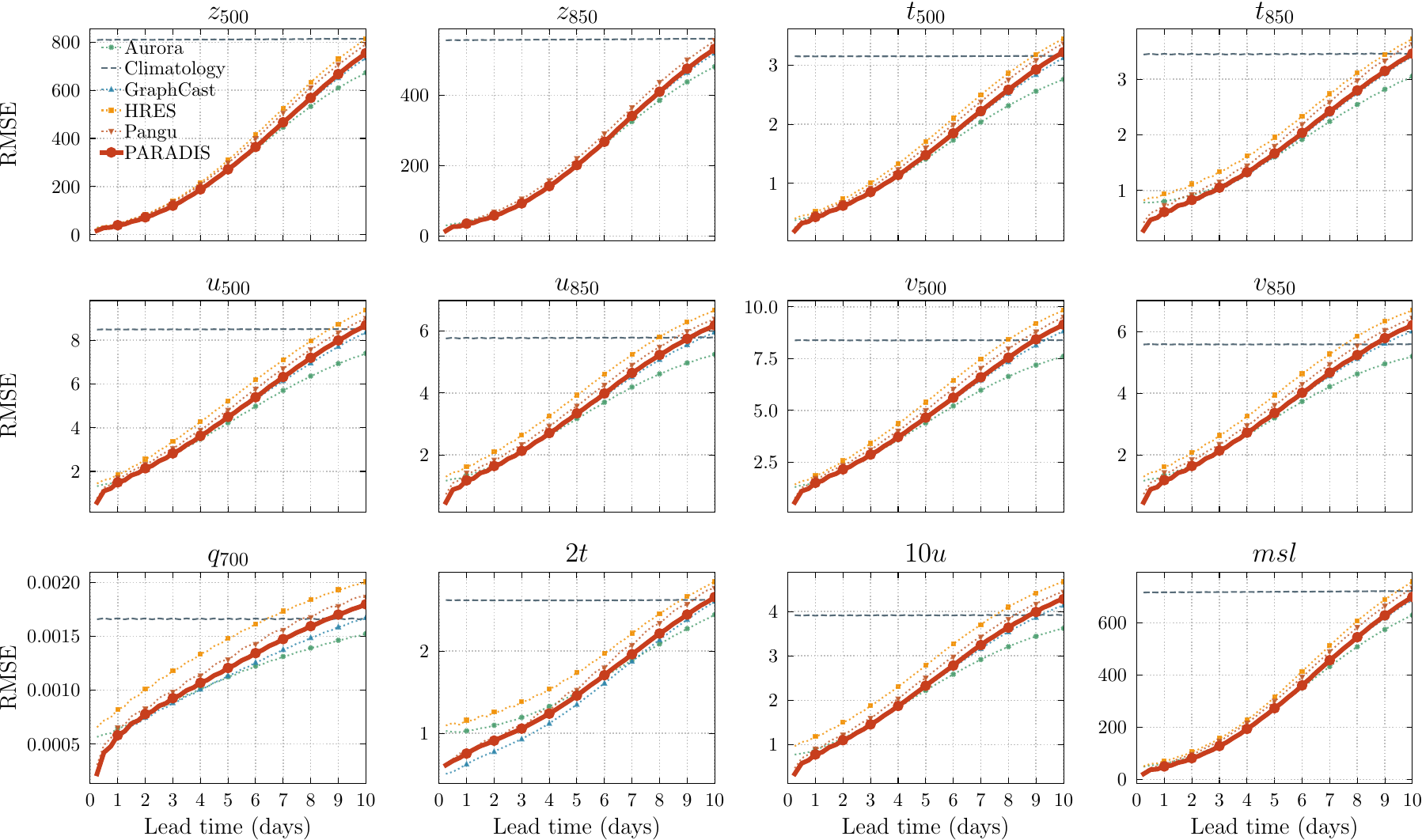}
        \caption{2022 Evaluation Period}
        \label{fig:paradis_others_rmse_2022}
    \end{subfigure}

    \caption{Comparison of forecast skill (RMSE; lower is better) between PARADIS and baseline models.}
    \label{fig:paradis_others_rmse}
\end{figure}

\Cref{fig:paradis_others_rmse} and \Cref{tab:paradis_rmse_z500} show the evolution of RMSE with forecast lead time. PARADIS is competitive with leading data-driven forecasting systems across the evaluated variables and performs particularly well at short to medium lead times. For $z_{500}$, PARADIS achieves the lowest RMSE among the evaluated models through five days in 2020 and remains among the strongest models over the corresponding range in 2022. Similar behavior is observed for several upper-air dynamical variables, including geopotential, temperature, and horizontal wind, where PARADIS generally improves upon ECMWF HRES and closely tracks the strongest machine-learning baselines.

At longer lead times, Baguan in 2020 and Aurora in 2022 obtain lower RMSE for several variables. As discussed in \Cref{sec:forecast_activity}, these models also exhibit stronger reductions in forecast activity at extended lead times, indicating greater smoothing. Consequently, RMSE alone does not characterize how well the forecasts preserve atmospheric variability, motivating the activity and spectral diagnostics considered below.

The behavior is broadly consistent across the two evaluation periods. Some variable-specific differences remain. In particular, PARADIS shows a modest degradation in 2-m temperature relative to GraphCast, while remaining competitive for the other surface variables. GraphCast uses a larger set of vertical pressure levels during training, which may contribute to differences in the representation of near-surface atmospheric structure.

\begin{figure}[htb]
    \centering

    \begin{subfigure}{\linewidth}
        \centering
        \includegraphics[width=\linewidth]{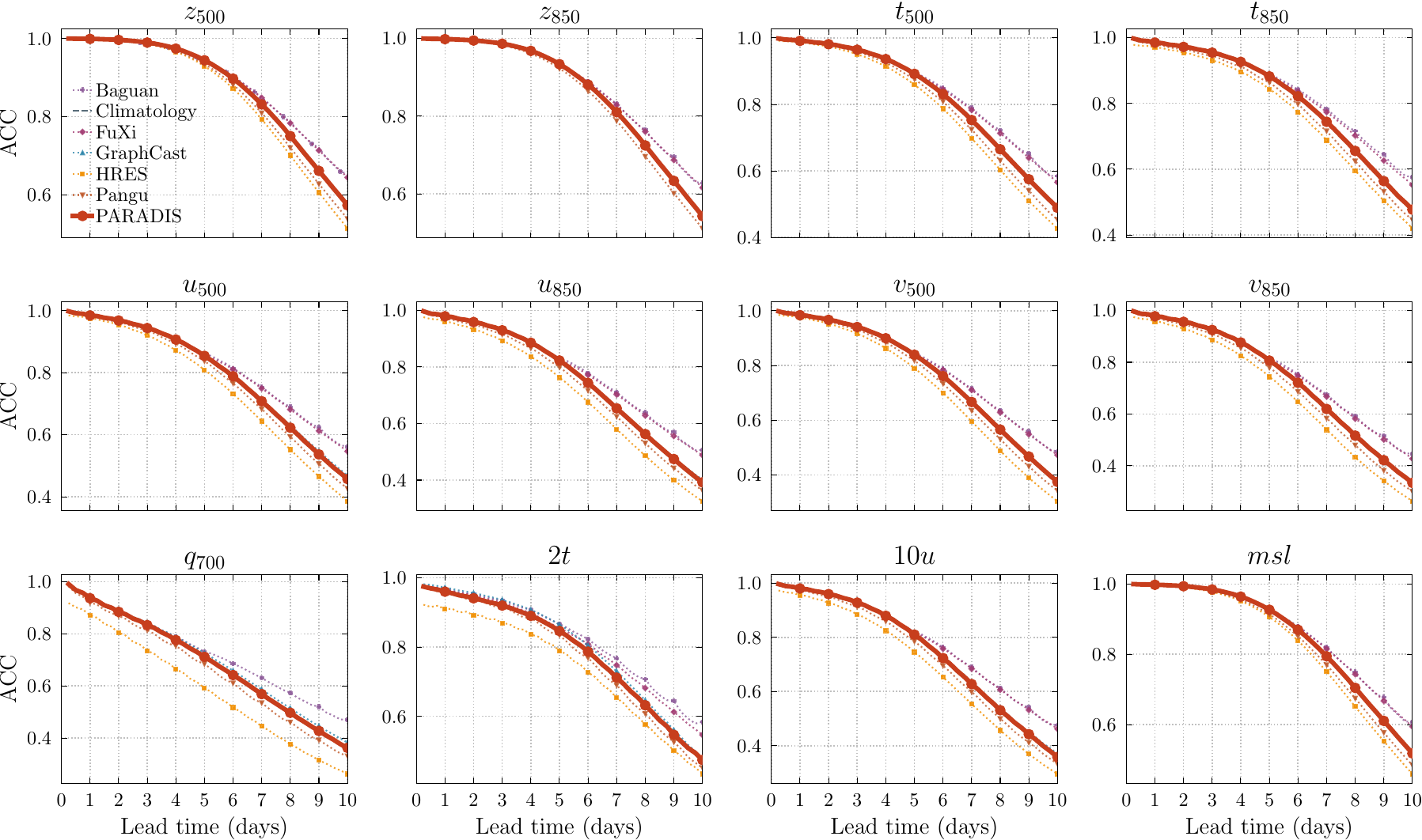}
        \caption{2020 Evaluation Period}
        \label{fig:paradis_others_acc_2020}
    \end{subfigure}

    \vspace{0.5em}

    \begin{subfigure}{\linewidth}
        \centering
        \includegraphics[width=\linewidth]{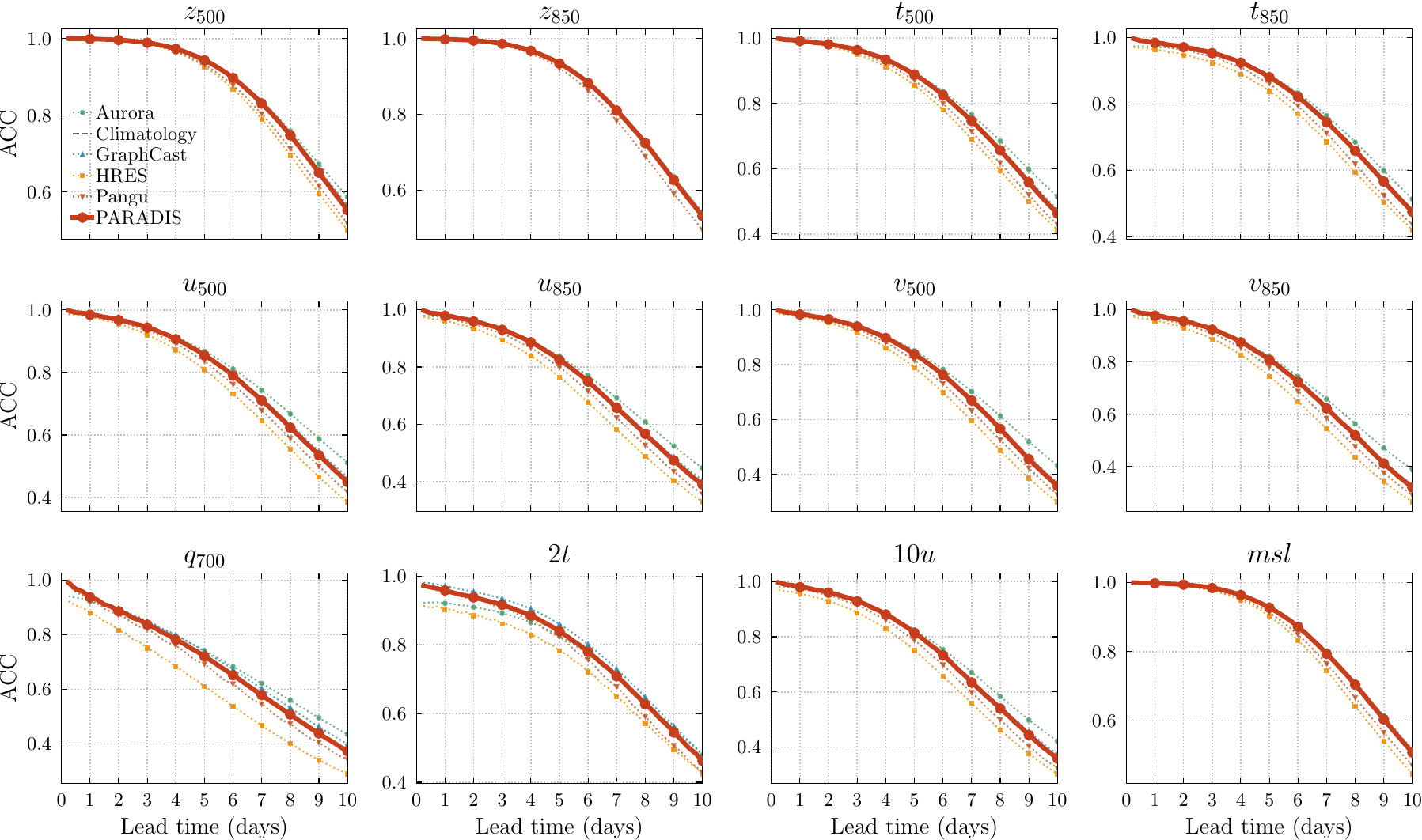}
        \caption{2022 Evaluation Period}
        \label{fig:paradis_others_acc_2022}
    \end{subfigure}

    \caption{Comparison of anomaly correlation coefficient (ACC; higher is better) between PARADIS and baseline models.}
    \label{fig:paradis_others_acc}
\end{figure}

The corresponding anomaly correlation coefficient (ACC) results are shown in \Cref{fig:paradis_others_acc}. PARADIS maintains competitive spatial correlation with ERA5 throughout the rollout horizon, with trends that closely follow the RMSE comparison. For several dynamical variables, PARADIS retains strong ACC at medium and long lead times even as pointwise RMSE increases, indicating that the large-scale spatial organization of the atmospheric state remains well represented.

\begin{table*}[htb]
\centering
\scriptsize
\caption{%
RMSE for $z_{500}$ at 24 h, 72 h, 120 h, and 240 h lead times for 2020 and 2022. Bold indicates the best-performing model, and underlining indicates the second-best.}
\label{tab:paradis_rmse_z500}
\begin{adjustbox}{max width=\linewidth}
\begin{tabular}{lcccccccc}
\toprule
 & \multicolumn{8}{c}{\textbf{$z_{500}$}} \\
\cmidrule(lr){2-9}
 & \multicolumn{4}{c}{2020} & \multicolumn{4}{c}{2022} \\
\cmidrule(lr){2-5}\cmidrule(lr){6-9}
Model & 24h & 72h & 120h & 240h & 24h & 72h & 120h & 240h \\
\midrule
PARADIS & \textbf{3.86e+01} & \textbf{1.19e+02} & \textbf{2.72e+02} & 7.48e+02 & \textbf{3.93e+01} & \textbf{1.20e+02} & \underline{2.71e+02} & 7.54e+02 \\
Aurora & - & - & - & - & 4.44e+01 & \underline{1.23e+02} & \textbf{2.66e+02} & \textbf{6.73e+02} \\
Baguan & 4.30e+01 & 1.29e+02 & \underline{2.75e+02} & \textbf{6.28e+02} & - & - & - & - \\
Climatology & 8.21e+02 & 8.22e+02 & 8.23e+02 & 8.23e+02 & 8.11e+02 & 8.11e+02 & 8.12e+02 & 8.15e+02 \\
IFS ens & 6.55e+01 & 1.97e+02 & 3.97e+02 & 8.72e+02 & 6.50e+01 & 1.97e+02 & 3.96e+02 & 8.79e+02 \\
FGN (member 1) & - & - & - & - & 5.13e+01 & 1.43e+02 & 3.17e+02 & 8.16e+02 \\
FuXi & 4.28e+01 & 1.28e+02 & 2.80e+02 & \underline{6.34e+02} & - & - & - & - \\
GenCast (member 1) & 5.58e+01 & 1.67e+02 & 3.55e+02 & 8.43e+02 & - & - & - & - \\
GraphCast & \underline{4.12e+01} & \underline{1.26e+02} & 2.76e+02 & 7.34e+02 & \underline{4.05e+01} & 1.24e+02 & 2.72e+02 & \underline{7.35e+02} \\
HRES & 4.86e+01 & 1.39e+02 & 3.08e+02 & 8.05e+02 & 4.82e+01 & 1.41e+02 & 3.10e+02 & 8.13e+02 \\
Pangu & 4.60e+01 & 1.36e+02 & 2.96e+02 & 7.81e+02 & 4.63e+01 & 1.36e+02 & 2.98e+02 & 7.92e+02 \\
\bottomrule
\end{tabular}
\end{adjustbox}
\end{table*}

\subsection{Forecast Activity and Smoothing}
\label{sec:forecast_activity}

\begin{figure}[htb]
    \centering

    \begin{subfigure}{\linewidth}
        \centering
        \includegraphics[width=\linewidth]{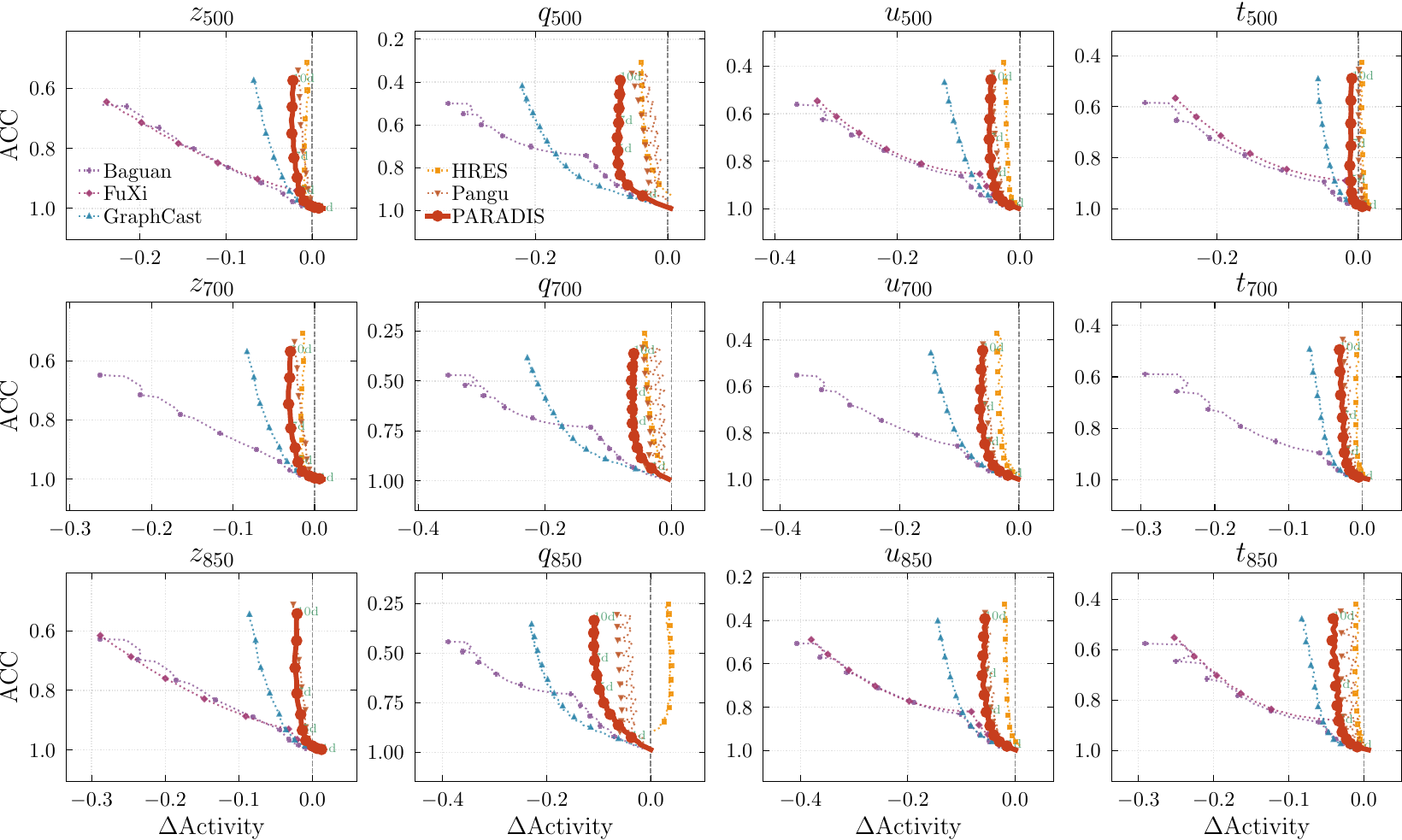}
        \caption{2020 Evaluation Period}
        \label{fig:paradis_others_activity_acc_2020}
    \end{subfigure}

    \vspace{0.5em}

    \begin{subfigure}{\linewidth}
        \centering
        \includegraphics[width=\linewidth]{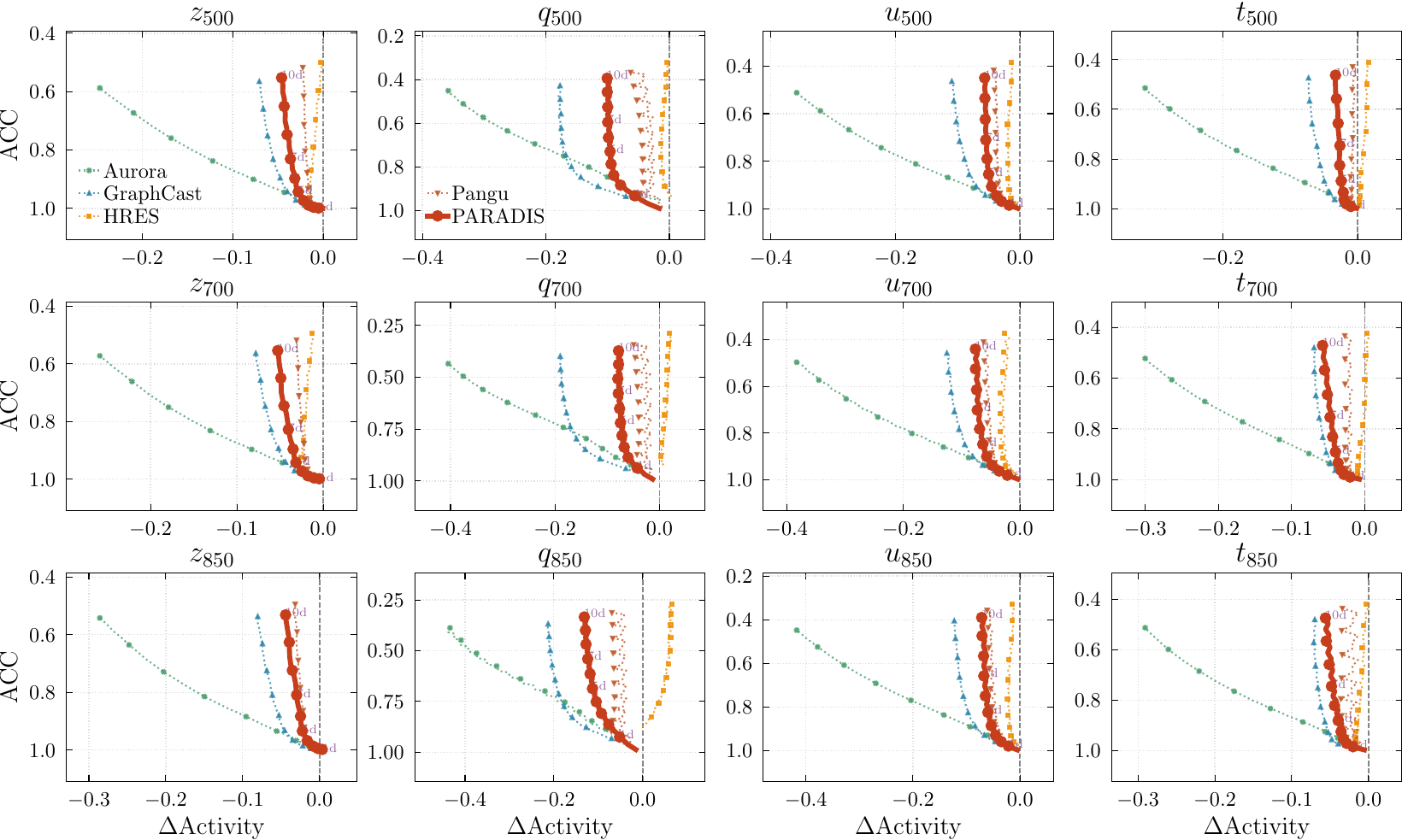}
        \caption{2022 Evaluation Period}
        \label{fig:paradis_others_activity_acc_2022}
    \end{subfigure}

    \caption{Relative forecast activity versus anomaly correlation coefficient for selected variables. Relative activity is computed with respect to ERA5, so negative values indicate smoothing, and values near zero indicate activity comparable to the verifying analysis.}
    \label{fig:activity}
\end{figure}

RMSE and ACC primarily measure pointwise accuracy and spatial agreement, but do not directly quantify whether a forecast preserves the amplitude of atmospheric variability. \Cref{fig:activity} therefore compares relative forecast activity, measured with respect to ERA5, as a function of ACC. Values near zero indicate variability comparable to the verifying analysis, while increasingly negative values indicate loss of forecast activity.

\begin{figure}[ht]
    \centering

    \begin{subfigure}{\linewidth}
        \centering
        \includegraphics[width=\linewidth]{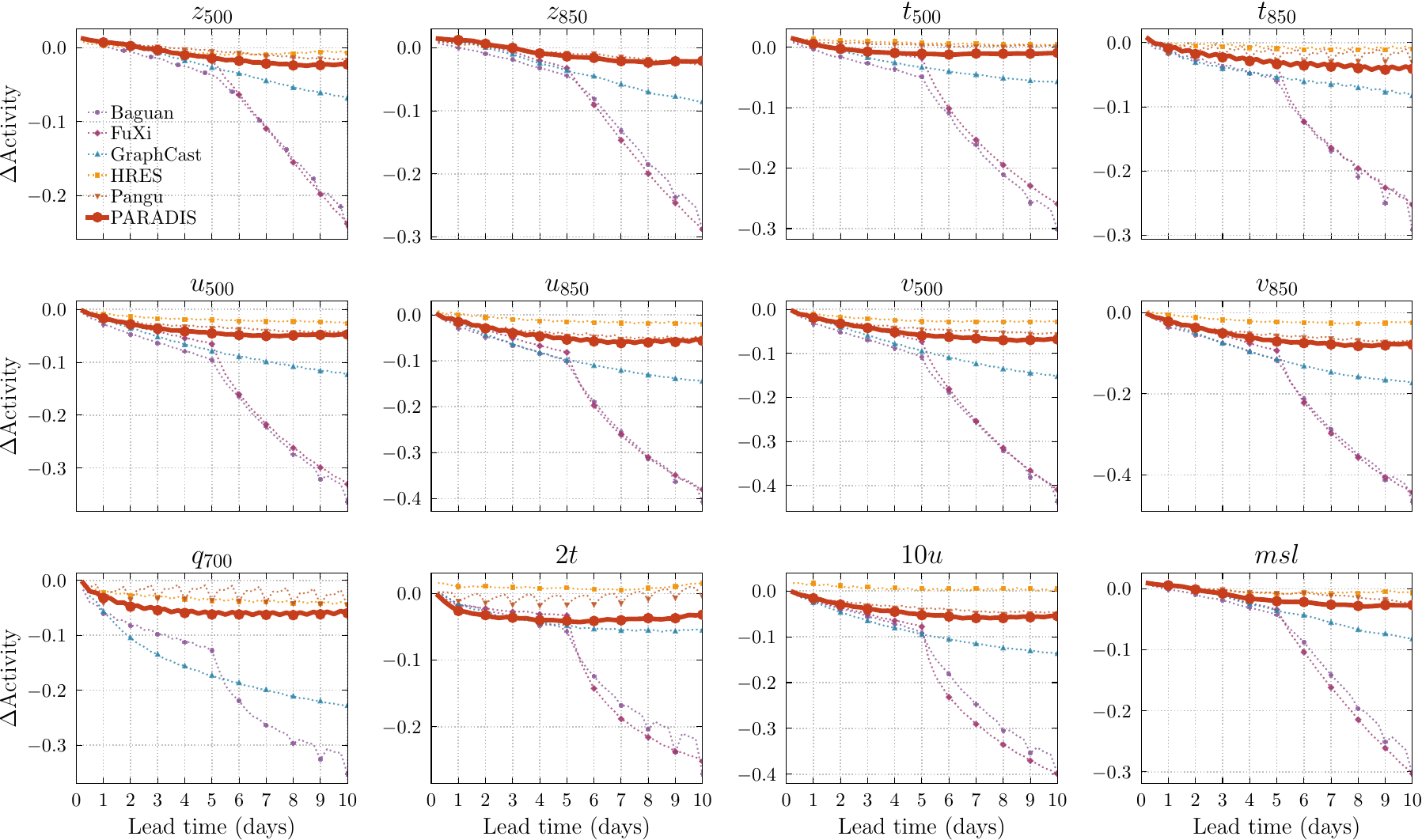}
        \caption{2020 Evaluation Period}
        \label{fig:paradis_others_activity_2020}
    \end{subfigure}

    \vspace{0.5em}

    \begin{subfigure}{\linewidth}
        \centering
        \includegraphics[width=\linewidth]{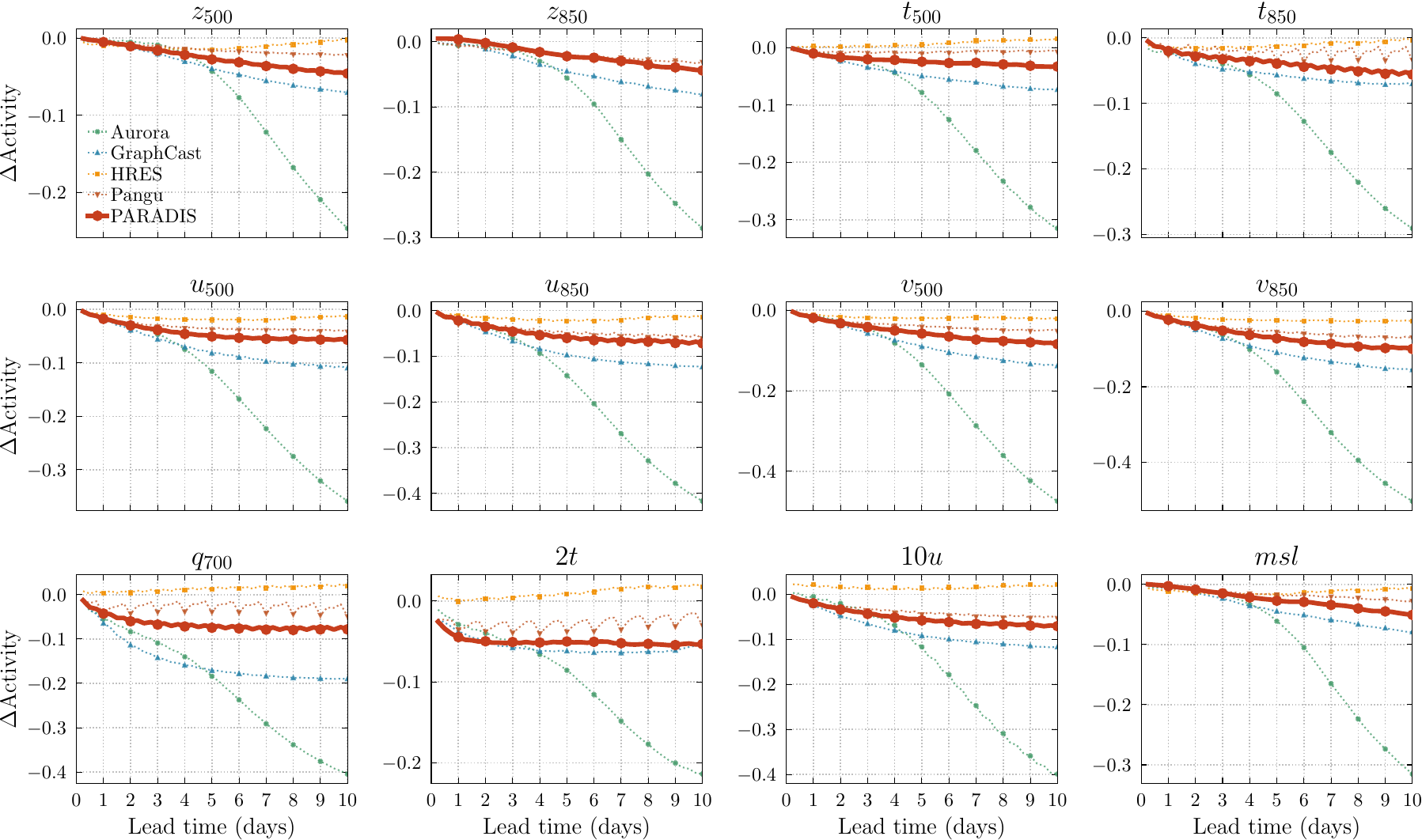}
        \caption{2022 Evaluation Period}
        \label{fig:paradis_others_activity_2022}
    \end{subfigure}

    \caption{Comparison of forecast activity (closer to zero is better) between PARADIS and baseline models.}
    \label{fig:paradis_others_activity}
\end{figure}

PARADIS preserves atmospheric activity consistently over long autoregressive rollouts. Baguan and FuXi exhibit substantial activity decay beyond approximately day five for several variables, indicating progressive smoothing of the forecast state. PARADIS remains substantially closer to ERA5 over the same period while maintaining competitive ACC. This distinction is particularly relevant at longer lead times, where models with reduced activity can obtain favorable pointwise RMSE while representing less of the variability present in the verifying analysis. The additional diagnostics in \Cref{fig:paradis_others_activity} provide a complementary view of this behavior.

This behavior is notable because PARADIS uses a single recurrent forecasting operator throughout the rollout. Pangu also maintains relatively stable activity for several variables, but constructs longer-range forecasts through hierarchical temporal aggregation using models trained at different forecast intervals.

\subsection{Spectral Fidelity and Phase Coherence}
\label{sec:spectral_fidelity}

Pointwise metrics such as RMSE and ACC do not fully characterize whether a forecast preserves atmospheric variability across spatial scales. In particular, a model can reduce RMSE through progressive smoothing while losing dynamically relevant small-scale structure~\cite{subich2025fixing}. We therefore complement the pointwise evaluation with spectral amplitude ratios and spectral coherence as functions of spherical wavenumber. The amplitude ratio measures the amount of variance retained relative to ERA5, with values near one indicating accurate spectral amplitude, while coherence measures the phase alignment of the forecast with the verifying analysis.

\begin{figure}[htb]
    \centering
    \includegraphics[width=0.48\linewidth]{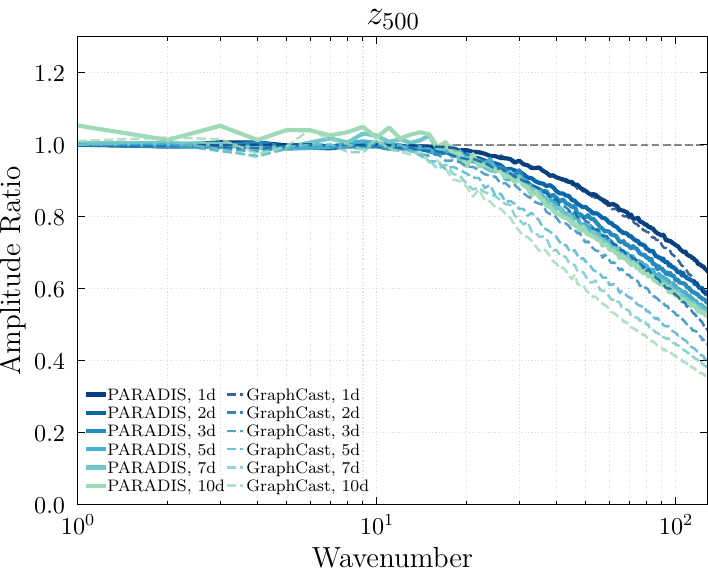}
    \includegraphics[width=0.48\linewidth]{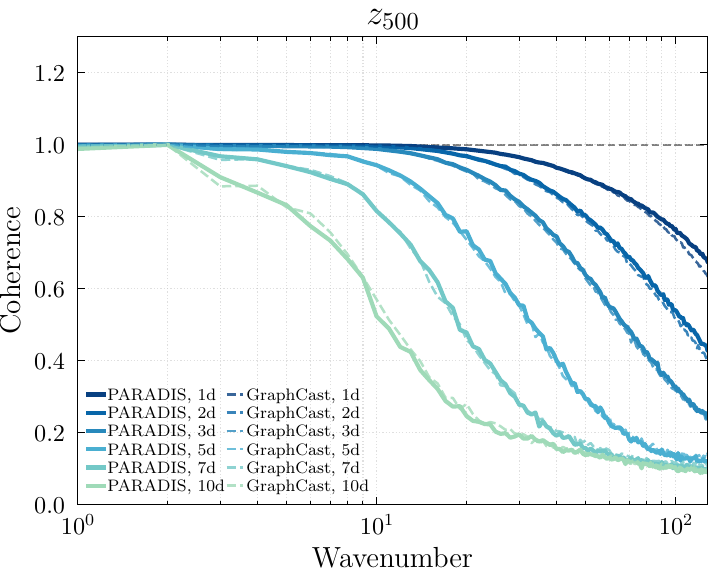}
    \caption{
    Spectral amplitude ratio (left) and coherence (right) for $z_{500}$ at
    selected lead times for PARADIS and GraphCast. An amplitude ratio near
    one indicates accurate retention of variance relative to ERA5, while
    higher coherence indicates better phase alignment with the verifying
    analysis. PARADIS retains more high-wavenumber variance and higher
    coherence, particularly at longer lead times.
    }
    \label{fig:spectra_graphcast}
\end{figure}

\Cref{fig:spectra_graphcast} compares PARADIS with GraphCast for $z_{500}$. At low wavenumbers, both models retain amplitudes close to ERA5 and maintain high coherence throughout the forecast range. Differences become increasingly apparent toward smaller spatial scales. GraphCast exhibits a stronger high-wavenumber roll-off in amplitude, particularly at longer lead times, whereas PARADIS retains substantially more variance. PARADIS also maintains higher coherence over these scales. The simultaneous improvement in amplitude retention and coherence is important: the additional high-wavenumber variance retained by PARADIS remains correlated with the verifying atmospheric state rather than appearing as unstructured small-scale variability.

\begin{figure}[htb]
    \centering
    \includegraphics[width=0.48\linewidth]{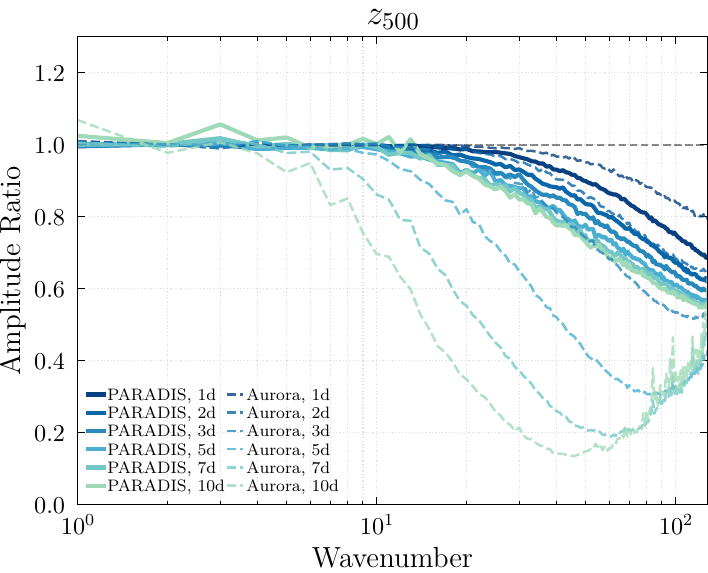}
    \includegraphics[width=0.48\linewidth]{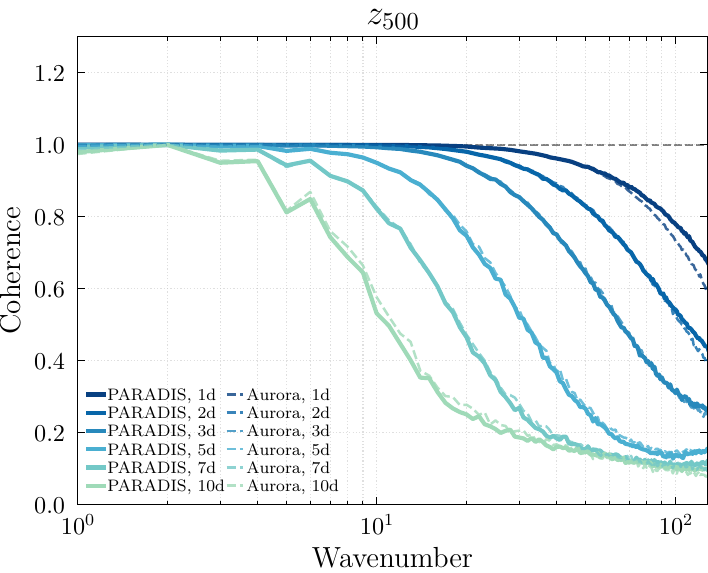}
    \caption{
    Spectral amplitude ratio (left) and coherence (right) for $z_{500}$ at
    selected lead times for PARADIS and Aurora. Aurora exhibits increasing
    spectral damping at medium and high wavenumbers with forecast lead time,
    while PARADIS retains more variance together with higher small-scale
    coherence.
    }
    \label{fig:spectra_aurora}
\end{figure}

A similar contrast is observed against Aurora in \Cref{fig:spectra_aurora}. Aurora develops pronounced spectral damping at medium and high wavenumbers as lead time increases. The effect becomes particularly strong at the longest lead times, where its amplitude ratio falls substantially below that of PARADIS over a broad range of scales. PARADIS preserves considerably more of the ERA5 spectral amplitude while also maintaining higher coherence toward the smaller scales. These spectral characteristics are consistent with the reduced forecast activity observed for Aurora at extended lead times and help explain why lower pointwise RMSE does not necessarily correspond to improved representation of atmospheric variability.

\begin{figure}[htb]
    \centering
    \includegraphics[width=0.48\linewidth]{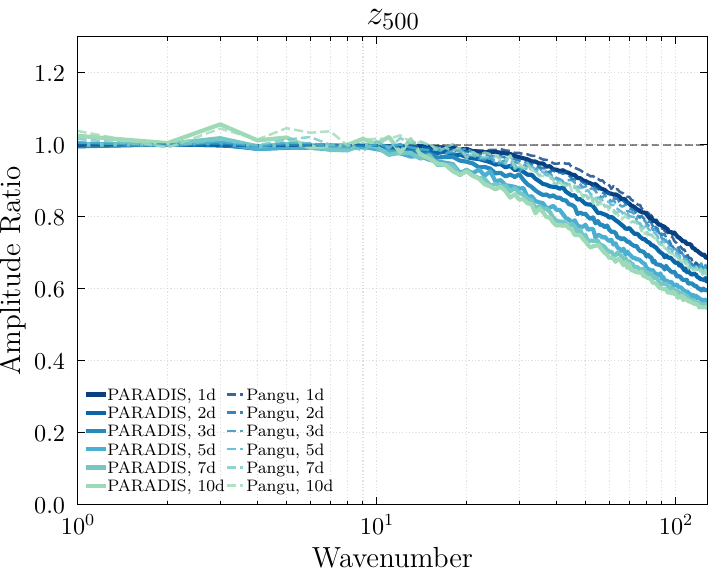}
    \includegraphics[width=0.48\linewidth]{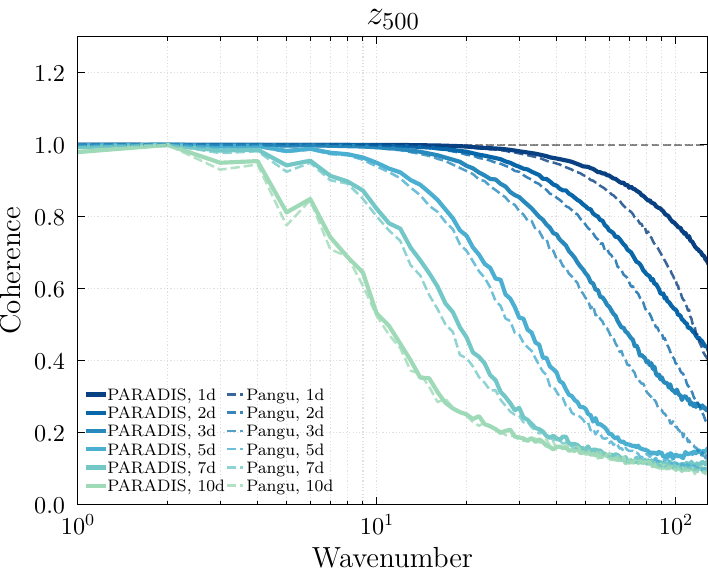}
    \caption{
    Spectral amplitude ratio (left) and coherence (right) for $z_{500}$ at
    selected lead times for PARADIS and Pangu. The two models show closer
    spectral behavior than in the GraphCast and Aurora comparisons, although
    PARADIS generally retains more high-wavenumber variance at longer lead
    times while maintaining higher coherence.
    }
    \label{fig:spectra_pangu}
\end{figure}

The comparison with Pangu in \Cref{fig:spectra_pangu} is closer. Both models retain substantially more small-scale variance than GraphCast or Aurora, and their amplitude spectra remain comparatively similar over much of the resolved range. PARADIS nevertheless tends to retain somewhat more high-wavenumber amplitude at longer lead times while maintaining  higher coherence. Pangu therefore represents a stronger spectral baseline, consistent with its relatively stable forecast activity discussed in \Cref{sec:forecast_activity}.

Taken together, the spectral results show that PARADIS preserves a broader range of spatial scales during autoregressive rollout. The improvement is visible both in spectral amplitude and in phase coherence, indicating that the retained small-scale energy corresponds to forecast structure that remains dynamically aligned with ERA5. This complements the activity analysis and supports the interpretation that the semi-Lagrangian latent dynamics can help mitigate the progressive smoothing commonly observed in deterministic neural weather forecasting systems.

\subsection{Cyclone Tracking and Intensity}\label{sec:cyclone_tracking_appendix}


\begin{figure}[htb]
\centering
\begin{subfigure}{0.48\linewidth}
  \centering
  \includegraphics[width=\linewidth]{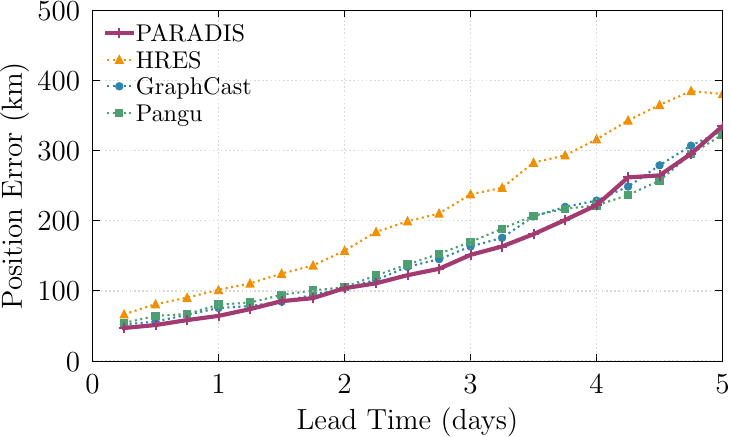}
  \caption{Position Error}
  \label{fig:cyclone_stats_a}
\end{subfigure}
\begin{subfigure}{0.48\linewidth}
  \centering
  \includegraphics[width=\linewidth]{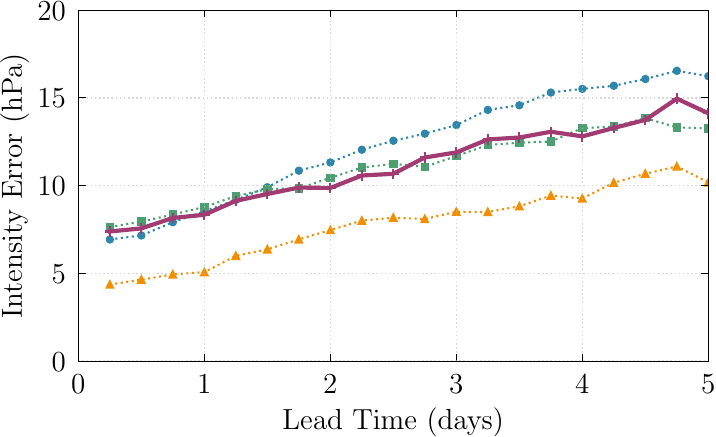}
  \caption{Intensity Error}
  \label{fig:cyclone_stats_b}
\end{subfigure}

\caption{Position error (left) and intensity error (right) for PARADIS and baseline models relative to IBTrACS over the 2020 evaluation period. Position error is measured by great-circle distance from the observed storm center, and intensity error by absolute central-pressure error; lower values indicate better performance.}
\label{fig:cyclone_stats}
\end{figure}

We evaluate the proposed implementation on tropical cyclone track and intensity errors, which provide stringent tests of coherent vortex propagation, dynamical consistency, and long-range transport. The evaluation is performed against the International Best Track Archive for Climate Stewardship (IBTrACS)~\citep{knapp2010international, gahtan2024ibtracs}, treating IBTrACS positions and central pressures as the verification reference. We aggregate over the 2020 evaluation year and restrict the analysis to storms for which all evaluated models provide complete forecast coverage, yielding 91 storms and ensuring a like-for-like comparison. Results are shown in Figure~\ref{fig:cyclone_stats}.

For each model forecast valid time, the predicted cyclone center is identified as the minimum mean sea-level pressure (MSLP; denoted $msl$) within a geographic search window centered on the last known storm position, seeded from IBTrACS. Model forecasts are verified at the same valid times as the IBTrACS records. When model output timestamps do not exactly coincide with IBTrACS times, we select the closest model valid time within a tolerance window of $\pm 3$ hours; forecasts without a match within this tolerance are excluded for that verification time. All great-circle distances are computed using the Haversine formula.

Track error is measured using the Direct Position Error (DPE), defined as the great-circle distance between the predicted cyclone center and the IBTrACS reference position:
\begin{equation}
\mathrm{DPE}(t) =
d_{\mathrm{gc}}\left(
(\phi_t^{\mathrm{pred}},\lambda_t^{\mathrm{pred}}),
(\phi_t^{\mathrm{ref}},\lambda_t^{\mathrm{ref}})
\right),
\end{equation}
where $d_{\mathrm{gc}}(\cdot,\cdot)$ denotes the Haversine distance.

Intensity error is computed as the absolute error in the minimum central pressure:
\begin{equation}
\mathrm{IE}(t) =
\left|
msl_{\min}^{\mathrm{pred}}(t) -
p_{\mathrm{WMO}}^{\mathrm{ref}}(t)
\right|,
\end{equation}
where $msl_{\min}^{\mathrm{pred}}(t)$ is the predicted minimum MSLP at the identified center and $p_{\mathrm{WMO}}^{\mathrm{ref}}(t)$ is the WMO-reported central pressure from IBTrACS. Both DPE and intensity error are aggregated as means across storms at each lead time.

\Cref{fig:cyclone_stats_a} shows that PARADIS produces strong tropical-cyclone track forecasts over the five-day rollout. Its position error remains among the lowest of the evaluated models throughout most of the forecast horizon. PARADIS generally improves upon GraphCast and HRES and remains close to Pangu, with the differences between PARADIS and Pangu becoming small toward the end of the forecast window. These results indicate that the learned semi-Lagrangian dynamics are able to preserve coherent storm translation over multi-day autoregressive forecasts.

The intensity comparison in \Cref{fig:cyclone_stats_b} is more mixed. PARADIS consistently produces lower intensity error than GraphCast and remains broadly comparable to Pangu over much of the forecast horizon, but HRES achieves substantially lower intensity error throughout the five-day period. The difference between the track and intensity results suggests that PARADIS captures cyclone propagation more successfully than the evolution of the localized pressure minimum associated with storm intensity.

Overall, the cyclone evaluation provides additional evidence that the proposed architecture preserves coherent advective structures during autoregressive forecasting. The particularly strong track performance is consistent with the spectral and forecast-activity diagnostics, which indicate improved preservation of dynamically organized structure. Cyclone intensity remains more challenging, however, suggesting that accurate transport alone is insufficient to fully represent the localized processes controlling the evolution of extreme central pressure. Improving the representation of these localized extremes remains an important direction for future work.

\subsection{Ablation Studies}\label{sec:ablation}

To assess the contribution of the main components and training choices in PARADIS, we perform a series of controlled ablation experiments. We study the effect of the loss function and isolate the roles of the advection, diffusion, and reaction components of the ADR formulation. Due to the computational cost of training the reference $0.25^\circ$ forecast configuration, all ablations are performed at $1^\circ$ resolution under a common experimental setup.

\subsubsection{Loss Function}

\begin{figure}[htb]
\centering
\includegraphics[width=\linewidth]{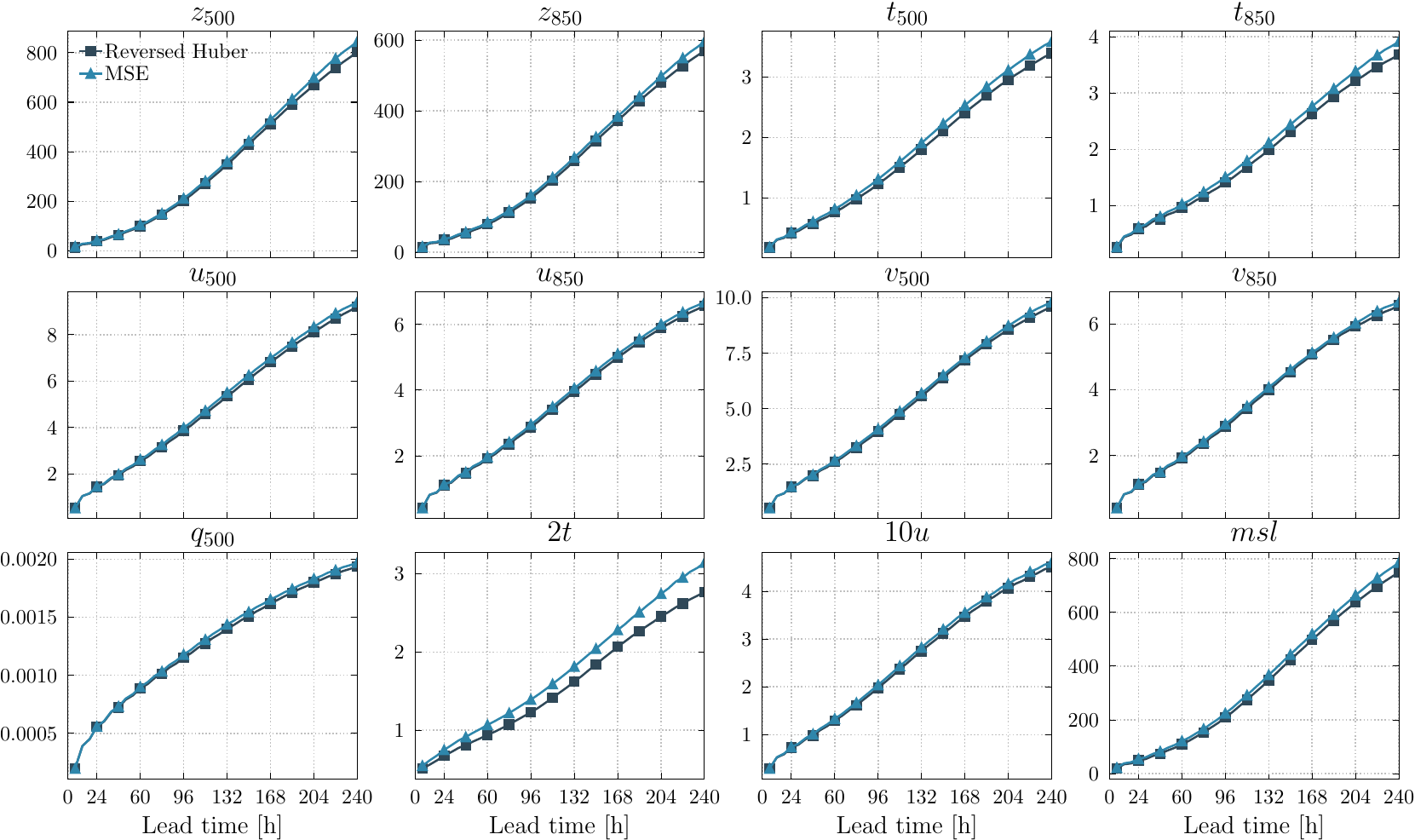}
\caption{RMSE as a function of forecast lead time for models trained with reversed Huber and MSE losses. Results are shown for representative upper-air and surface variables using the $1^\circ$ ablation configuration. Reversed Huber generally matches or improves upon MSE, with the largest difference observed for 2-m temperature at medium and long lead times.}
\label{fig:ablation_loss}
\end{figure}

We first evaluate the effect of the training objective by comparing the reversed Huber loss considered in this work with a standard mean squared error (MSE) objective. Both models use the same architecture and training configuration, so differences in forecast skill can be attributed to the loss function.~\Cref{fig:ablation_loss} shows the resulting RMSE over a ten-day rollout for representative upper-air and surface variables.

The reversed Huber objective consistently matches or improves upon MSE across the evaluated variables. For most upper-air fields, including geopotential, temperature, and horizontal wind, the differences are small at short lead times but become more apparent as forecast errors accumulate during the rollout. The clearest improvement occurs for 2-m temperature, where the reversed Huber model exhibits substantially slower error growth beyond approximately four days. Smaller but persistent improvements are also visible for several dynamical variables and mean sea-level pressure at medium and long lead times. Specific humidity shows little sensitivity to the choice of objective.

These results motivate the use of the reversed Huber loss in the full configuration of the model. Although both objectives produce similar short-range skill, reversed Huber provides equal or lower RMSE across the evaluated forecast horizon and yields a clear improvement for some variables without an evident degradation in others. Its advantage becomes most relevant during extended autoregressive rollouts, which are the primary setting considered in this work.

\subsubsection{ADR Components}

We next assess the contribution of the advection, diffusion, and reaction operators by removing each component independently while retaining the remaining two. All ablation experiments use the $1^\circ$ PARADIS configuration because of the computational cost of retraining multiple variants at $0.25^\circ$. Although absolute forecast errors differ from those of the reference $0.25^\circ$ model, this setting provides a controlled comparison of the contribution of each operator.

The parameter counts of the resulting configurations are reported in \Cref{tab:ablations}. Removing a component reduces both model capacity and computational cost, although the magnitude of the reduction differs substantially between operators.

\begin{table}[htb]
\centering
\caption{Parameter count for the $1^\circ$ ADR ablation configurations.}
\label{tab:ablations}
\begin{adjustbox}{max width=\linewidth}
\begin{tabular}{lc}
\toprule
\textbf{Configuration} & \textbf{Parameter count}  \\
\midrule
Full PARADIS ($A+D+R$) & 61.5M  \\
No Advection ($D+R$)   & 40.1M  \\
No Diffusion ($A+R$)   & 52.2M  \\
No Reaction ($A+D$)    & 32.4M  \\
\bottomrule
\end{tabular}
\end{adjustbox}
\end{table}

\begin{figure}[ht]
\centering
\includegraphics[width=\linewidth]{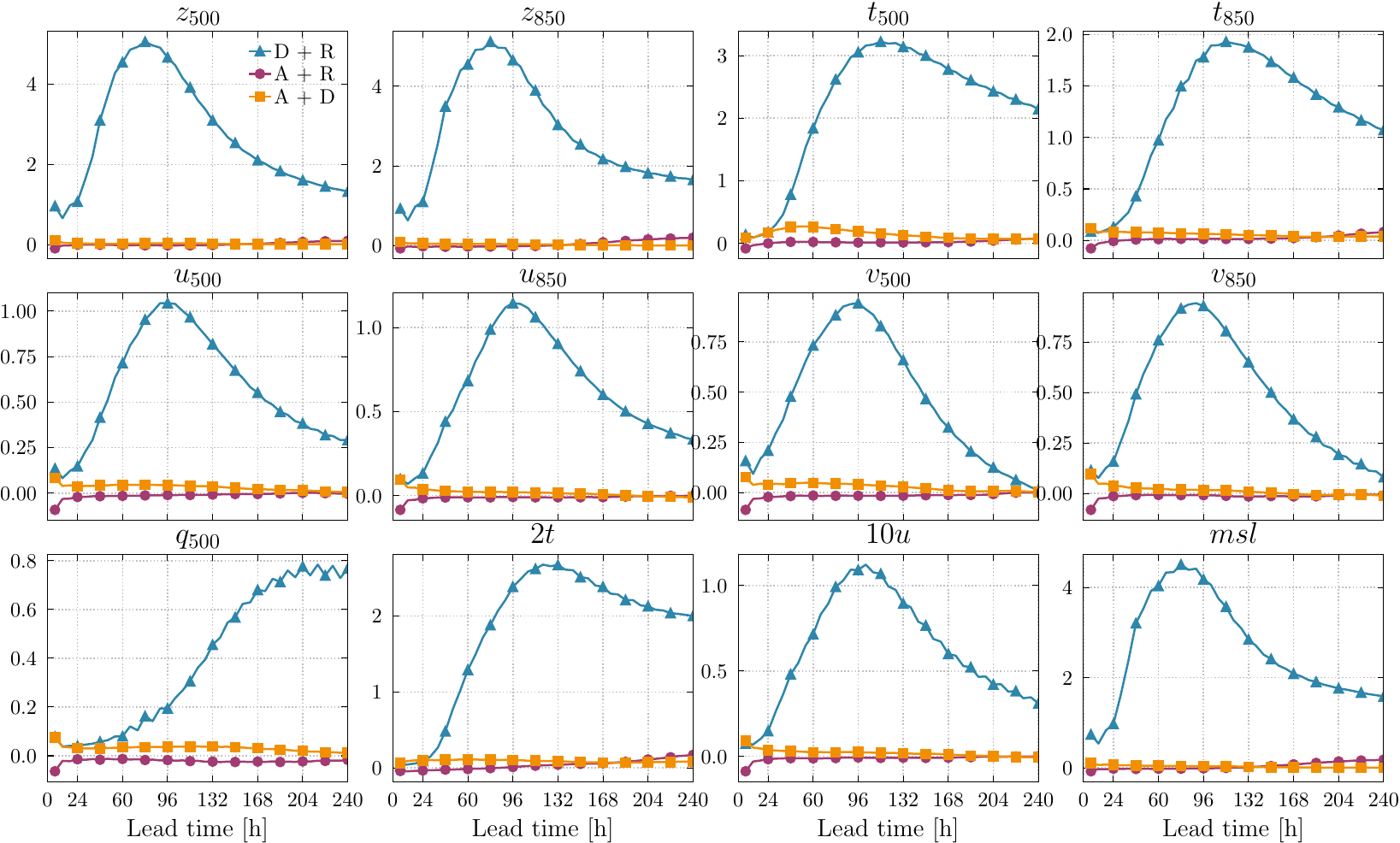}
\caption{RMSE relative to the full PARADIS model for ablations of advection ($D+R$), diffusion ($A+R$), and reaction ($A+D$). Lower values indicate performance closer to or better than the complete model. Removing advection causes by far the largest increase in forecast error across variables and lead times, whereas removing diffusion or reaction results in comparatively small changes.}
\label{fig:ablation_rmse}
\end{figure}

\Cref{fig:ablation_rmse} shows the change in RMSE relative to the complete $A+D+R$ model. The effect of removing advection is markedly larger than that of removing either diffusion or reaction. The $D+R$ configuration rapidly accumulates additional error across all evaluated upper-air and surface variables, with the degradation becoming particularly pronounced at medium lead times. This behavior is visible for geopotential, temperature, horizontal wind, 2-m temperature, and mean sea-level pressure, and persists through the remainder of the ten-day rollout.

In comparison, removing diffusion or reaction produces much smaller changes in RMSE. The $A+R$ configuration generally remains close to the complete model, while the $A+D$ configuration incurs a modest degradation for several variables. The difference in sensitivity is especially notable given that the reaction block contains a substantial fraction of the model parameters. Forecast skill therefore cannot be explained solely by the parameter capacity contributed by each component: the explicit transport mechanism supplied by advection has a disproportionately large effect on predictive performance.

\begin{figure}[htb]
\centering
\begin{subfigure}{0.75\linewidth}
  \centering
  \includegraphics[width=\linewidth]{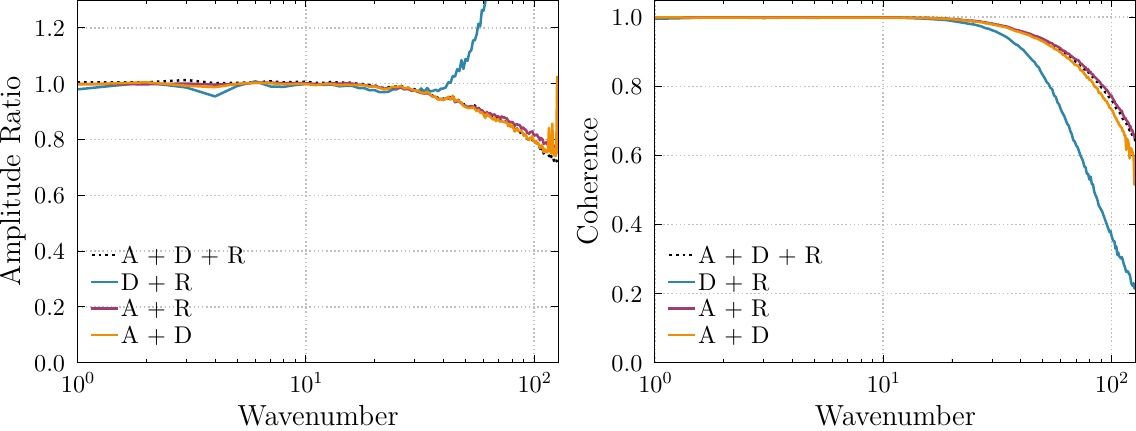}
  \caption{1 day}
\end{subfigure}

\vspace{0.5em}

\begin{subfigure}{0.75\linewidth}
  \centering
  \includegraphics[width=\linewidth]{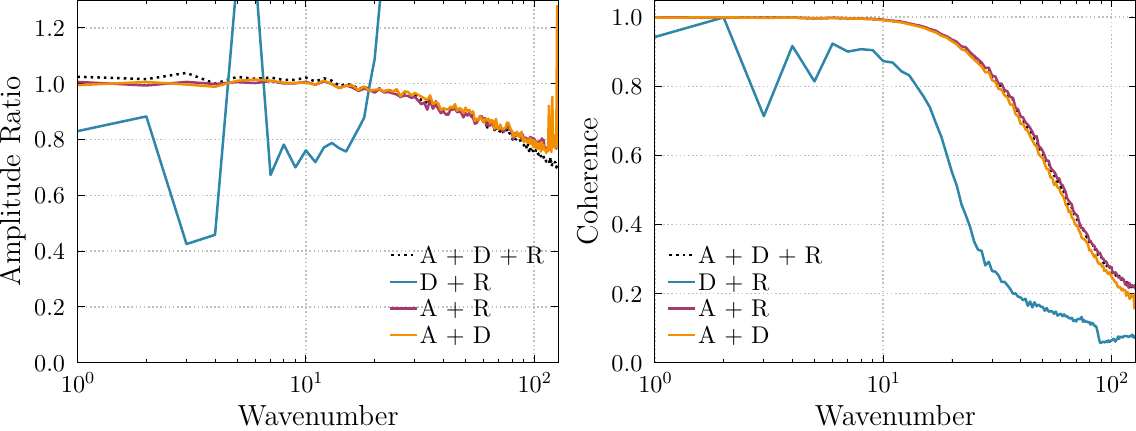}
  \caption{3 days}
\end{subfigure}

\vspace{0.5em}

\begin{subfigure}{0.75\linewidth}
  \centering
  \includegraphics[width=\linewidth]{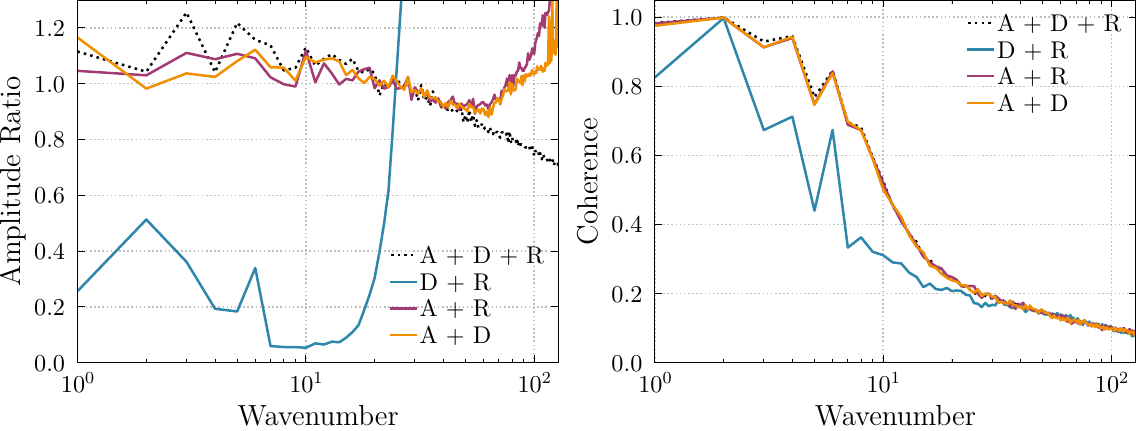}
  \caption{10 days}
\end{subfigure}

\caption{Amplitude ratio (left; closer to one is better) and spectral coherence (right; higher is better) for $z_{500}$ as functions of total spherical wavenumber. Results compare the complete PARADIS model ($A+D+R$) with configurations without advection ($D+R$), diffusion ($A+R$), or reaction ($A+D$) at lead times of 1, 3, and 10 days. Removing advection produces a rapid deterioration of both spectral amplitude and coherence, whereas configurations retaining advection remain substantially closer to the complete model.}
\label{fig:ablation_spectra}
\end{figure}

The spectral diagnostics in \Cref{fig:ablation_spectra} provide further insight into the role of each operator. At one day, the configurations containing advection remain close to the full model in both amplitude ratio and coherence. In contrast, the $D+R$ model already shows a substantially earlier loss of coherence at intermediate and high wavenumbers. By three days, the separation is much stronger: removing advection produces severe distortion of the amplitude spectrum together with a rapid collapse of spectral coherence, while the $A+R$, $A+D$, and complete models remain closely aligned.

At ten days, forecast coherence has decreased for all configurations at sufficiently small scales, as expected for an extended deterministic forecast. The distinction at larger and intermediate scales remains clear. The no-advection configuration loses coherence substantially earlier and develops a highly distorted amplitude spectrum, whereas the models retaining advection continue to follow the spectral behavior of the complete model much more closely.

Together, the pointwise and spectral ablations identify advection as the most consequential component of the ADR decomposition for forecast skill. Its removal causes substantially larger errors despite the remaining diffusion and reaction operators retaining considerable learnable capacity. The spectral results further show that this degradation is associated with an inability to propagate coherent structure across spatial scales. These findings support the use of explicit semi-Lagrangian transport as an inductive bias in PARADIS: rather than requiring the remaining learned operators to approximate transport implicitly, the architecture provides a dedicated mechanism for propagating latent atmospheric features through time.

\section{Learned Operators and Latent Fields} \label{sec:learned_operators}
To assess the internal representations learned by the model, we take a look at the structure of some learned operators and resulting latent feature maps. By visualizing components of the hidden state, we can qualitatively verify that the model's physically inspired inductive biases translate into physically structured, geographically coherent corrections and transport patterns.

\subsection{Latent Winds}

The NSL layer relies on an internally estimated velocity field $\mathbf{u}_l$ to transport information in latent space, and this is the dedicated mechanism of long-range information transport.  For weather processes governed by the Euler equations, information and energy are largely transported through wave processes.  The mode-1 baroclinic gravity wave (with a typical phase speed of about $50\, \mathrm{m}\,\mathrm{s}^{-1}$) is the fastest wave of meteorological significance, but large-scale weather is also driven by slower but more globally energetic Rossby waves.

Relevant atmospheric wave modes propagate over a range of characteristic speeds. We therefore expect the learned departure points to span a region
consistent with physically relevant propagation distances over the 6-hour forecast interval, while being displaced by the background flow. A characteristics-based PDE solver would need to integrate over this volume, and we expect the NSL layer to allow PARADIS to approximately accomplish this. Additionally, the domains of dependence for each of these wave modes are also Doppler-shifted by the local wind field, reflecting local Galilean invariance of the governing Euler equations.  This combination gives us two testable propositions:

\begin{itemize}
\item The set of upstream points selected by the NSL layer should approximately span the circular domain bounded by the forecast interval ($6\,\mathrm{hr}$) times the baroclinic wave speed, allowing the model to integrate over the correct domain of dependence, and
\item The set of upstream points should be centered on the point given by backward advection with respect to the physical wind fields, roughly $-6\, \mathrm{hr} \cdot \bar{\mathbf{u}}(x)$ for vertically-averaged physical wind fields $\bar{\mathbf{u}}$.
\end{itemize}

\begin{figure}[htb]
\centering
\includegraphics[width=\linewidth]{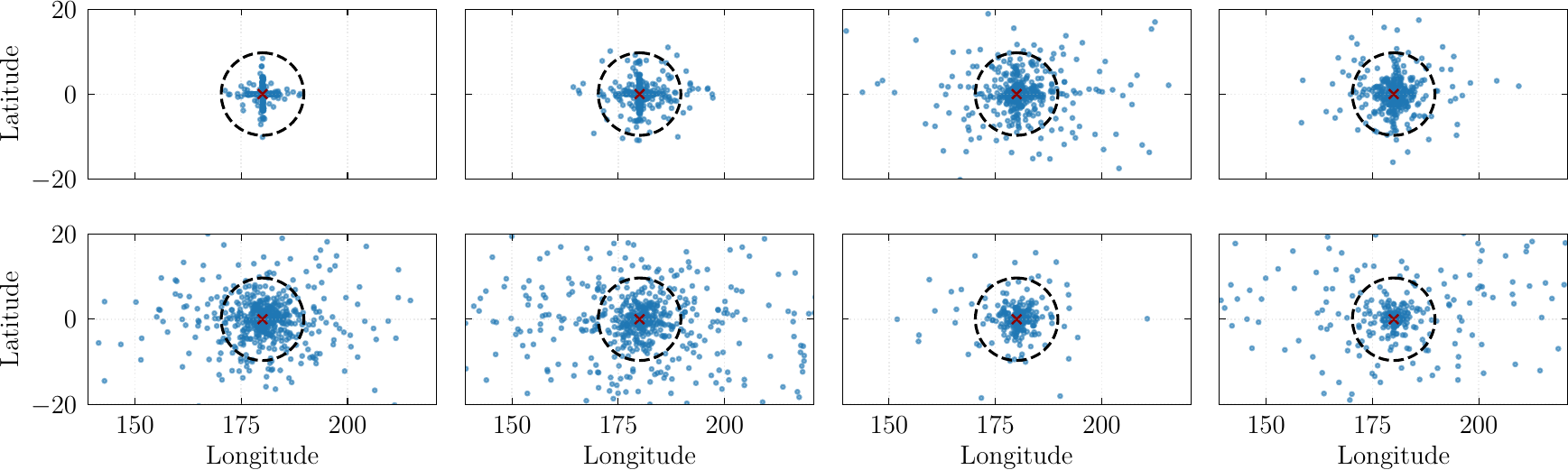}
\caption{Upstream points selected by the Neural Semi-Lagrangian layer for an arbitrary grid point, by model layer.  The collection of points (dots) generally spans the region covered by the domain of dependence of the first baroclinic wave mode (dashed circle).}
\label{fig:upstream_points}
\end{figure}

\begin{figure}[htb]
\centering

\begin{subfigure}{\linewidth}
  \centering
  \includegraphics[width=\linewidth]{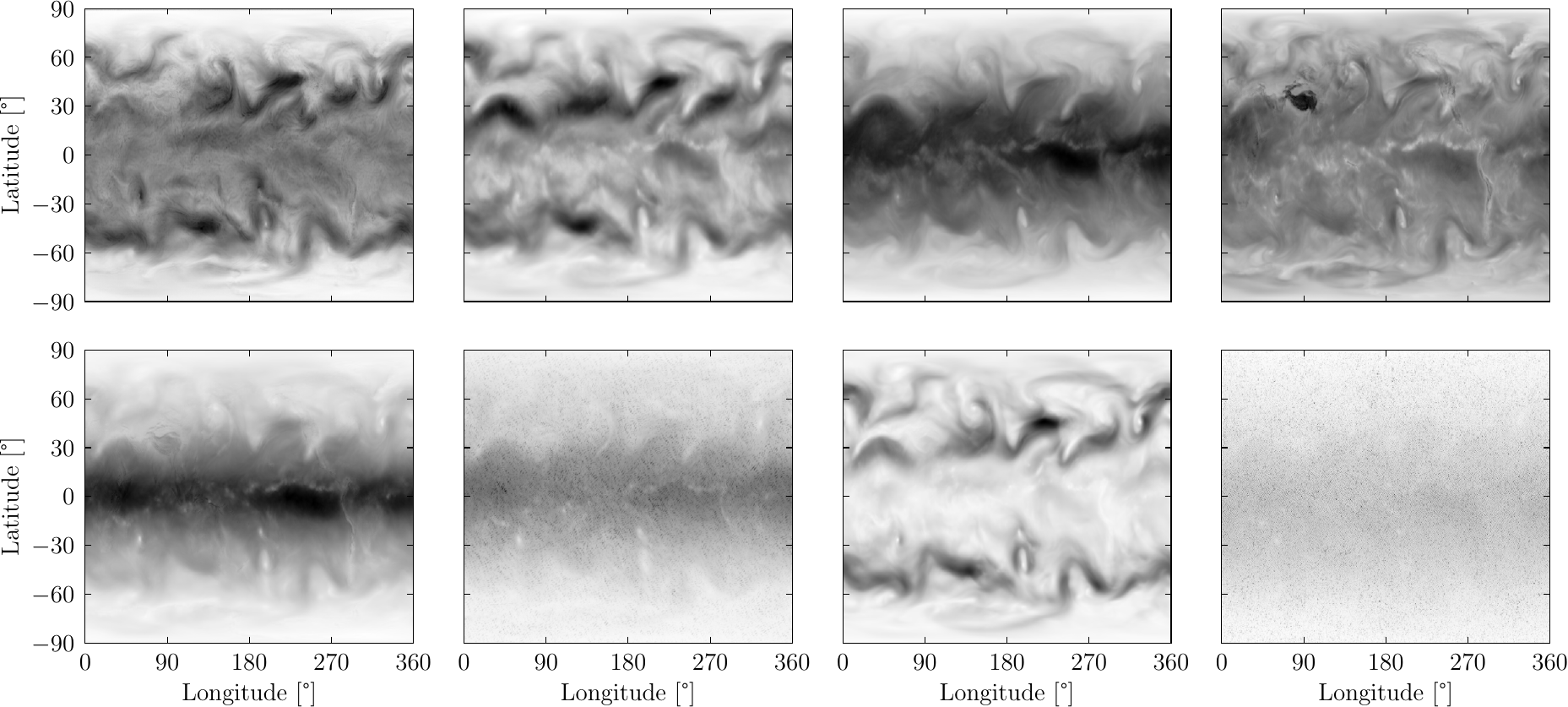}
  \caption{Latent winds per layer}
\end{subfigure}

\vspace{0.5em}

\begin{subfigure}{\linewidth}
  \centering
  \includegraphics[width=0.5\linewidth]{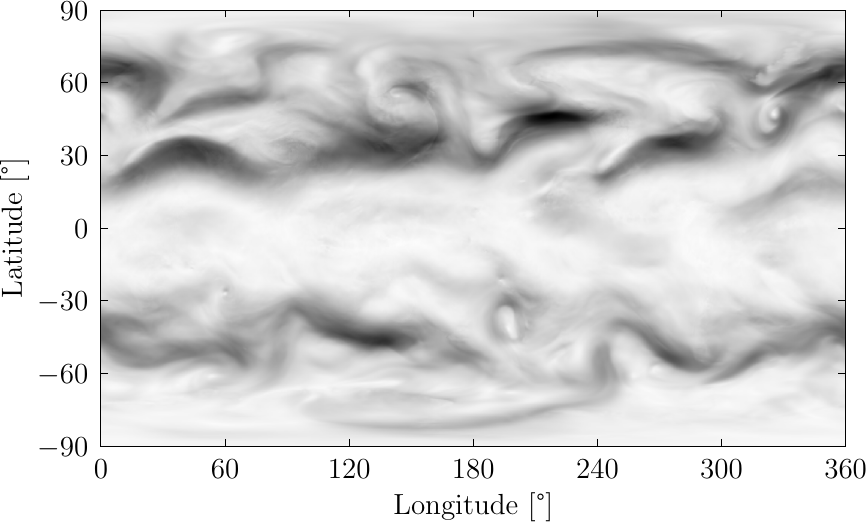}
  \caption{Physical winds}
\end{subfigure}

\caption{Evolution of sample learned velocity fields across forecast lead times. \textbf{Left} panels show vertically averaged physical wind speed, while \textbf{right} panels display the corresponding averaged latent wind speed, aggregated over all channels and layers of the Neural Semi-Lagrangian advection module. Results are shown for (a) 6 hours, (b) 5 days, and (c) 10 days lead time. The fields exhibit coherent, large-scale flow structures and increasing spatial complexity with lead time, indicating that the model learns physically consistent transport patterns in latent space.}
\label{fig:latent_velocities}
\end{figure}

Examining the velocities and upstream points learned by the NSL provides general support for both of these propositions.  \Cref{fig:upstream_points} shows the upstream points selected for an arbitrary target location and inference date through each layer of the model.  These points generally align with the maximum distance of baroclinic wave propagation\footnote{We speculate that slower wave modes are also represented here, but there is no clear way to separate them from the overall domain of dependence.}, and we suspect that outlying points are used for long-range regularization and reflect residual faster processes like diurnal heating.  \Cref{fig:latent_velocities} averages the latent wind speed estimates over all channels per layer, showing a clear correlation to the corresponding physical wind fields across all lead times.

\subsection{Learned Bias Fields}
\begin{figure}[htb]
\centering
\includegraphics[width=\linewidth]{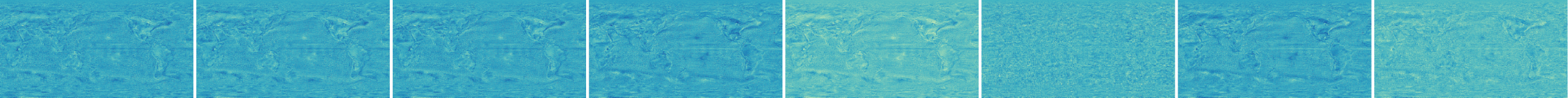}
\includegraphics[width=\linewidth]{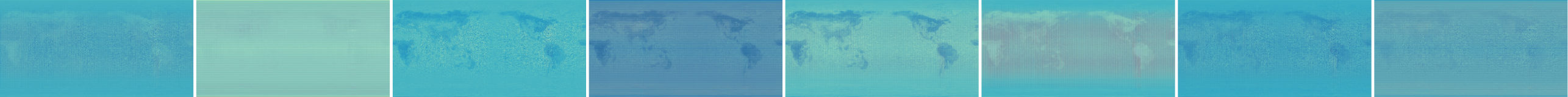}
\includegraphics[width=\linewidth]{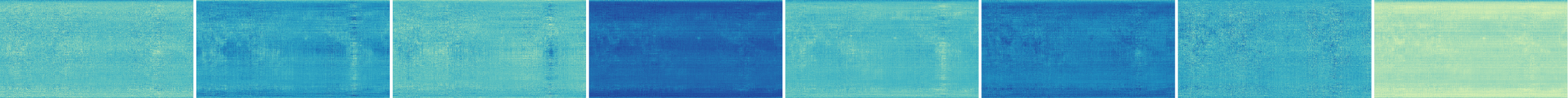}
\includegraphics[width=\linewidth]{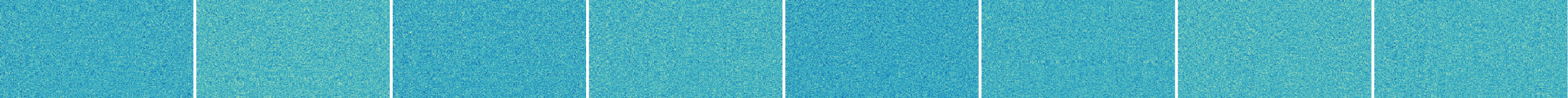}
\caption{Representative examples of reconstructed low-rank bias maps. Each panel shows a selected channel of the intermediate bias tensor $B'$ reconstructed from the learned rank-$K$ factors ($A,U,V$) in the low-rank bias module (\Cref{sec:low_rank_bias}). For clarity, only a subset of channels is shown. The bias maps exhibit spatially coherent structure, suggesting that the model learns large-scale, geographically structured corrections in latent space.}
\label{fig:bias_maps}
\end{figure}

 A key component for the Reaction and Diffusion blocks is the low-rank bias module. \Cref{fig:bias_maps} displays a representative subset of the reconstructed bias maps $B'$ within the model. Many of these maps exhibit clear, large-scale spatial structures that correlate with geographic features or zonal atmospheric patterns, suggesting that the network is learning systematic, geographically-structured corrections. However, it is important to note that this structured behavior is not universal across all channels. Some bias maps contain high-frequency or weakly structured patterns, which may reflect unconstrained degrees of freedom in the latent representation.

\section{Conclusion}\label{sec:conclusion}
We introduced a novel physics-inspired neural architecture for weather forecasting that decomposes the forecast operator into advection, diffusion, and reaction components, embedding a differentiable semi-Lagrangian operator to transport latent atmospheric features along learned trajectories. Instantiated in the PARADIS reference model, this architecture achieves competitive deterministic skill, with particularly strong short- to medium-range RMSE, while preserving greater forecast activity and spectral
fidelity than several leading data-driven baselines at longer lead times.

The results suggest that explicit transport is a useful inductive bias for neural weather prediction. In particular, the proposed architecture combines competitive pointwise accuracy with improved preservation of coherent small-scale variance, and the ablations indicate that the semi-Lagrangian advection component is central to this behavior.

Current limitations of the reference implementation include its deterministic nature and its dependence on the biases and data constraints of the reanalysis used for training. Future work will extend the proposed approach toward calibrated probabilistic forecasting, diagnostic-variable prediction, and improved representation of extreme events.

\section*{Code Availability}

The implementation of the PARADIS model presented in this work is available on GitHub at\\ \href{https://github.com/Wx-Alliance-Alliance-Meteo/paradis_model}{https://github.com/Wx-Alliance-Alliance-Meteo/paradis\_model}.

\section*{Declaration of generative AI and AI-assisted technologies in the manuscript preparation process}
During the preparation of this work, the authors used Claude, Gemini, and ChatGPT for software coding assistance, copyediting and language refinement. After using these tools, the authors reviewed and edited the content as needed and take full responsibility for the content of the published article.

\section*{Acknowledgements}
We would like to acknowledge the collaboration of our colleagues and the support of the managers of the \textit{Direction de la recherche en m\'et\'eorologie} and the \textit{Division du d\'eveloppement des pr\'evisions nationales} at Environment and Climate Change Canada. We also gratefully acknowledge the valuable contributions of Ben Adcock, Mahta Abdollahzadehzare, Avi Gupta, Mehdi Haned, Erika Kember, Francis Poulin, Steven Ruuth, Aryan Sharma, Seth Taylor, and Xiangye Xu.

\FloatBarrier

\bibliography{sn-bibliography}
\bibliographystyle{elsarticle-num}

\newpage

\appendix
\numberwithin{equation}{section}
\counterwithin{figure}{section}
\counterwithin{table}{section}

\section{Model Architecture}\label{sec:appendix_paradis_model}
The implemented architecture is designed to emulate an advection--diffusion--reaction process within a latent space. The overall design can be described as an encoder--processor--decoder structure. Let  $\phi$ and $\lambda$ be the latitude and longitude angles, respectively, of a spherical model of the Earth, i.e. $\mathbf{x} = \mathbf{x}(\phi, \lambda)$. This physical inspiration imposes hard constraints on layer behavior, ensuring the model evolves a high-dimensional representation $\mathbf{h}(\phi,\lambda) \in \mathbb{R}^{C}$ through components resembling partial differential equations (PDEs).

\subsection{Composite Blocks}\label{subsec:legos}

The architecture components are built from combinations of elementary linear operators: a \emph{channel mixer}, which acts independently at each spatial location, a \emph{spatial mixer}, which couples neighboring grid points through a separable convolution, and a bias term, which provides learned spatially varying offsets. More complex blocks are formed by composing these operators and adding nonlinearities as well as learned bias terms.

\paragraph{Channel mixer}{\label{sec:channel_mixer}}
The channel mixer is a pointwise linear transformation that mixes feature channels without spatial coupling.
It is implemented as a $1\times1$ convolution applied
independently at each latitude--longitude location. It is defined by
\begin{equation}
\mathbf{h}_{\text{out}}  = \mathbf{W} \mathbf{h}_{\text{in}} + \mathbf{b},
\end{equation}
where $\mathbf{W} \in \mathbb{R}^{C_{\text{out}} \times C_{\text{in}}}$ is a shared weight matrix and $\mathbf{b} \in \mathbb{R}^{C_{\text{out}}}$ is a learned bias. The number of trainable parameters in this layer is $(C_{\text{in}}+1)C_{\text{out}}$. 

\paragraph{Spatial mixer}{\label{sec:spatial_mixer}}
The spatial mixer introduces local spatial coupling while maintaining computational efficiency through a depthwise-separable convolution. This operation factorizes a standard convolution into a two-stage process that decouples spatial filtering from channel interaction. Formally, for an input latent field $\mathbf{h}_{\text{in}} \in \mathbb{R}^{C_{\text{in}} \times H \times W}$, we first apply geocyclic padding (\Cref{appendix:padding}) to ensure continuity on the sphere. The mixer is then defined by
\begin{equation}
\mathbf{h}_{\text{out}} = \tilde{\mathbf{W}} \left( \mathbf{K}_{d} \star \mathbf{h}_{\text{in}} \right) + \mathbf{b},
\end{equation}
where the operations are defined as follows
\begin{itemize}
    \item Depthwise Convolution ($\star$): A channel-wise spatial filter $\mathbf{K}_d \in \mathbb{R}^{C_{\text{in}} \times k \times k}$ applied independently to each channel. This stage couples neighboring grid points within the same channel using a $3 \times 3$ kernel ($k=3$), effectively acting as a learnable spatial stencil.
    \item Pointwise Projection: The channel mixer of \Cref{sec:channel_mixer} with weights $\tilde{\mathbf{W}} \in \mathbb{R}^{C_{\text{out}} \times C_{\text{in}}}$ and bias $\mathbf{b} \in \mathbb{R}^{C_{\text{out}}}$ linearly recombines the spatially-filtered features.
\end{itemize}
By separating these steps, the mixer reduces the parameter complexity from
\begin{equation}
O(k^2 C_{\text{in}} C_{\text{out}})
\quad\text{to}\quad
O(k^2 C_{\text{in}} + C_{\text{in}} C_{\text{out}}).
\end{equation}
In total, the layer contains $9C_{\text{in}}$ parameters for the depthwise kernels and $(C_{\text{in}}+1)C_{\text{out}}$ parameters for the subsequent channel-mixing projection.

\paragraph{Low-rank bias}\label{sec:low_rank_bias}
To encode large-scale, spatially coherent corrections while maintaining parameter efficiency, we employ a low-rank factorization of a global bias field. Rather than store a full $C_{\text{out}} \times H \times W$ bias tensor, which is computationally expensive for high-dimensional latent spaces, we define a low-rank bottleneck with $C_{\text{in}}$ channels and decompose it into a rank-$K$ separable representation.

The core of this mechanism consists of three factor matrices: $\mathbf{A} \in \mathbb{R}^{C_{\text{in}} \times K}$ (per-channel coefficients), $\mathbf{U} \in \mathbb{R}^{K \times H}$ (latitudinal factors), and $\mathbf{V} \in \mathbb{R}^{K \times W}$ (longitudinal factors). An intermediate bias map $\mathbf{B}' \in \mathbb{R}^{C_{\text{in}} \times H \times W}$ is first reconstructed via a rank-$K$ decomposition
\begin{equation}
\mathbf{B}'_{c,i,j} = \sum_{k=1}^{K} \mathbf{A}_{ck} \mathbf{U}_{ki} \mathbf{V}_{kj}.
\end{equation}
This intermediate field is then mapped to the target latent dimension $C_{\text{out}}$ through a learned linear projection $\hat{\mathbf{W}} \in \mathbb{R}^{C_{\text{out}} \times C_{\text{in}}}$. The final bias field $\mathbf{B} \in \mathbb{R}^{C_{\text{out}} \times H \times W}$ is computed as
\begin{equation}
\mathbf{B} _{o,i,j} = \sum_{c=1}^{C_{\text{in}}} \hat{\mathbf{W}}_{oc} \mathbf{B} '_{c,i,j}.
\end{equation}

This two-stage approach provides a significant reduction in parameters, from $C_{\text{out}} \times H \times W$ down to $K(C_{\text{in}} + H + W) + C_{\text{out}}C_{\text{in}}$, while retaining the capacity to represent smooth, large-scale spatial patterns across the latent dimensions. By decoupling the spatial rank of the bias from the total number of latent channels, the model can efficiently incorporate spatially coherent, channel-specific corrections throughout the model. We use $K=128$ by default.

\paragraph{Normalization layer}

We employ a channel-wise normalization operator, referred to as \emph{Channel Norm}, which is equivalent to applying Layer Normalization across the latent channel dimension at each spatial location. For an input field $\mathbf{x}(\phi, \lambda)\in\mathbb{R}^{C}$, \emph{Channel Norm} normalizes each latent vector independently, without coupling statistics across space or batch elements.

Formally, the normalized latent state $\hat{\mathbf{h}}$ is given by
\begin{equation} \hat{\mathbf{h}} = \boldsymbol{\gamma} \odot \frac{\mathbf{h} - \mu}{\sqrt{\sigma^2 + \varepsilon}} + \boldsymbol{\beta}, \end{equation}
where $\mu$ and $\sigma^2$ are the mean and variance computed across the channel dimension $C$, and $\{\boldsymbol{\gamma}, \boldsymbol{\beta}\} \in \mathbb{R}^C$ are learned affine parameters. A small constant $\varepsilon$ is added for numerical stability. This results in affine parameters with $2C$ degrees of freedom.

This form of normalization is standard in attention-based and graph neural network
architectures, where each spatial location (or token) is normalized independently in latent space. It is used throughout the implementation to stabilize training while preserving local spatial structure.

\subsection{Neural Semi-Lagrangian Advection}
\label{sec:nsl_advection}

Given a velocity field $\mathbf{u} = (u, v)$ and a time step $\Delta t$, the layer computes the advected state $\mathcal{A}_{\Delta t}(\mathbf{h}; \mathbf{u})$ by sampling $\mathbf{h}$ at departure points traced backward along characteristics.

\paragraph{Latent space projection}
To balance computational expressivity with memory efficiency, advection is performed in a reduced latent subspace of dimension $C_\text{adv}<C$. This reflects the idea that more than a single feature must likely travel with the same velocity field. We define a learned channel-wise down-projection $\mathcal{P}_\downarrow: \mathbb{R}^{C} \to \mathbb{R}^{V}$ and a corresponding lifting operator $\mathcal{P}_\uparrow: \mathbb{R}^{V} \to \mathbb{R}^{C}$. The advection step is applied to the projected state $\mathbf{z}$
\begin{equation}
  \mathbf{z} = \mathcal{P}_\downarrow(\mathbf{h}), \quad \mathbf{h}_{\text{adv}} = \mathcal{P}_\uparrow(\widetilde{\mathbf{z}}).
\end{equation}
This projection allows the model to identify the most salient features for transport, effectively decoupling the advective dynamics from the full hidden state dimensionality.

\paragraph{Pole continuity and geometric constraints}

The latitude--longitude representation possesses singularities at the poles where meridians converge. To ensure the latent state remains single-valued at the poles and to prevent numerical instabilities, we enforce a zonal mean constraint at this location
\begin{equation}
\mathbf{z}^{\text{pole}}_{c,i} = \frac{1}{W}\sum_{j=0}^{W-1} \mathbf{z}_{c,i,j}, \quad i \in \{0, H-1\}.
\end{equation}
By applying this pole constraint both before and after the advection interpolation step, we filter out spurious high-frequency longitudinal noise that typically accumulates near the poles in grid-based transport schemes.

\paragraph{Spherical backtracing via rotated coordinates}
Standard planar backtracing ($\mathbf{x}_d = \mathbf{x}_a - \mathbf{u}\Delta t$) introduces significant metric distortion near the poles. Following \cite{mcdonald1989semi}, we compute departure points $(\phi_d, \lambda_d)$ by treating the local displacement as a rotation on the unit sphere. For an arrival point $(\phi_a, \lambda_a)$, we define local angular displacements $\phi' = -v\Delta t$ and $\lambda' = -u\Delta t$. The departure coordinates are derived via spherical trigonometry
\begin{align}
\phi_d &= \arcsin\bigl( \sin \phi' \cos \phi_a + \cos \phi' \cos \lambda' \sin \phi_a \bigr), \\
\lambda_d &= \lambda_a + \operatorname{atan2}\bigl( \cos \phi' \sin \lambda' , \cos \phi' \cos \lambda' \cos \phi_a - \sin \phi' \sin \phi_a \bigr).
\end{align}
This formulation ensures that the trajectory remains valid across the entire sphere, including the polar regions.

\paragraph{Differentiable sampling and interpolation} The calculated departure coordinates $(\phi_d, \lambda_d)$ are mapped to continuous grid indices $(x, y)$. The advected state $\widetilde{\mathbf{z}}$ is then obtained via
\begin{equation}
\widetilde{\mathbf{z}} = \mathcal{I} \left( \mathbf{z}^{\text{pad}}, \phi_d, \lambda_d \right),
\end{equation}
where $\mathcal I$ denotes a bicubic interpolation kernel, implemented in PyTorch via \texttt{grid\_sample}.

\subsection{Diffusion and Reaction Terms}

The diffusion operator $\mathcal{D}_{\mathrm{net}}$ models local spatial
mixing and sub-grid dissipation. Like the advection and reaction operators,
it acts directly on the common $0.5^\circ$ latent processor grid. Spatial
mixing is implemented using the depthwise-separable SpatialMixer described
in \Cref{subsec:legos}.

The diffusion update is defined as
\begin{align}
\mathbf{h}_{\text{diff}}
= \mathcal{D}_{\text{net}}(\mathbf{h})
= 
    \operatorname{SpatialMixer}
    \left(
        \mathbf{h}
    \right)
+ \mathbf{B}_{\text{g}},
\end{align}
This structure allows the model to represent anisotropic diffusion patterns with a larger effective receptive field, while preserving the spatial dimensions of the latent processor state.

The reaction operator is evaluated on the same $0.5^\circ$ processor grid
and handles pointwise nonlinear transformations and source/sink terms.  The operator acts independently at each grid point through a two-layer channel mixer, while also conditioning on static fields such as topography, land--sea mask, and other time-invariant boundary information. These static inputs provide spatial context for locally varying physical processes without requiring the reaction operator itself to perform spatial mixing.

The reaction mapping is given by
\begin{subequations}\label{eq:react_layer}
\begin{align}
\mathbf{z}
&=
\mathbf{W}_1
\left[
    \mathbf{h}, \mathbf{s}
\right]
+ \mathbf{b}_1
+ \mathbf{B}_{\text{g}}, \\
\mathbf{h}_{\text{react}}
&=
\mathbf{W}_2 \operatorname{SiLU}(\mathbf{z}) + \mathbf{b}_2,
\end{align}
\end{subequations}
where $\mathbf{s}$ denotes the encoded static variables, $\mathbf{W}_1$ and $\mathbf{W}_2$ are channel-wise linear operators implemented as pointwise convolutions, and $\mathbf{B}_{\text{g}}$ is the low-rank global bias field. By incorporating static information and the global bias before the nonlinearity, the model can learn spatially varying reaction rates while preserving the spatial structure of the latent processor state.

\subsection{Feature Encoding}

\paragraph{Encoder}\label{subsubsec:input}

Let $\mathbf{x}(\phi, \lambda)\in\mathbb{R}^{C_{\text{input}}}$ denote the assembled input feature vector at grid location $(\phi, \lambda)$ (see \Cref{tab:variable_inventory} in the Appendix for a list of these input variables). The encoder maps these physical features into latent space $\mathbf{h}\in\mathbb{R}^C$.
This projection is implemented as a sequence of pointwise channel mixers.  For the single-layer case, which we apply in our work, the mapping is defined as
\begin{equation}\label{eq:input_layer}
\mathbf{h}(\phi, \lambda)  = \sigma\left(\mathbf{W}_{\text{enc}}\mathbf x(\phi, \lambda)  + \mathbf{b}\right),
\end{equation}
where $\mathbf{W}_{\text{enc}} \in \mathbb{R}^{C \times C_{\text{in}}}$.
By applying this transformation independently at each spatial location, the encoder initializes the latent representation while strictly preserving the local spatial structure of the input fields.

The encoded latent field remains on the $0.25^\circ$ input grid. Before
entering the ADR processor, it is spatially downsampled to the $0.5^\circ$
processor grid.

\paragraph{Decoder}\label{subsubsec:output}

After the final ADR processor layer, the latent state
$\mathbf{h}_N$ is interpolated from the $0.5^\circ$ processor grid back to the
$0.25^\circ$ grid. The interpolated state is then combined with the
full-resolution latent representation produced by the encoder:
\begin{equation}
\mathbf{h}_{\mathrm{dec}}
=
\mathcal{I}_{\uparrow}(\mathbf{h}_N)
+
\mathbf{h}^{f}_0.
\end{equation}

The resulting representation is mapped to the physical forecast variables
through the output projection,
\begin{subequations}\label{eq:output_layer}
\begin{align}
\mathbf{h}_{\mathrm{out}}
&=
\operatorname{SpatialMixer}(\mathbf{h}_{\mathrm{dec}})
+ \mathbf{B}_{\mathrm{g}}, \\
\mathbf{q}_{t+1}
&=
\mathbf{W}_{\mathrm{dec}}
\mathbf{h}_{\mathrm{out}}
+
\mathbf{b}_{\mathrm{dec}}.
\end{align}
\end{subequations}
where $\mathbf{W}_{\text{dec}} \in \mathbb{R}^{C_{\text{out}} \times C}$ is a learned linear operator that maps the $C$ latent channels back to the $C_{\text{out}}$ physical dimensions. By applying a spatial mixer (\Cref{subsec:legos}) immediately before the final projection, the decoder can smooth potential grid-scale artifacts and capture local spatial correlations that may have developed during the latent evolution. No additional non-linearities are applied in this stage, ensuring that $\mathbf{q}$ represents a direct linear decoding of the final processed features.

\paragraph{Static encoder}\label{subsubsec:static_encoder}

In addition to the prognostic atmospheric fields, we condition the latent evolution on static geographic variables. Let $\mathbf{s}(\phi,\lambda)\in\mathbb{R}^{C_{\text{static}}}$ denote the static feature vector at grid location $(\phi,\lambda)$, containing time-invariant fields such as topography, land--sea mask, and other constant information. These fields are encoded in addition to the prognostic state
\begin{equation}
\mathbf{h}_{\text{static}}(\phi,\lambda)
=
\mathcal{E}_{\text{static}}\bigl(\mathbf{s}(\phi,\lambda)\bigr),
\end{equation}
where $\mathbf{h}_{\text{static}}\in\mathbb{R}^{C_{\text{static-lat}}}$ is a lower-dimensional latent representation of the static inputs.

Unlike the prognostic encoder, which is purely pointwise, the static encoder includes local spatial mixing. It is implemented as a shallow convolutional network using depthwise-separable convolutions and geocyclic padding, allowing the model to extract spatially coherent geographic features while respecting the spherical boundary conditions. Because the static fields do not evolve in time, this representation is computed once per forecast step and supplied to each reaction block.

The encoded static representation is subsequently downsampled to the
processor grid using the same spatial coarsening as the prognostic latent
state and is supplied to each reaction block.

\section{Training Details}

\subsection{Training Objective}
\label{subsec:training_objective}

Training is performed by minimizing a \emph{reversed Huber} loss, which combines
linear behavior for small errors with quadratic behavior for large errors. Given a
prediction $\hat{y}$ and a reference value $y$, and a scale parameter
$\delta>0$, the loss is defined as
\begin{equation}
L_\delta(y,\hat{y}) =
\begin{cases}
\delta\,|y-\hat{y}|, & |y-\hat{y}| \le \delta, \\[4pt]
\frac{1}{2}(y-\hat{y})^2 + \frac{1}{2}\delta^2,
& |y-\hat{y}| > \delta.
\end{cases}
\end{equation}
In contrast to the classical Huber loss \citep{huber1992robust}, which is quadratic
near the origin and linear for large errors, the reversed Huber loss enforces
linear behavior for small residuals while penalizing large errors quadratically.
The construction satisfies the following requirements
\begin{enumerate}
\item[(i)] linear growth for small errors,
\item[(ii)] quadratic growth for large errors,
\item[(iii)] continuity at the transition point $|y-\hat{y}|=\delta$.
\end{enumerate}

The reversed Huber loss is non-negative, convex, and satisfies
$L_\delta(y,y)=0$. Its gradient with respect to the prediction $\hat{y}$ is given
by
\begin{equation}
\frac{\partial L_\delta}{\partial \hat{y}} =
\begin{cases}
-\delta\,\mathrm{sign}(y-\hat{y}),
& |y-\hat{y}| \le \delta, \\[4pt]
-(y-\hat{y}),
& |y-\hat{y}| > \delta.
\end{cases}
\end{equation}
The gradient is discontinuous at $|y-\hat{y}|=\delta$, which may hinder numerical
optimization.

To mitigate this issue, we employ a smooth approximation. Let $e=y-\hat{y}$ denote the prediction error.
The smoothed loss, which is used in this model, is defined as
\begin{equation}
\tilde{L}_\delta(e) =
\bigl(1-w(e)\bigr)\,\delta|e|
+ w(e)\,\frac{1}{2}e^2,
\end{equation}
where $w(e)$ is a sigmoid weighting function,
\begin{equation}
w(e) =
\frac{1}{1+\exp\!\bigl(-2(|e|-\delta)\bigr)}.
\end{equation}
This formulation replaces the sharp transition at $|e|=\delta$ with a smooth
interpolation between the linear and quadratic regimes. The gradient of the reversed Huber loss with respect to $\hat{y}$ is
\begin{equation}
\begin{aligned}
\frac{\partial \tilde{L}_\delta}{\partial \hat{y}}
&=
-\delta\,\mathrm{sign}(e)\bigl(1-w(e)\bigr)
- e\,w(e)
- \Bigl(\tfrac{1}{2}e^2-\delta|e|\Bigr)
\frac{\partial w}{\partial \hat{y}},
\end{aligned}
\end{equation}
with
\begin{equation}
\frac{\partial w}{\partial \hat{y}} =
-\frac{2\,\mathrm{sign}(e)\,
\exp\!\bigl(-2(|e|-\delta)\bigr)}
{\bigl(1+\exp\!\bigl(-2(|e|-\delta)\bigr)\bigr)^2}.
\end{equation}

The asymptotic behavior of $\tilde{L}_\delta$ matches that of the original reversed
Huber loss:
\begin{itemize}
\item for $|e|\ll\delta$, $w(e)\approx0$ and
$\tilde{L}_\delta(e)\approx\delta|e|$;
\item for $|e|\gg\delta$, $w(e)\approx1$ and
$\tilde{L}_\delta(e)\approx\frac{1}{2}e^2$.
\end{itemize}

The total training loss is obtained by applying $\tilde{L}_\delta$ independently
to each predicted variable and pressure level, followed by a weighted aggregation. To this end, we make use of latitude and pressure-level weights described as follows.

\subsubsection{Integration Weights}
To account for spherical geometry and the vertical distribution of atmospheric mass in the training objective, we apply spatial and pressure-level weights.

\paragraph{Latitude weighting} Since the grid cells on a regular latitude--longitude grid decrease in area toward the poles, we apply latitude weights $w_i$ to prevent over-representing polar regions. Following the methodology in WeatherBench~\citep{rasp_weatherbench_2024}, we define these weights to account for the integration of area on a sphere:
\begin{equation} w_i \propto
\begin{cases}
\sin^2\left(\frac{\Delta\phi}{4}\right) & \text{if } \phi_i = \pm 90^\circ \text{ (poles)},
\\ \cos(\phi_i) \sin\left(\frac{\Delta\phi}{2}\right) & \text{otherwise}
\end{cases}
\end{equation}
These weights are normalized to have a unit mean across the latitude dimension, ensuring that the magnitude of the loss metric remains comparable across different grid resolutions.

\paragraph{Pressure-level weighting} In addition to spatial weighting, we apply pressure-level weights $v_k$ when aggregating metrics across multiple atmospheric layers. This gives greater weight to the troposphere when aggregating the training loss, where the majority of the atmospheric mass and weather phenomena reside, and maintains a meaningful contribution from the upper atmosphere. Following Lang et al.~\citep{lang_aifs-crps_2024}, the weight for a given pressure level $p_k$ (expressed in hPa) is defined via
\begin{equation}
v_k = \max\left( \frac{p_k}{1000}, 0.2 \right),
\end{equation}
This linear scaling assigns higher importance to levels near the surface and applies a constant floor for levels above 200 hPa to prevent the stratosphere from being excessively disregarded in the loss function.

\paragraph{Variable (feature) weighting} We adopt a uniform weighting strategy across all physical output variables. While spatial and vertical weights account for the Earth's geometry and atmospheric mass distribution, the relative contribution of each prognostic variable---such as geopotential, temperature, specific humidity, and wind components---to the aggregate skill scores is considered equal for the purposes of the integration of the loss function.

\subsection{Optimization}

We use a learning rate of $5\times10^{-4}$ during single-step pre-training
and reduce it to $5\times10^{-6}$ during autoregressive fine-tuning. We apply momentum hyperparameters $\beta_1 = 0.9$ and $\beta_2 = 0.95$ for the $0.25^\circ$ training. A weight decay coefficient of $10^{-2}$ is applied. We maintain the maximum representational capacity of the high-dimensional latent space and do not employ any dropout mechanism. Furthermore, we find that the structural architecture of PARADIS does not require global gradient clipping for stabilization during any of the training stages.

\subsection{List of Hyperparameters}

Key architectural and training hyperparameters are listed in \Cref{tab:hyperparameters}.

\begin{table}[ht]
\centering
\caption{Hyperparameters and training configuration for PARADIS.}
\label{tab:hyperparameters}
\small
\begin{adjustbox}{max width=\linewidth}
\begin{tabular}{@{}llc@{}}
\toprule
\textbf{Category} & \textbf{Hyperparameter} & \textbf{Value} \\ \midrule
\multirow{9}{*}{Model Architecture}
    & Latent dimension ($C$) & 1024 \\
    & Number of processor layers ($\nsubstep$) & 8 \\
    & Velocity subspace vectors ($C_{\text{adv}}$) & 768 \\
    & Bias latent channels ($C_{\text{in}}$) & 8 \\
    & Advection interpolation & Bicubic \\
    & Processor resolution & $0.5^\circ$ \\
    & Encoder layers & 1 \\
    & Bias rank & 128 \\
    & Time-history inputs & 2 \\ \midrule
\multirow{12}{*}{Training \& Optimization}
    & Pre-training steps & 260,000 \\
    & Total AR fine-tuning steps & $6\times1,000$ \\
    & Global batch size & 16  \\
    & Learning rate & $5 \times 10^{-4}$ (pre-training), $5 \times 10^{-6}$ (fine-tuning) \\
    & Optimizer & AdamW \\
    & Weight decay & $10^{-2}$ \\
    & Betas ($\beta_1, \beta_2$) & 0.9, 0.95 \\
    & Loss function & Pseudo reversed Huber \\
    & Huber delta ($\delta$) & 1.0 \\
    & Scheduler & WSD \\
    & Warmup steps (Pre-training) & 1,000 \\
    & Decay period & 20\% \\ \midrule
\multirow{5}{*}{Dataset \& Compute}
    & Time resolution & 6h \\
    & Input/Output resolution & $0.25^\circ$ \\
    & Vertical levels & 13 \\
    & Total number of input features & 216 \\
    & Automatic mixed precision (AMP) & bf16 \\ \bottomrule
\end{tabular}
\end{adjustbox}
\end{table}

\subsection{Training Curriculum}\label{sec:training_curriculum}
PARADIS is trained using a two-phase curriculum consisting of single-step pre-training followed by progressively longer autoregressive fine-tuning. The model is first trained for 260,000 optimizer steps using a 6-hour forecast horizon. This stage establishes accurate single-step dynamics before the model is exposed to its own predictions during autoregressive training.

Following pre-training, the forecast horizon is progressively increased from 12 to 72 hours. Starting from the pre-trained model, we fine-tune sequentially using rollout lengths of 2, 4, 6, 8, 10, and 12 forecast steps, corresponding to horizons of 12, 24, 36, 48, 60, and 72 hours. Each stage consists of 1,000 optimizer updates and is initialized from the preceding stage. This curriculum gradually exposes the model to increasingly long trajectories while avoiding the optimization and memory costs associated with training directly on long rollouts. The complete training schedule is summarized in \Cref{tab:unified_curriculum}.

\subsubsection{Pre-training: Learning Rate Schedule}
\label{sec:phase1_wsd}

During pre-training, we perform 260,000 optimization steps using a 6-hour prediction horizon. We employ a Warmup-Steady-Decay (WSD) learning-rate scheduler~\citep{hu_minicpm_2024}. The schedule consists of a short warmup period followed by a constant-learning-rate phase covering most of training, and a final decay phase that progressively reduces the learning rate as optimization approaches convergence.

Concretely, we warm up for 1,000 steps and apply the decay over the final 20\% of the pre-training schedule. This long single-step training phase provides the initialization for the subsequent autoregressive curriculum.

\subsubsection{Autoregressive Fine-tuning}

The pre-trained model is subsequently fine-tuned using progressively longer autoregressive rollouts. Since each model application advances the atmospheric state by 6 hours, rollout lengths of
$N_{\mathrm{rollout}}=\{2,4,6,8,10,12\}$
correspond to forecast horizons ranging from 12 to 72 hours. Each rollout stage is trained for 1,000 optimizer updates before its weights are used to initialize the next stage.

For a rollout of length $N_{\mathrm{rollout}}$, the model recursively produces
\begin{equation}
    \left(
    \mathbf{q}_{t+1},
    \mathbf{q}_{t+2},
    \ldots,
    \mathbf{q}_{t+N_{\mathrm{rollout}}}
    \right),
\end{equation}
with each predicted state used as input to the subsequent forecast step. Losses are accumulated over the predicted lead times with full backpropagation over the rollout window.

The progressive rollout curriculum exposes the model to errors generated by its own predictions and encourages stable behavior under repeated application of the forecasting operator. The learning rate is reduced relative to pre-training during this phase to preserve the single-step forecast skill while adapting the model to longer autoregressive trajectories.

\begin{table}[ht]
\centering
\small
\caption{Training curriculum for PARADIS. Pre-training uses a single 6-hour forecast step, followed by progressively longer autoregressive rollouts. Steps denote the number of optimizer updates.}
\label{tab:unified_curriculum}
\begin{adjustbox}{max width=\linewidth}
\begin{tabular}{@{}llccc@{}}
\toprule
\textbf{Phase} & \textbf{Stage} & \textbf{Rollout Steps} &
\textbf{Horizon} & \textbf{Updates} \\
\midrule
\textbf{Pre-training}
    & Single-step & 1 & 6h & 260,000 \\
\midrule
\multirow{6}{*}{\textbf{AR Fine-tuning}}
    & AR Stage 1 & 2  & 12h & 1,000 \\
    & AR Stage 2 & 4  & 24h & 1,000 \\
    & AR Stage 3 & 6  & 36h & 1,000 \\
    & AR Stage 4 & 8  & 48h & 1,000 \\
    & AR Stage 5 & 10 & 60h & 1,000 \\
    & AR Stage 6 & 12 & 72h & 1,000 \\
\bottomrule
\end{tabular}
\end{adjustbox}
\end{table}

\subsubsection{Compute Details}
\label{sec:compute_details}

PARADIS was trained on a distributed cluster comprising 16 NVIDIA H100 GPUs using Distributed Data Parallelism (DDP). Training was performed with automatic mixed precision using the \texttt{bf16} format to reduce memory consumption and improve computational throughput. All training stages use $0.25^\circ$ input and target fields, with the
latent processor operating at $0.5^\circ$.

The initial single-step pre-training phase required approximately $0.30$ GPU-years. The subsequent autoregressive fine-tuning curriculum, in which the rollout horizon was progressively increased from 2 to 12 forecast steps, required an additional $0.05$ GPU-years. The total training cost of the final model was therefore approximately $0.35$ GPU-years.

To accommodate the memory requirements of the high-dimensional latent architecture, we employ gradient checkpointing during training. This is particularly important during autoregressive fine-tuning, where gradients are propagated through multiple forecast steps. Gradient checkpointing allows the 1024-dimensional latent representation and long autoregressive
training graphs to fit within H100 device memory.

Once trained, the model produces a complete 10-day autoregressive forecast at $0.25^\circ$ resolution in approximately 35\,s on a single NVIDIA H100 GPU.

\section{Datasets}

\subsection{Training Dataset: ECMWF's ERA5}
For training and evaluation, we employed ECMWF's ERA5 reanalysis (Copernicus and CC-BY licenses), which delivers a high-resolution global reconstruction of the atmospheric state from 1959 onward. ERA5 is generated using the HRES model within ECMWF's 4D-Var assimilation framework \citep{andersson2008ecmwf}, combining observations over 12-hour windows with short-range forecasts. The dataset provides hourly output at $0.25^\circ$ resolution.

Our study uses a tailored subset spanning 1979-01-01 to 2019-12-31, sampled at four synoptic times per day (00, 06, 12, and 18 UTC). The selected configuration includes surface variables and 13 pressure levels: 50, 100, 150, 200, 250, 300, 400, 500, 600, 700, 850, 925, and 1000 hPa.

To ensure efficiency during training, the dataset was directly obtained from WeatherBench~\cite{rasp_weatherbench_2024}, reprocessed, and stored in re-chunked form along the time dimension to reduce random-access overhead.

\begin{table}[ht]
\centering
\footnotesize
\caption{Input and output variable inventory for the PARADIS model. Atmospheric variables are evaluated at 13 pressure levels (50, 100, 150, 200, 250, 300, 400, 500, 600, 700, 850, 925, 1000 hPa).}
\label{tab:variable_inventory}
\setlength{\tabcolsep}{3pt}
\begin{adjustbox}{max width=\linewidth}
\begin{tabular}{llcc}
\toprule
\textbf{Category} & \textbf{Variable Name} & \textbf{Input} & \textbf{Output} \\
\midrule
\multirow{5}{*}{Atmospheric} & Geopotential  & \cmark & \cmark \\
 & Cartesian horizontal wind vector & \cmark & \cmark \\
 & Specific humidity & \cmark & \cmark \\
 & Temperature  & \cmark & \cmark \\
 & Vertical velocity & \cmark & \cmark \\
\midrule
\multirow{5}{*}{Surface} & 10m Cartesian horizontal wind vector   & \cmark & \cmark \\
 & 2m temperature  & \cmark & \cmark \\
 & Mean sea-level pressure & \cmark & \cmark \\
 & Surface pressure  & \cmark & \cmark \\
 & Total column water  & \cmark & \cmark \\
\midrule
\multirow{3}{*}{Forcings} & TOA incident solar radiation & \cmark & \xmark \\
 & Time of day ($\sin, \cos$) & \cmark & \xmark \\
 & Year progress ($\sin, \cos$) & \cmark & \xmark \\
\midrule
\multirow{8}{*}{Static fields} & Geopotential at surface  & \cmark & \xmark \\
 & Land-sea mask & \cmark & \xmark \\
 & Topographic slope & \cmark & \xmark \\
 & Sub-grid topographic standard deviation & \cmark & \xmark \\
 & Inverse longitude spacing & \cmark & \xmark \\
 & Latitude / longitude & \cmark & \xmark \\
 & $\cos(\phi)$ & \cmark & \xmark \\
 & $\sin(\lambda), \cos(\lambda)$ & \cmark & \xmark \\
\bottomrule
\end{tabular}
\end{adjustbox}
\end{table}

\subsection{Evaluation Datasets}
\label{subsec:analysis_datasets}
To benchmark the performance of the proposed model, we compare our forecasts against both traditional numerical weather prediction systems (ECMWF HRES) and a diverse set of state-of-the-art data-driven forecasting models spanning a wide range of architectural approaches, including graph neural networks (GraphCast), transformer-based models (Pangu-Weather, Aurora, and Baguan), diffusion-based generative forecasting (GenCast), spectral operator methods (FGN), and cascaded forecasting systems (FuXi).

\begin{table}[ht]
\centering
\caption{Variable reference: short names and long names for meteorological fields used in evaluation.}
\label{tab:variable_reference}
\begin{adjustbox}{max width=\linewidth}
\begin{tabular}{@{}lll@{}}
\toprule
\textbf{Category} & \textbf{Short Name} & \textbf{Long Name} \\ \midrule
\multirow{5}{*}{Atmospheric}
    & $z$ & Geopotential \\
    & $t$ & Temperature \\
    & $u$ & $u$-component of wind (zonal) \\
    & $v$ & $v$-component of wind (meridional) \\
    & $q$ & Specific humidity \\ \midrule
\multirow{3}{*}{Surface}
    & $2t$& 2-metre temperature \\
    & $10u$ & 10-metre $u$-component of wind \\
    & $msl$ & Mean sea-level pressure \\ \bottomrule
\end{tabular}
\end{adjustbox}
\end{table}

Short and long names of evaluated variables are listed in \Cref{tab:variable_reference}.

\subsection{Data Normalization}

For each variable, we compute the mean and standard deviation independently at each pressure level. Then, $z$-score normalization is applied as follows
\begin{equation}
q_{\text{normalized}} = \frac{q - \mu(q)}{\sigma(q)}.
\end{equation}
Here, $\mu(q)$ and $\sigma(q)$ represent the mean and standard deviation of the variable $q$ for a given pressure level.

\section{Verification Strategies}\label{sec:evaluation_metrics}

To assess the performance of the model, we employ a suite of metrics that evaluate the forecast accuracy in both physical and spectral space.
\subsection{RMSE Skill}
The RMSE provides a measure of the average magnitude of the forecast error. To account for the convergence of meridians toward the poles, we utilize an area-weighted RMSE
\begin{equation}\text{RMSE}(t) = \sqrt{\frac{1}{\sum_{i,j} w_i} \sum_{i=1}^{H} \sum_{j=1}^{W} w_i \left( \hat{y}_{i,j}(t) - y_{i,j}(t) \right)^2},
\end{equation}
where $\hat{y}$ and $y$ are the predicted and ground-truth fields, respectively. The weight $w_i$ corresponds to that of the latitude $\phi_i$ at grid row $i$. This weighting ensures that errors near the equator (which represent a larger physical area) are prioritized appropriately. These weights are computed according to~\cite{rasp_weatherbench_2024, lam2023learning} for datasets containing the poles.
    \subsection{ACC skill}
The anomaly correlation coefficient (ACC) measures the spatial similarity between the predicted and observed anomalies (deviations from the long-term climatology). It is highly sensitive to the phase and structure of weather systems. Following~\cite{rasp_weatherbench_2024}
\begin{equation}
\text{ACC}(t) = \frac{\sum_{i,j} w_i \left( f_{i,j}(t) \cdot o_{i,j}(t) \right)}{\sqrt{\sum_{i,j} w_i (f_{i,j}(t))^2 \cdot \sum_{i,j} w_i (o_{i,j}(t))^2}},
\end{equation}
where ${f} = \hat{y} - \bar{y}$ and $o = y - \bar{y}$ are the predicted and observed anomalies relative to the climatological mean $\bar{y}$. An ACC of 1 indicates perfect spatial correlation, while a value of 0.6 is typically considered the threshold for a ``useful'' synoptic forecast \citep{murphy_skill_1989}.
\subsection{Prediction Activity}
Prediction Activity measures the amplitude of the forecast anomalies relative to climatology and provides an estimate of the dynamical variability represented by the model forecasts. In contrast to ACC, which evaluates the spatial agreement between forecast and observed anomalies, Prediction Activity quantifies the strength of the predicted anomalous patterns themselves. Following the implementation used in \cite{rasp_weatherbench_2024} and WeatherBench-X, the activity is defined as
\begin{equation}
\text{Activity}(t) = \sqrt{\frac{\sum_{i,j} w_i \left(f_{i,j}(t)\right)^2}{\sum_{i,j} w_i}},
\end{equation}
where \(f = \hat{y} - \bar{y}\)
denotes the forecast anomaly relative to the climatological mean
\(\bar{y}\),
and \(w_i\) are the latitude-dependent area weights. This metric can be interpreted as the spatial standard deviation of the forecast anomalies and is useful for assessing whether the model underestimates or overestimates the intensity of atmospheric variability at different lead times. We compute the relative activity with respect to ERA5 via
\begin{equation}
    \Delta\text{Activity} = \frac{\text{Activity}_f -\text{Activity}_\text{ERA5}}{\text{Activity}_\text{ERA5}},
\end{equation}
where negative values indicate smoothing.

\subsection{Power Spectra} \label{subsec:powerspectra}
The amplitude ratio is the square root of the power spectral density ratio between the forecast and verifying analysis, where we add together all spherical harmonic modes of the same total wavenumber.  Since the full power spectrum decays over several orders of magnitude from the synoptic scales to the finest scales, this normalization highlights the departure from the ground-truth reference and depicts the scale-dependent smoothing (or noise generation) of a model.

The spectral coherence is the correlation coefficient between the individual amplitudes of the forecast and the verifying ground truth, again grouped by total wavenumber.  This shows the scale-dependent forecast skill, independently of smoothing.

For plotting purposes, these values are computed independently per forecast and then averaged over the 2020 verification period.

Given fixed forecast skill, MSE optimality is reached when a model's smoothing (amplitude ratio) is equal to its coherence on a mode-by-mode basis, and a similar relationship holds under MAE-style loss functions with mild assumptions about distribution regularity.  ML-based models such as PARADIS, GraphCast, and Pangu-Weather approach this optimum ratio.  The physics-based HRES baseline is designed to produce realistic forecast fields rather than to give an MSE-optimal forecast, so its amplitude ratio is approximately 1 for all modes.

\section{Tables of Comparison with Baseline Models}
\label{sec:baseline_comparison}

\begin{table*}
\centering
\tiny
\caption{RMSE for selected atmospheric variables at multiple pressure levels, evaluated across forecast lead times. Bold values denote the best-performing model; underlined values denote the second-best.}
\renewcommand{\arraystretch}{0.8}
\setlength{\tabcolsep}{6pt}
\begin{adjustbox}{max width=\linewidth}
\begin{tabular}{lcccccccc}
\toprule
 & \multicolumn{8}{c}{\textbf{$z_{500}$}} \\
\cmidrule(lr){2-9}
 & \multicolumn{4}{c}{2020} & \multicolumn{4}{c}{2022} \\
\cmidrule(lr){2-5}\cmidrule(lr){6-9}
Model & 24h & 72h & 120h & 240h & 24h & 72h & 120h & 240h \\
\midrule
PARADIS & \textbf{3.86e+01} & \textbf{1.19e+02} & \textbf{2.72e+02} & 7.48e+02 & \textbf{3.93e+01} & \textbf{1.20e+02} & \underline{2.71e+02} & 7.54e+02 \\
Aurora & - & - & - & - & 4.44e+01 & \underline{1.23e+02} & \textbf{2.66e+02} & \textbf{6.73e+02} \\
Baguan & 4.30e+01 & 1.29e+02 & \underline{2.75e+02} & \textbf{6.28e+02} & - & - & - & - \\
Climatology & 8.21e+02 & 8.22e+02 & 8.23e+02 & 8.23e+02 & 8.11e+02 & 8.11e+02 & 8.12e+02 & 8.15e+02 \\
IFS ens & 6.55e+01 & 1.97e+02 & 3.97e+02 & 8.72e+02 & 6.50e+01 & 1.97e+02 & 3.96e+02 & 8.79e+02 \\
FGN (member 1) & - & - & - & - & 5.13e+01 & 1.43e+02 & 3.17e+02 & 8.16e+02 \\
FuXi & 4.28e+01 & 1.28e+02 & 2.80e+02 & \underline{6.34e+02} & - & - & - & - \\
GenCast (member 1) & 5.58e+01 & 1.67e+02 & 3.55e+02 & 8.43e+02 & - & - & - & - \\
GraphCast & \underline{4.12e+01} & \underline{1.26e+02} & 2.76e+02 & 7.34e+02 & \underline{4.05e+01} & 1.24e+02 & 2.72e+02 & \underline{7.35e+02} \\
HRES & 4.86e+01 & 1.39e+02 & 3.08e+02 & 8.05e+02 & 4.82e+01 & 1.41e+02 & 3.10e+02 & 8.13e+02 \\
Pangu & 4.60e+01 & 1.36e+02 & 2.96e+02 & 7.81e+02 & 4.63e+01 & 1.36e+02 & 2.98e+02 & 7.92e+02 \\
\bottomrule
\end{tabular}
\end{adjustbox}

\begin{adjustbox}{max width=\linewidth}
\begin{tabular}{lcccccccc}
\toprule
 & \multicolumn{8}{c}{\textbf{$t_{850}$}} \\
\cmidrule(lr){2-9}
 & \multicolumn{4}{c}{2020} & \multicolumn{4}{c}{2022} \\
\cmidrule(lr){2-5}\cmidrule(lr){6-9}
Model & 24h & 72h & 120h & 240h & 24h & 72h & 120h & 240h \\
\midrule
PARADIS & \underline{6.13e-01} & 1.05e+00 & 1.67e+00 & 3.49e+00 & \underline{6.15e-01} & \underline{1.05e+00} & 1.67e+00 & 3.46e+00 \\
Aurora & - & - & - & - & 8.05e-01 & 1.08e+00 & \textbf{1.60e+00} & \textbf{3.05e+00} \\
Baguan & 6.37e-01 & \underline{1.04e+00} & \textbf{1.62e+00} & \textbf{2.89e+00} & - & - & - & - \\
Climatology & 3.49e+00 & 3.48e+00 & 3.48e+00 & 3.49e+00 & 3.46e+00 & 3.46e+00 & 3.46e+00 & 3.47e+00 \\
IFS ens & 1.06e+00 & 1.61e+00 & 2.34e+00 & 3.98e+00 & 1.13e+00 & 1.64e+00 & 2.34e+00 & 3.97e+00 \\
FGN (member 1) & - & - & - & - & 9.72e-01 & 1.32e+00 & 1.95e+00 & 3.74e+00 \\
FuXi & 6.45e-01 & 1.07e+00 & 1.68e+00 & \underline{2.98e+00} & - & - & - & - \\
GenCast (member 1) & 8.52e-01 & 1.40e+00 & 2.11e+00 & 3.86e+00 & - & - & - & - \\
GraphCast & \textbf{6.01e-01} & \textbf{1.03e+00} & \underline{1.63e+00} & 3.42e+00 & \textbf{5.95e-01} & \textbf{1.02e+00} & \underline{1.62e+00} & \underline{3.42e+00} \\
HRES & 8.57e-01 & 1.30e+00 & 1.94e+00 & 3.73e+00 & 9.42e-01 & 1.34e+00 & 1.96e+00 & 3.73e+00 \\
Pangu & 7.31e-01 & 1.15e+00 & 1.79e+00 & 3.61e+00 & 7.32e-01 & 1.15e+00 & 1.79e+00 & 3.62e+00 \\
\bottomrule
\end{tabular}
\end{adjustbox}

\begin{adjustbox}{max width=\linewidth}
\begin{tabular}{lcccccccc}
\toprule
 & \multicolumn{8}{c}{\textbf{$q_{700}$}} \\
\cmidrule(lr){2-9}
 & \multicolumn{4}{c}{2020} & \multicolumn{4}{c}{2022} \\
\cmidrule(lr){2-5}\cmidrule(lr){6-9}
Model & 24h & 72h & 120h & 240h & 24h & 72h & 120h & 240h \\
\midrule
PARADIS & \underline{5.89e-04} & 9.44e-04 & 1.24e-03 & 1.83e-03 & \underline{5.82e-04} & 9.26e-04 & 1.21e-03 & 1.80e-03 \\
Aurora & - & - & - & - & 6.46e-04 & \underline{8.97e-04} & \textbf{1.12e-03} & \textbf{1.52e-03} \\
Baguan & 6.04e-04 & \underline{9.17e-04} & \textbf{1.17e-03} & \textbf{1.51e-03} & - & - & - & - \\
Climatology & 1.68e-03 & 1.68e-03 & 1.68e-03 & \underline{1.68e-03} & 1.67e-03 & 1.67e-03 & 1.67e-03 & \underline{1.67e-03} \\
IFS ens & 9.55e-04 & 1.35e-03 & 1.64e-03 & 2.08e-03 & 9.35e-04 & 1.33e-03 & 1.64e-03 & 2.11e-03 \\
FGN (member 1) & - & - & - & - & 8.62e-04 & 1.17e-03 & 1.44e-03 & 1.95e-03 \\
GenCast (member 1) & 8.26e-04 & 1.27e-03 & 1.56e-03 & 2.04e-03 & - & - & - & - \\
GraphCast & \textbf{5.79e-04} & \textbf{9.10e-04} & \underline{1.17e-03} & 1.68e-03 & \textbf{5.68e-04} & \textbf{8.82e-04} & \underline{1.13e-03} & 1.67e-03 \\
HRES & 8.39e-04 & 1.20e-03 & 1.49e-03 & 1.99e-03 & 8.19e-04 & 1.18e-03 & 1.48e-03 & 2.01e-03 \\
Pangu & 6.61e-04 & 1.00e-03 & 1.31e-03 & 1.89e-03 & 6.51e-04 & 9.81e-04 & 1.28e-03 & 1.86e-03 \\
\bottomrule
\end{tabular}
\end{adjustbox}

\begin{adjustbox}{max width=\linewidth}
\begin{tabular}{lcccccccc}
\toprule
 & \multicolumn{8}{c}{\textbf{$u_{850}$}} \\
\cmidrule(lr){2-9}
 & \multicolumn{4}{c}{2020} & \multicolumn{4}{c}{2022} \\
\cmidrule(lr){2-5}\cmidrule(lr){6-9}
Model & 24h & 72h & 120h & 240h & 24h & 72h & 120h & 240h \\
\midrule
PARADIS & \textbf{1.18e+00} & \textbf{2.15e+00} & 3.38e+00 & 6.23e+00 & \textbf{1.18e+00} & 2.13e+00 & 3.34e+00 & 6.18e+00 \\
Aurora & - & - & - & - & 1.34e+00 & \underline{2.13e+00} & \textbf{3.17e+00} & \textbf{5.24e+00} \\
Baguan & 1.23e+00 & \underline{2.16e+00} & \textbf{3.26e+00} & \textbf{5.05e+00} & - & - & - & - \\
Climatology & 5.81e+00 & 5.82e+00 & 5.82e+00 & 5.82e+00 & 5.78e+00 & 5.78e+00 & 5.78e+00 & \underline{5.80e+00} \\
IFS ens & 2.02e+00 & 3.31e+00 & 4.68e+00 & 7.08e+00 & 2.00e+00 & 3.28e+00 & 4.64e+00 & 7.08e+00 \\
FGN (member 1) & - & - & - & - & 1.76e+00 & 2.72e+00 & 4.00e+00 & 6.73e+00 \\
FuXi & 1.25e+00 & 2.20e+00 & 3.37e+00 & \underline{5.13e+00} & - & - & - & - \\
GenCast (member 1) & 1.71e+00 & 2.95e+00 & 4.32e+00 & 6.89e+00 & - & - & - & - \\
GraphCast & \underline{1.20e+00} & 2.16e+00 & \underline{3.33e+00} & 5.95e+00 & \underline{1.18e+00} & \textbf{2.12e+00} & \underline{3.29e+00} & 5.96e+00 \\
HRES & 1.64e+00 & 2.68e+00 & 3.97e+00 & 6.68e+00 & 1.63e+00 & 2.64e+00 & 3.93e+00 & 6.67e+00 \\
Pangu & 1.43e+00 & 2.36e+00 & 3.60e+00 & 6.38e+00 & 1.42e+00 & 2.33e+00 & 3.58e+00 & 6.38e+00 \\
\bottomrule
\end{tabular}
\end{adjustbox}

\begin{adjustbox}{max width=\linewidth}
\begin{tabular}{lcccccccc}
\toprule
 & \multicolumn{8}{c}{\textbf{$v_{850}$}} \\
\cmidrule(lr){2-9}
 & \multicolumn{4}{c}{2020} & \multicolumn{4}{c}{2022} \\
\cmidrule(lr){2-5}\cmidrule(lr){6-9}
Model & 24h & 72h & 120h & 240h & 24h & 72h & 120h & 240h \\
\midrule
PARADIS & \textbf{1.18e+00} & \textbf{2.14e+00} & 3.38e+00 & 6.22e+00 & \textbf{1.18e+00} & 2.14e+00 & 3.36e+00 & 6.22e+00 \\
Aurora & - & - & - & - & 1.34e+00 & \textbf{2.13e+00} & \textbf{3.20e+00} & \textbf{5.20e+00} \\
Baguan & 1.23e+00 & 2.16e+00 & \textbf{3.28e+00} & \textbf{5.03e+00} & - & - & - & - \\
Climatology & 5.60e+00 & 5.60e+00 & 5.59e+00 & 5.59e+00 & 5.60e+00 & 5.59e+00 & 5.59e+00 & \underline{5.60e+00} \\
IFS ens & 1.99e+00 & 3.26e+00 & 4.66e+00 & 7.05e+00 & 1.97e+00 & 3.24e+00 & 4.63e+00 & 7.07e+00 \\
FGN (member 1) & - & - & - & - & 1.76e+00 & 2.72e+00 & 4.03e+00 & 6.78e+00 \\
FuXi & 1.25e+00 & 2.20e+00 & 3.39e+00 & \underline{5.11e+00} & - & - & - & - \\
GenCast (member 1) & 1.71e+00 & 2.93e+00 & 4.32e+00 & 6.89e+00 & - & - & - & - \\
GraphCast & \underline{1.20e+00} & \underline{2.15e+00} & \underline{3.33e+00} & 5.94e+00 & \underline{1.19e+00} & \underline{2.13e+00} & \underline{3.32e+00} & 6.02e+00 \\
HRES & 1.63e+00 & 2.65e+00 & 3.96e+00 & 6.71e+00 & 1.62e+00 & 2.63e+00 & 3.94e+00 & 6.70e+00 \\
Pangu & 1.42e+00 & 2.35e+00 & 3.62e+00 & 6.38e+00 & 1.42e+00 & 2.34e+00 & 3.61e+00 & 6.44e+00 \\
\bottomrule
\end{tabular}
\end{adjustbox}

\renewcommand{\arraystretch}{1.0}
\setlength{\tabcolsep}{6pt}
\label{tab:paradis_rmse_full}
\end{table*}

\begin{table*}
\centering
\tiny
\caption{ACC for selected atmospheric variables at multiple pressure levels, evaluated across forecast lead times. Bold values denote the best-performing model; underlined values denote the second-best.}
\renewcommand{\arraystretch}{0.8}
\setlength{\tabcolsep}{6pt}
\begin{adjustbox}{max width=\linewidth}
\begin{tabular}{lcccccccc}
\toprule
 & \multicolumn{8}{c}{\textbf{$z_{500}$}} \\
\cmidrule(lr){2-9}
 & \multicolumn{4}{c}{2020} & \multicolumn{4}{c}{2022} \\
\cmidrule(lr){2-5}\cmidrule(lr){6-9}
Model & 24h & 72h & 120h & 240h & 24h & 72h & 120h & 240h \\
\midrule
PARADIS & \textbf{9.99e-01} & \textbf{9.90e-01} & \textbf{9.44e-01} & 5.74e-01 & \textbf{9.99e-01} & \textbf{9.89e-01} & \underline{9.43e-01} & 5.52e-01 \\
Aurora & - & - & - & - & 9.99e-01 & \underline{9.89e-01} & \textbf{9.45e-01} & \textbf{5.88e-01} \\
Baguan & 9.99e-01 & 9.88e-01 & \underline{9.42e-01} & \textbf{6.54e-01} & - & - & - & - \\
IFS ens & 9.97e-01 & 9.71e-01 & 8.83e-01 & 4.39e-01 & 9.97e-01 & 9.71e-01 & 8.81e-01 & 4.27e-01 \\
FGN (member 1) & - & - & - & - & 9.98e-01 & 9.84e-01 & 9.24e-01 & 5.01e-01 \\
FuXi & 9.99e-01 & 9.88e-01 & 9.41e-01 & \underline{6.45e-01} & - & - & - & - \\
GenCast (member 1) & 9.98e-01 & 9.79e-01 & 9.06e-01 & 4.75e-01 & - & - & - & - \\
GraphCast & \underline{9.99e-01} & \underline{9.88e-01} & 9.42e-01 & 5.73e-01 & \underline{9.99e-01} & 9.88e-01 & 9.42e-01 & \underline{5.65e-01} \\
HRES & 9.98e-01 & 9.86e-01 & 9.29e-01 & 5.14e-01 & 9.98e-01 & 9.85e-01 & 9.26e-01 & 5.01e-01 \\
Pangu & 9.98e-01 & 9.86e-01 & 9.34e-01 & 5.39e-01 & 9.98e-01 & 9.86e-01 & 9.32e-01 & 5.18e-01 \\
\bottomrule
\end{tabular}
\end{adjustbox}

\begin{adjustbox}{max width=\linewidth}
\begin{tabular}{lcccccccc}
\toprule
 & \multicolumn{8}{c}{\textbf{$t_{850}$}} \\
\cmidrule(lr){2-9}
 & \multicolumn{4}{c}{2020} & \multicolumn{4}{c}{2022} \\
\cmidrule(lr){2-5}\cmidrule(lr){6-9}
Model & 24h & 72h & 120h & 240h & 24h & 72h & 120h & 240h \\
\midrule
PARADIS & \underline{9.85e-01} & 9.54e-01 & 8.82e-01 & 4.77e-01 & \underline{9.84e-01} & \underline{9.53e-01} & 8.81e-01 & 4.75e-01 \\
Aurora & - & - & - & - & 9.72e-01 & 9.50e-01 & \textbf{8.86e-01} & \textbf{5.13e-01} \\
Baguan & 9.83e-01 & \underline{9.54e-01} & \textbf{8.87e-01} & \textbf{5.76e-01} & - & - & - & - \\
IFS ens & 9.53e-01 & 8.93e-01 & 7.76e-01 & 3.50e-01 & 9.46e-01 & 8.87e-01 & 7.70e-01 & 3.46e-01 \\
FGN (member 1) & - & - & - & - & 9.60e-01 & 9.27e-01 & 8.40e-01 & 4.14e-01 \\
FuXi & 9.83e-01 & 9.52e-01 & 8.80e-01 & \underline{5.52e-01} & - & - & - & - \\
GenCast (member 1) & 9.70e-01 & 9.19e-01 & 8.16e-01 & 3.85e-01 & - & - & - & - \\
GraphCast & \textbf{9.85e-01} & \textbf{9.55e-01} & \underline{8.85e-01} & 4.78e-01 & \textbf{9.85e-01} & \textbf{9.56e-01} & \underline{8.86e-01} & \underline{4.81e-01} \\
HRES & 9.69e-01 & 9.30e-01 & 8.43e-01 & 4.20e-01 & 9.62e-01 & 9.24e-01 & 8.38e-01 & 4.20e-01 \\
Pangu & 9.78e-01 & 9.44e-01 & 8.64e-01 & 4.47e-01 & 9.77e-01 & 9.43e-01 & 8.62e-01 & 4.36e-01 \\
\bottomrule
\end{tabular}
\end{adjustbox}

\begin{adjustbox}{max width=\linewidth}
\begin{tabular}{lcccccccc}
\toprule
 & \multicolumn{8}{c}{\textbf{$q_{700}$}} \\
\cmidrule(lr){2-9}
 & \multicolumn{4}{c}{2020} & \multicolumn{4}{c}{2022} \\
\cmidrule(lr){2-5}\cmidrule(lr){6-9}
Model & 24h & 72h & 120h & 240h & 24h & 72h & 120h & 240h \\
\midrule
PARADIS & \underline{9.37e-01} & 8.33e-01 & 7.12e-01 & 3.62e-01 & \underline{9.37e-01} & 8.37e-01 & 7.20e-01 & 3.73e-01 \\
Aurora & - & - & - & - & 9.22e-01 & \underline{8.44e-01} & \textbf{7.42e-01} & \textbf{4.35e-01} \\
Baguan & 9.33e-01 & \underline{8.39e-01} & \textbf{7.32e-01} & \textbf{4.70e-01} & - & - & - & - \\
IFS ens & 8.33e-01 & 6.66e-01 & 5.06e-01 & 2.05e-01 & 8.43e-01 & 6.83e-01 & 5.25e-01 & 2.23e-01 \\
FGN (member 1) & - & - & - & - & 8.66e-01 & 7.52e-01 & 6.23e-01 & 3.05e-01 \\
GenCast (member 1) & 8.80e-01 & 7.15e-01 & 5.68e-01 & 2.66e-01 & - & - & - & - \\
GraphCast & \textbf{9.39e-01} & \textbf{8.40e-01} & \underline{7.26e-01} & \underline{3.83e-01} & \textbf{9.40e-01} & \textbf{8.48e-01} & \underline{7.38e-01} & \underline{3.99e-01} \\
HRES & 8.72e-01 & 7.35e-01 & 5.91e-01 & 2.62e-01 & 8.79e-01 & 7.50e-01 & 6.10e-01 & 2.90e-01 \\
Pangu & 9.20e-01 & 8.15e-01 & 6.85e-01 & 3.31e-01 & 9.21e-01 & 8.20e-01 & 6.93e-01 & 3.44e-01 \\
\bottomrule
\end{tabular}
\end{adjustbox}

\begin{adjustbox}{max width=\linewidth}
\begin{tabular}{lcccccccc}
\toprule
 & \multicolumn{8}{c}{\textbf{$u_{850}$}} \\
\cmidrule(lr){2-9}
 & \multicolumn{4}{c}{2020} & \multicolumn{4}{c}{2022} \\
\cmidrule(lr){2-5}\cmidrule(lr){6-9}
Model & 24h & 72h & 120h & 240h & 24h & 72h & 120h & 240h \\
\midrule
PARADIS & \textbf{9.79e-01} & \textbf{9.30e-01} & 8.23e-01 & 3.92e-01 & \textbf{9.79e-01} & \underline{9.30e-01} & 8.25e-01 & 3.90e-01 \\
Aurora & - & - & - & - & 9.73e-01 & 9.30e-01 & \textbf{8.37e-01} & \textbf{4.48e-01} \\
Baguan & 9.77e-01 & \underline{9.29e-01} & \textbf{8.31e-01} & \textbf{5.06e-01} & - & - & - & - \\
IFS ens & 9.41e-01 & 8.39e-01 & 6.78e-01 & 2.58e-01 & 9.41e-01 & 8.40e-01 & 6.80e-01 & 2.69e-01 \\
FGN (member 1) & - & - & - & - & 9.54e-01 & 8.90e-01 & 7.61e-01 & 3.30e-01 \\
FuXi & 9.77e-01 & 9.26e-01 & 8.21e-01 & \underline{4.88e-01} & - & - & - & - \\
GenCast (member 1) & 9.57e-01 & 8.71e-01 & 7.24e-01 & 3.00e-01 & - & - & - & - \\
GraphCast & \underline{9.79e-01} & 9.28e-01 & \underline{8.24e-01} & 4.01e-01 & \underline{9.79e-01} & \textbf{9.30e-01} & \underline{8.26e-01} & \underline{4.05e-01} \\
HRES & 9.60e-01 & 8.92e-01 & 7.62e-01 & 3.25e-01 & 9.60e-01 & 8.94e-01 & 7.65e-01 & 3.29e-01 \\
Pangu & 9.69e-01 & 9.15e-01 & 8.00e-01 & 3.64e-01 & 9.69e-01 & 9.16e-01 & 8.00e-01 & 3.58e-01 \\
\bottomrule
\end{tabular}
\end{adjustbox}

\begin{adjustbox}{max width=\linewidth}
\begin{tabular}{lcccccccc}
\toprule
 & \multicolumn{8}{c}{\textbf{$v_{850}$}} \\
\cmidrule(lr){2-9}
 & \multicolumn{4}{c}{2020} & \multicolumn{4}{c}{2022} \\
\cmidrule(lr){2-5}\cmidrule(lr){6-9}
Model & 24h & 72h & 120h & 240h & 24h & 72h & 120h & 240h \\
\midrule
PARADIS & \textbf{9.78e-01} & \textbf{9.25e-01} & 8.06e-01 & 3.34e-01 & \textbf{9.77e-01} & 9.24e-01 & 8.08e-01 & 3.20e-01 \\
Aurora & - & - & - & - & 9.71e-01 & \textbf{9.25e-01} & \textbf{8.20e-01} & \textbf{3.88e-01} \\
Baguan & 9.76e-01 & 9.23e-01 & \textbf{8.13e-01} & \textbf{4.46e-01} & - & - & - & - \\
IFS ens & 9.38e-01 & 8.31e-01 & 6.52e-01 & 2.00e-01 & 9.38e-01 & 8.33e-01 & 6.58e-01 & 2.01e-01 \\
FGN (member 1) & - & - & - & - & 9.51e-01 & 8.82e-01 & 7.40e-01 & 2.65e-01 \\
FuXi & 9.75e-01 & 9.20e-01 & 8.03e-01 & \underline{4.28e-01} & - & - & - & - \\
GenCast (member 1) & 9.53e-01 & 8.62e-01 & 6.99e-01 & 2.35e-01 & - & - & - & - \\
GraphCast & \underline{9.77e-01} & \underline{9.23e-01} & \underline{8.07e-01} & 3.36e-01 & \underline{9.77e-01} & \underline{9.25e-01} & \underline{8.09e-01} & \underline{3.29e-01} \\
HRES & 9.57e-01 & 8.86e-01 & 7.44e-01 & 2.63e-01 & 9.58e-01 & 8.87e-01 & 7.45e-01 & 2.65e-01 \\
Pangu & 9.67e-01 & 9.09e-01 & 7.80e-01 & 3.01e-01 & 9.67e-01 & 9.09e-01 & 7.81e-01 & 2.90e-01 \\
\bottomrule
\end{tabular}
\end{adjustbox}

\renewcommand{\arraystretch}{1.0}
\setlength{\tabcolsep}{6pt}
\label{tab:paradis_acc_full}
\end{table*}

\clearpage 




\section{Methods}

\subsection{Spherical Geometry}\label{appendix:padding}

The implementation of the semi-Lagrangian scheme in spherical coordinates is inspired by \cite{mcdonald1989semi}, who introduced an ``auxiliary spherical coordinate system'' to address difficulties near the poles. For each arrival point $(\phi_a, \lambda_a)$, a rotated coordinate system $(\phi', \lambda')$ is defined, where the origin coincides with the arrival point, and the equator of this rotated system passes through the point in question. In this local system, $\lambda'$ measures the angular distance along the rotated equator, while $\phi'$ measures the angular distance perpendicular to it.

When applying the geocyclic padding described in \Cref{subsec:geometry}, there is a difficulty in handling wind vectors near the Earth's poles due to the convergence of meridians, which creates artificial discontinuities in the latitude--longitude coordinate system. This would force the neural networks to learn unnatural, discontinuous transformations in their latent space representations. Hence, we propose to transform wind vectors from spherical coordinates into Cartesian coordinates in a pre-processing step. This transformation avoids the coordinate singularity at the poles and ensures a locally continuous representation of winds. The transformation from spherical velocity components $(u, v, w)$ to Cartesian components $(u_x, u_y, u_z)$ is given by
\begin{align*}
    u_x &= -u\sin(\lambda) - v\sin(\phi)\cos(\lambda) - w\cos(\phi)\cos(\lambda), \\
    u_y &= u\cos(\lambda) - v\sin(\phi)\sin(\lambda) - w\cos(\phi)\sin(\lambda), \\
    u_z &= v\cos(\phi) - w\sin(\phi).
\end{align*}
The neural network operates on these continuous Cartesian components. During post-processing, its output is transformed back to spherical coordinates. The inverse transformation is given by
\begin{align*}
    u &= -u_x\sin(\lambda) + u_y\cos(\lambda), \\
    v &= -u_x\sin(\phi)\cos(\lambda) - u_y\sin(\phi)\sin(\lambda) + u_z\cos(\phi), \\
    w &= -u_x\cos(\phi)\cos(\lambda) - u_y\cos(\phi)\sin(\lambda) - u_z\sin(\phi).
\end{align*}

\subsection{Radiation Parameterization}
\label{subsec:radiation}

Following the approach of \cite{lam2023learning}, we include the one-hour accumulated top-of-atmosphere (TOA) incoming solar radiation as an external conditioning channel for the forecast model. This variable provides the network with an implicit representation of the time of day and seasonality. Because the model is not provided with sufficient information to compute a full radiative balance, this signal serves primarily as a temporal and geometric proxy rather than a physically complete radiation scheme.

The computation closely follows \cite{wald_basics_2019}, accounting for the ellipticity of Earth's orbit while neglecting variability in the solar cycle. Let $\phi$ denote latitude, $\lambda$ longitude, and $T$ the Julian day referenced to 1 January 2000 at 12:00 UTC. The instantaneous TOA incoming solar radiation, $R$ is given by
\begin{equation}\label{eqn:toa_rad_local}
\begin{aligned}
R(\phi,\lambda,T) &=
\frac{1360.56}{d(T)^2}\,
\max\Bigl(0,\;
\sin(\phi)\sin(\delta(T)) \\
&\qquad + \cos(\phi)\cos(\delta(T))
\cos\!\bigl(t(T,\lambda)\bigr)
\Bigr).
\end{aligned}
\end{equation}
The orbital and solar geometry parameters required to evaluate \eqref{eqn:toa_rad_local}---including the Earth--Sun distance $d(T)$ and solar declination $\delta(T)$---are summarized in \Cref{tab:orbital_params}. All angular quantities are reduced to the interval $[-\pi,\pi]$ during computation.

The local solar time $t(T,\lambda)$ depends on the equation of time $\text{EOT}(T) = L(T) - \alpha(T)$, and is computed as:
\begin{equation}
t(T,\lambda) = \lambda + 2\pi T +\text{EOT}(T).
\end{equation}
The final conditioning variable provided to the model corresponds to the one-hour accumulated TOA radiation, obtained by integrating \eqref{eqn:toa_rad_local} over the one-hour interval ending at the specified time.

\begin{table}[ht]
\centering
\caption{Orbital and solar geometry parameters used for radiation evaluation.}
\label{tab:orbital_params}
\addtolength{\tabcolsep}{5pt}
\begin{adjustbox}{max width=\linewidth}
\begin{tabular}{ll}
\toprule
\textbf{Parameter} & \textbf{Equation} \\
\midrule
Obliquity & $\epsilon(T) = \frac{\pi}{180}(23.439 - 3.6\times10^{-7}T)$ \\
Mean Anomaly & $M(T) = \frac{\pi}{180}(357.529 + 0.985600028T)$ \\
Mean Longitude & $L(T) = \frac{\pi}{180}(280.459 + 0.98564736T)$ \\
Sun Longitude & $\lambda_\odot(T) = L(T) + \frac{\pi}{180} (1.915\sin M + 0.020\sin 2M)$ \\
Earth--Sun Dist. & $d(T) = 1.00014 - 0.01671\cos M - 1.4\times10^{-4}\cos 2M$ \\
Right Ascension & $\alpha(T) = \arctan(\cos \epsilon \tan \lambda_\odot)$ \\
Declination & $\delta(T) = \arcsin(\sin \epsilon \sin \lambda_\odot)$ \\
\bottomrule
\end{tabular}
\end{adjustbox}
\end{table}

\section{Visual Forecast Examples}\label{appendix:visual_examples}

\begin{figure}[tbh]
\centering

\begin{subfigure}{\linewidth}
  \centering
  \hspace{\fill} Ground truth \hspace{10em} \hspace{\fill} PARADIS \hspace{\fill}\, \\[1ex]
  \includegraphics[width=\linewidth]{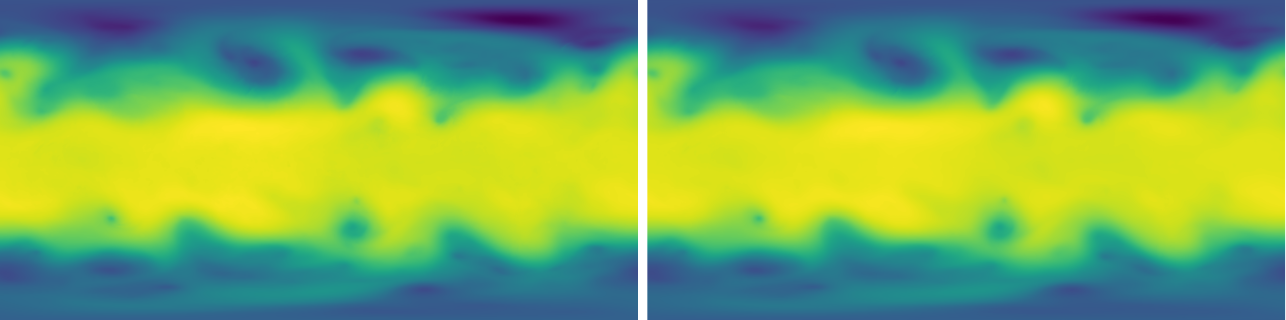}
  \caption{1 day}
\end{subfigure}

\vspace{0.5em}

\begin{subfigure}{\linewidth}
  \centering
  \includegraphics[width=\linewidth]{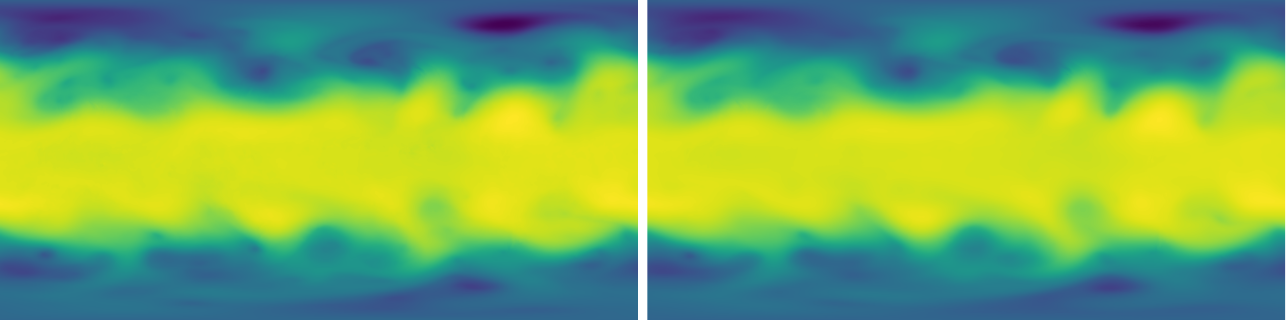}
  \caption{2 days}
\end{subfigure}

\vspace{0.5em}

\begin{subfigure}{\linewidth}
  \centering
  \includegraphics[width=\linewidth]{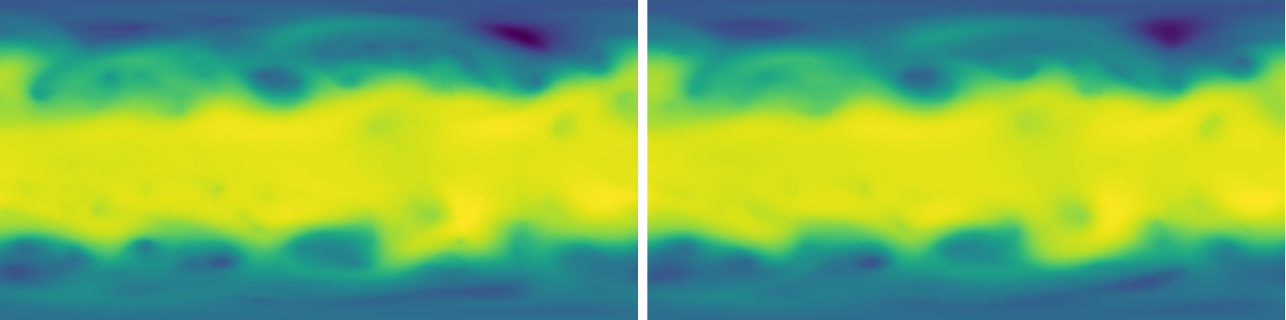}
  \caption{5 days}
\end{subfigure}
\caption{Field: Geopotential at 500 hPa. Left plot is the reference, right plot is PARADIS.}
\label{fig:l1}
\end{figure}

\begin{figure}[tbh]
\centering

\begin{subfigure}{\linewidth}
  \centering
  \hspace{\fill} Ground truth \hspace{10em} \hspace{\fill} PARADIS \hspace{\fill}\, \\[1ex]
  \includegraphics[width=\linewidth]{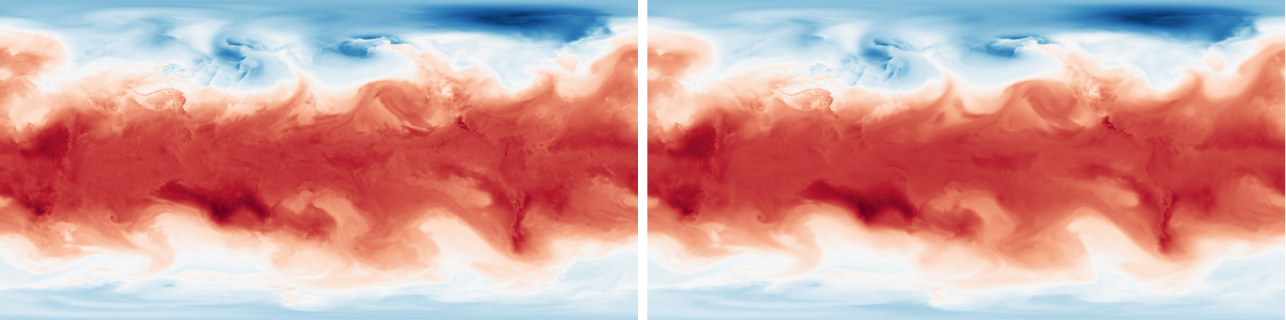}
  \caption{1 day}
\end{subfigure}

\vspace{0.5em}

\begin{subfigure}{\linewidth}
  \centering
  \includegraphics[width=\linewidth]{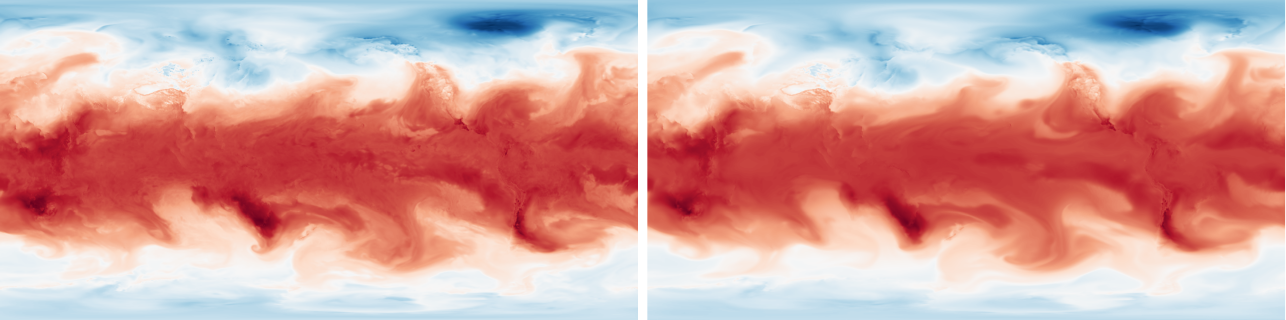}
  \caption{2 days}
\end{subfigure}

\vspace{0.5em}

\begin{subfigure}{\linewidth}
  \centering
  \includegraphics[width=\linewidth]{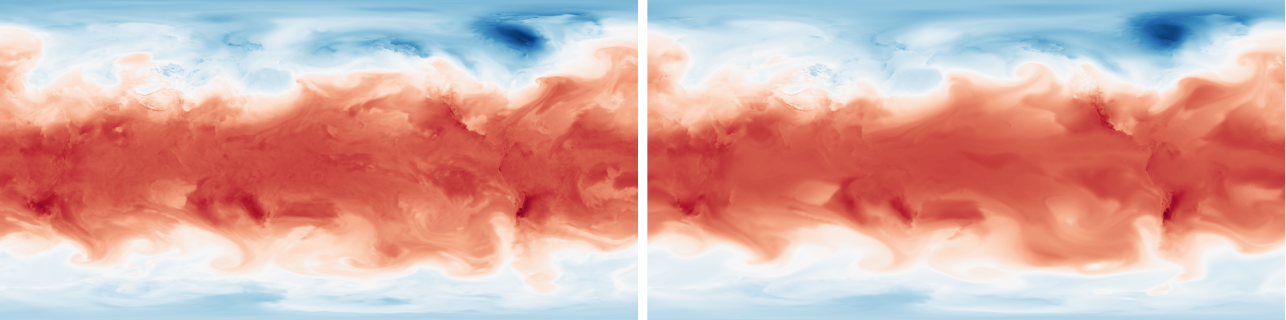}
  \caption{5 days}
\end{subfigure}
\caption{Field: Temperature at 850 hPa. Left plot is the reference, right plot is PARADIS.}
\label{fig:l2}
\end{figure}

\begin{figure}[tbh]
\centering

\begin{subfigure}{\linewidth}
  \centering
  \hspace{\fill} Ground truth \hspace{10em} \hspace{\fill} PARADIS \hspace{\fill}\, \\[1ex]
  \includegraphics[width=\linewidth]{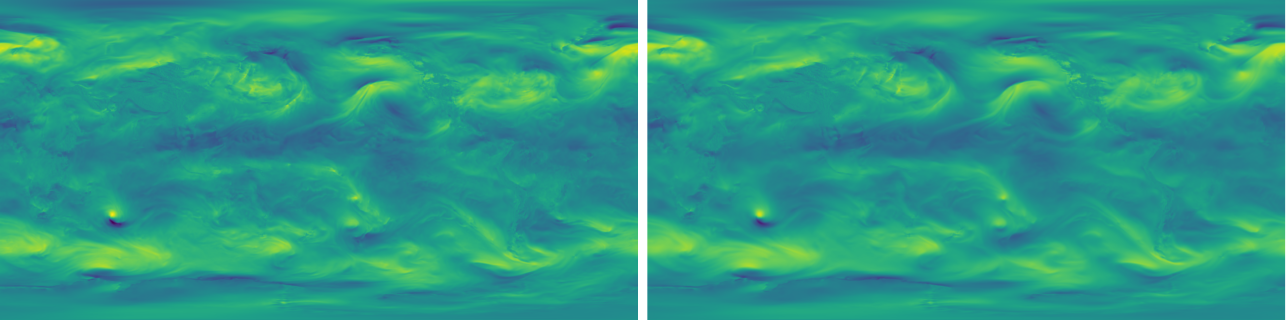}
  \caption{1 day}
\end{subfigure}

\vspace{0.5em}

\begin{subfigure}{\linewidth}
  \centering
  \includegraphics[width=\linewidth]{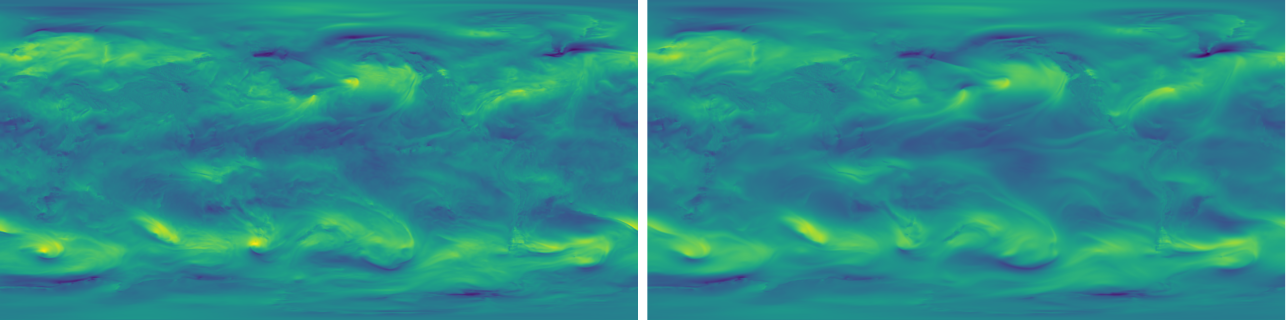}
  \caption{2 days}
\end{subfigure}

\vspace{0.5em}

\begin{subfigure}{\linewidth}
  \centering
  \includegraphics[width=\linewidth]{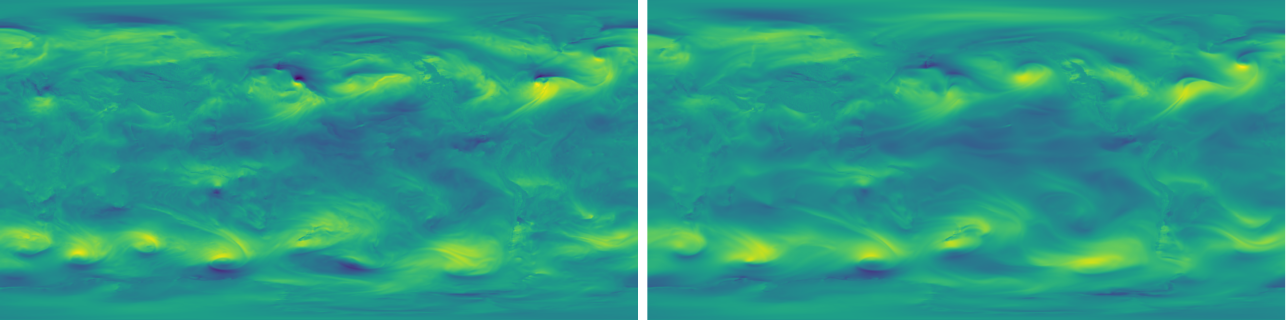}
  \caption{5 days}
\end{subfigure}
\caption{Field: $u$-component of wind at 850 hPa. Left plot is the reference, right plot is PARADIS.}
\label{fig:l3}
\end{figure}

\begin{figure}[tbh]
\centering

\begin{subfigure}{\linewidth}
  \centering
  \hspace{\fill} Ground truth \hspace{10em} \hspace{\fill} PARADIS \hspace{\fill}\, \\[1ex]
  \includegraphics[width=\linewidth]{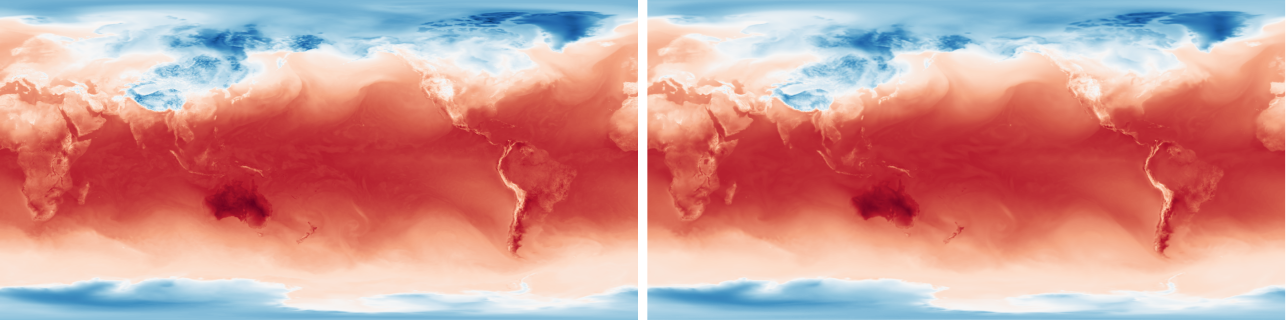}
  \caption{1 day}
\end{subfigure}

\vspace{0.5em}

\begin{subfigure}{\linewidth}
  \centering
  \includegraphics[width=\linewidth]{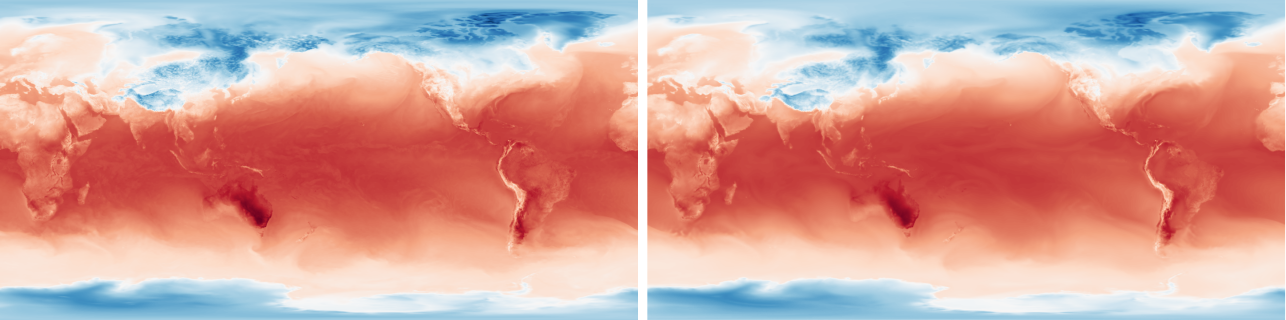}
  \caption{2 days}
\end{subfigure}

\vspace{0.5em}

\begin{subfigure}{\linewidth}
  \centering
  \includegraphics[width=\linewidth]{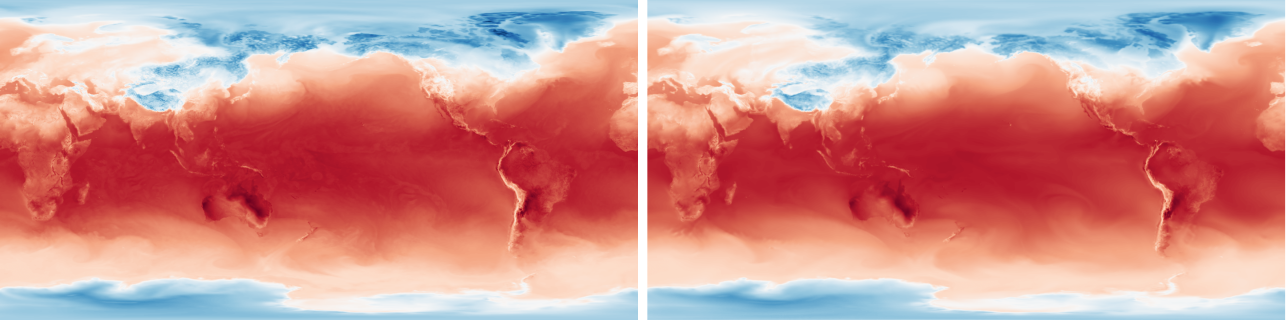}
  \caption{5 days}
\end{subfigure}
\caption{Field: $2m$-temperature. Left plot is the reference, right plot is PARADIS}
\label{fig:l4}
\end{figure}

For the sake of completeness, we present qualitative results for the PARADIS model against the ERA5 reference analysis in figures \Cref{fig:l1,fig:l2,fig:l3,fig:l4}. Each figure shows the evolution of a specific meteorological field at 1, 2, and 5-day lead times. For each time step, the subplots represent (left) the ERA5 reference and (right) the PARADIS forecast.

\section{Table of Symbols}

The following table summarizes the primary symbols and notations used to describe the implemented architecture and its underlying physical principles.
\begin{table}[ht]
    \centering
    \scriptsize
    \renewcommand{\arraystretch}{0.88}
    \begin{adjustbox}{max width=\linewidth}
\begin{tabular}{ll}
        \toprule
        \textbf{Symbol} & \textbf{Description / Definition} \\
        \midrule
        \multicolumn{2}{l}{\textbf{State Variables and Domains}} \\
        $\mathbf{q}$ & Physical state vector of atmospheric variables. \\
        $\mathbf{h}$ & Latent state vector/features. \\
        $\mathbf{x}$ & Grid point location. \\
        $\mathbf{x}_d$ & Departure point at time $t$ found by tracing trajectories backward. \\
        $\mathbf{u}$ & Velocity field or transport velocities satisfying $d\mathbf{x}/dt = \mathbf{u}$. \\
        $\phi$ &  Latitude. \\
        $\lambda$ &  Longitude. \\
        \midrule
        \multicolumn{2}{l}{\textbf{Model Operators}} \\
        $\mathcal{A}_{\text{net}}$, $\mathcal{A}$ & Neural Semi-Lagrangian (NSL) operator for advection. \\
        $\mathcal{D}_{\text{net}}$,  $\mathcal{D}$& Neural operator for diffusion-like spatial mixing. \\
        $\mathcal{R}_{\text{net}}$,  $\mathcal{R}$ & Neural operator for local reaction/source terms. \\
        $\mathcal{V}_{\text{net}}$ & Velocity network estimating transport fields from the latent state. \\
        $\mathcal{I}$ & Bicubic interpolation. \\
        $\nsubstep$ & Number of processor layers/substeps to advance from $t$ to $t+\Delta t$. \\
        $\mathcal{E}$ & Encoder mapping physical variables to latent space. \\
        $\mathcal{G}$ & Decoder mapping latent space back to physical space. \\
        $\frac{D}{Dt}$ & Material derivative ($\frac{\partial}{\partial t} + u \cdot \nabla$). \\
        \midrule
        \multicolumn{2}{l}{\textbf{Model Architecture}} \\
        $K$ & Rank for low-rank factorization of the global bias field. \\
        $H,W$ & Dimension of spatial grid. \\
        $\mathbf{W}$ & Weight matrix. \\
        $\mathbf{K}_d$ & A channel-wise spatial filter applied independently on each channel.\\
        $\mathbf{b}$ & Learned bias. \\
        $\mathbf{A}, \mathbf{U}, \mathbf{V}$ & Factor matrices for low-rank bias decomposition. \\
        $\mathbf{B}$ & Low-rank bias map. \\
        $\mathbf{z}$ & Projected state in latent space projection. \\
        \midrule
        \multicolumn{2}{l}{\textbf{Optimization and Training}} \\
        $L_\delta$ & Reversed Huber loss function. \\
        $\tilde{L}_\delta$ & reversed Huber loss function. \\
        $\delta$ & Threshold parameter for the Huber loss transition. \\
        $e$ & Prediction error. \\
        $w$ & A sigmoid weighting function. \\
        \midrule
        \multicolumn{2}{l}{\textbf{Meteorological variables}}  \\
        $z$ & Geopotential \\
        $t$ & Temperature \\
        $u$ & $u$-component of wind (zonal) \\
        $v$ & $v$-component of wind (meridional) \\
        $w$ & vertical velocity \\
        $q$ & Specific humidity \\
        $2t$& 2-metre temperature \\
        $10u$ & 10-metre $u$-component of wind \\
        $msl$ & Mean sea-level pressure \\
        \bottomrule
    \end{tabular}
\end{adjustbox}
    \caption{Summary of mathematical notation in the model architecture.}
    \label{tab:tab_of_symbols}
\end{table}

\end{document}